\documentclass[botnum, fleqn]{unmeethesis}
\usepackage{textcomp}
\usepackage{graphicx}
\usepackage{caption}
\usepackage{subcaption}
\usepackage{tabularx}
\usepackage{array,ragged2e}
\newcolumntype{P}[1]{>{\RaggedRight\arraybackslash}p{#1}}
\usepackage[utf8x]{inputenc}
\usepackage{float}
\usepackage{url}
\usepackage[ruled,vlined]{algorithm2e}
\usepackage{geometry}
\usepackage{algorithmic}
\usepackage{atbegshi}
\AtBeginDocument{\AtBeginShipoutNext{\AtBeginShipoutDiscard}}
\newcommand{\comment}[1]{}
\newcommand{\tabitem}{~~\llap{\textbullet}~~}

\begin{document}
\frontmatter


\title {Long Term Object Detection and Tracking in Collaborative Learning Environments}

\author{Sravani Teeparthi}

\degree{Master of Science \\ Computer Engineering}

\documenttype{Thesis}

\previousdegrees{B.Tech., Electronics and Communication Engineering, 2015}

\date{July, \thisyear}

\maketitle

\begin{dedication}
  To my husband and parents for their support and encouragement. \\[3ex]
\end{dedication}

\begin{acknowledgments}
  \vspace{1.1in} I would first like to thank my advisor Prof. Marios
  Pattichis for all his patience in advicing me. His guidance and
  ideas were crucial for my development as a graduate student.

  I would also like to acknowledge Miguel A Hombrados-Herrera,
  and Krishna Ashok-Poddar for their valuable insights and feedback.

  This material is based upon work supported by the National Science
  Foundation under Grant No. 1613637 and No. 1842220. Any opinions or
  findings of this thesis reflect the views of the author.  They do not
  necessarily reflect the views of NSF.

\end{acknowledgments}

\maketitleabstract 

\begin{abstract}
Human activity recognition in videos is a challenging problem that has
drawn a lot of interest, particularly when the goal requires the analysis of a large video database. The Advancing Out-of-School Learning in Mathematics and Engineering (AOLME) project provides a collaborative learning environment for middle school students to explore mathematics, computer science, and engineering by processing digital images and videos. As part of this project, around
2200 hours of video data were collected for analysis. This data
was collected to understand how children learn in situations involving
mathematical and programming challenges so as to recognize best
teaching practices that support broadening participation of
underrepresented students in STEM fields. Because of the size of
the dataset, it is hard to analyze all the videos of the dataset
manually. Thus, there is a huge need for reliable computer-based
methods that can detect activities of interest.  \par My thesis is
focused on the development of accurate methods for detecting and
tracking objects in collaborative learning environments in long videos
($>1$ hour). Long-term object detection and tracking face fundamental
challenges due to occlusion, illumination variations, and pose
variations.  \par For collaborative learning groups, the thesis
contributes robust methods for computer keyboard detection, tracking, and student hand detection. For hand detection, the thesis
integrates object detection with clustering and time-projections for
accurate, long-term assessment of student participation. The hand detection method was
integrated into a writing detection system and can also be used for
later research on recognizing student gestures.\par

All the models are validated on videos from 7 different sessions,
ranging from 45 minutes to 90 minutes. The keyboard detector achieved
a very high average precision (AP) of $92\%$ at 0.5 intersection over
union (IoU). Furthermore, a combined system of the detector with a
fast tracker KCF (159fps) was developed so that the algorithm runs
significantly faster without sacrificing accuracy. For a video of 23
minutes having resolution 858 $\times$ 480 @ 30 fps, the detection
alone runs at $4.7\times$the real-time, and the combined
algorithm runs at $21\times$the real-time for an average IoU of 0.84 and 0.82, respectively. The
hand detector achieved average precision (AP) of $72\%$ at 0.5
intersection over union (IoU). The detection results were improved to
$81\%$ using optimal data augmentation parameters. The hand detector runs at $4.7\times$the real-time with AP of $81\%$ at 0.5 intersection over union. The hand
detection method was integrated with projections and clustering for
accurate proposal generation. This approach reduced the number of false-positive hand detections by 80$\%$. The overall hand detection system runs at $4\times$the real-time, capturing all the activity regions of the current collaborative group.
  
  \clearpage 
\end{abstract}

\tableofcontents
\listoffigures
\listoftables

\begin{glossary}{Longest  string}
\item[IoU Ratio]Intersection over Union is an evaluation metric used to measure the accuracy of an object detector on a particular dataset. The IoU is the ratio of the overlapping area of ground truth and predicted area to the total area. 
\item[AOLME] The Advancing Out-of-School Learning in Mathematics and Engineering research study. 
\item[SOTA] State of the Art. 
\item[Binary Image] An image consisting only of black and white values.
\item[Centroid] The center point in a countour.
\item[Ground Truth] A set of labelled data that serves as a point of comparison 
\item[CNN] Convolutional Neural Network.
\item[Precision] Fraction of relevant instances among all retrieved instances
\item[Recall] Fraction of retrieved instances among all relevant instances
\item[AP] Performance evaluation terms computed for each category. AP is the mean of the precision scores after each relevant document is retrieved.
\item[AR] Average Recall.
\item[TP, FP] True Positive, False Positive
\item[TN, FN] True Negative, False Negative

\end{glossary} 

\mainmatter

\chapter{Introduction}
Object detection and tracking have advanced significantly over the last decade. It is possible to detect multiple objects in an image accurately thanks to advancements in deep learning \cite{aziz2020exploring}. Even though it works for images, accurate detection and tracking proved quite challenging for videos. In addition to image object detection problems, such as occlusion, scaling, variation in lighting, camera angle, we also face other video-related problems. These problems include but are not limited to (i) a large number of frames (100K images per hour), (ii) disappearance and reappearance of an object, and (iii) dynamic object pose variation.

\begin{figure}[]
  \begin{subfigure}{0.3\columnwidth}
    \centering
    \includegraphics[width=\textwidth]{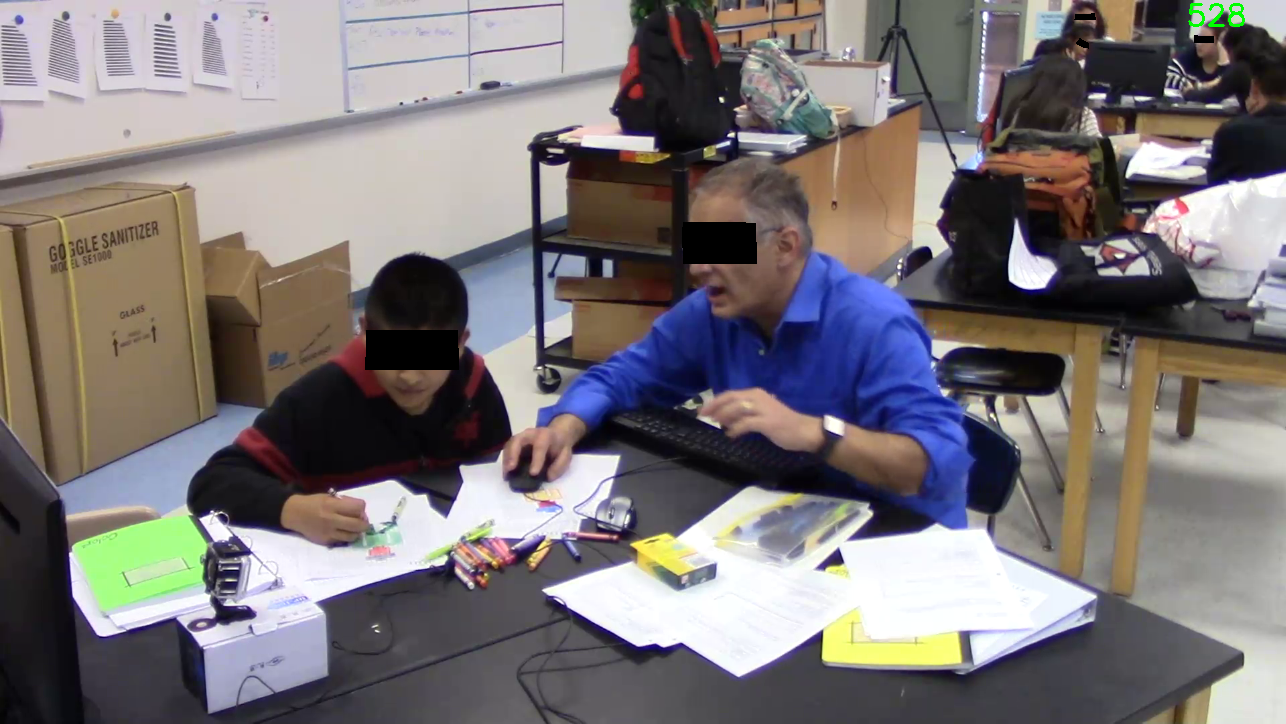}
    \caption{}
    \label{fig:1}
  \end{subfigure}\hfill
  \begin{subfigure}{0.3\columnwidth}
    \centering
    \includegraphics[width=\textwidth]{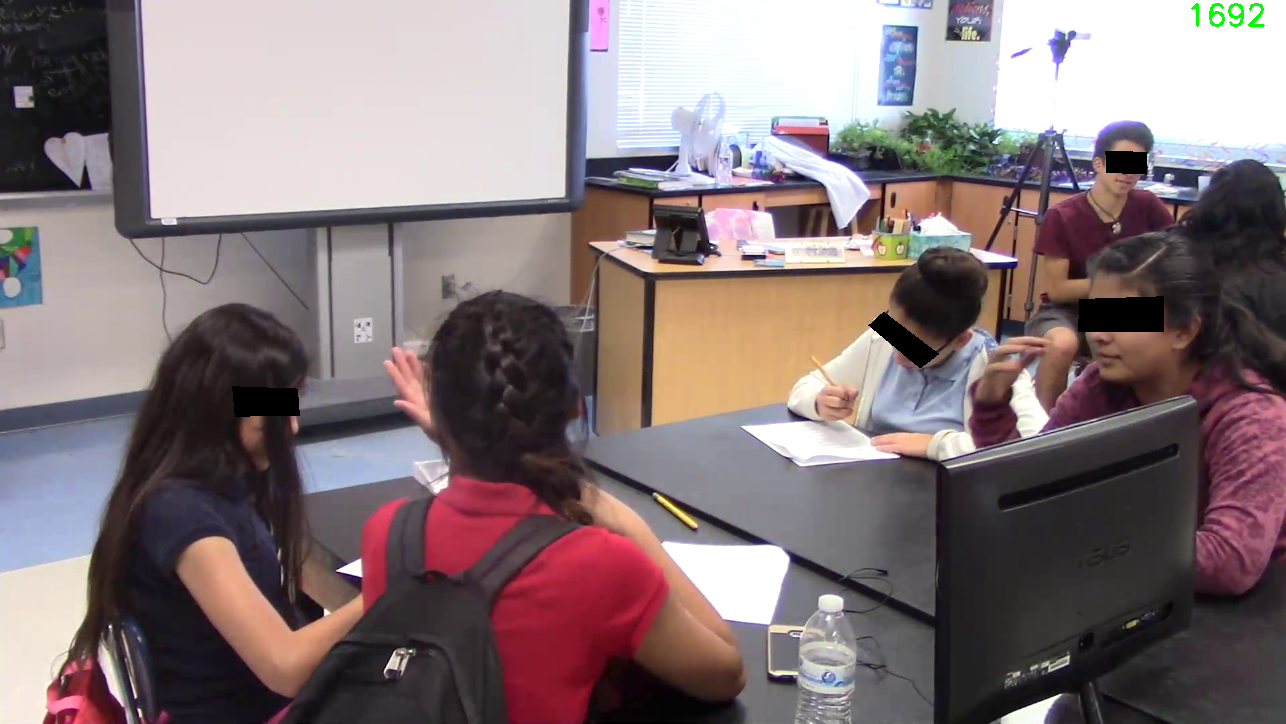}
    \caption{}
    \label{fig:2}
  \end{subfigure}\hfill
  \begin{subfigure}{0.3\columnwidth}
    \centering
    \includegraphics[width=\textwidth]{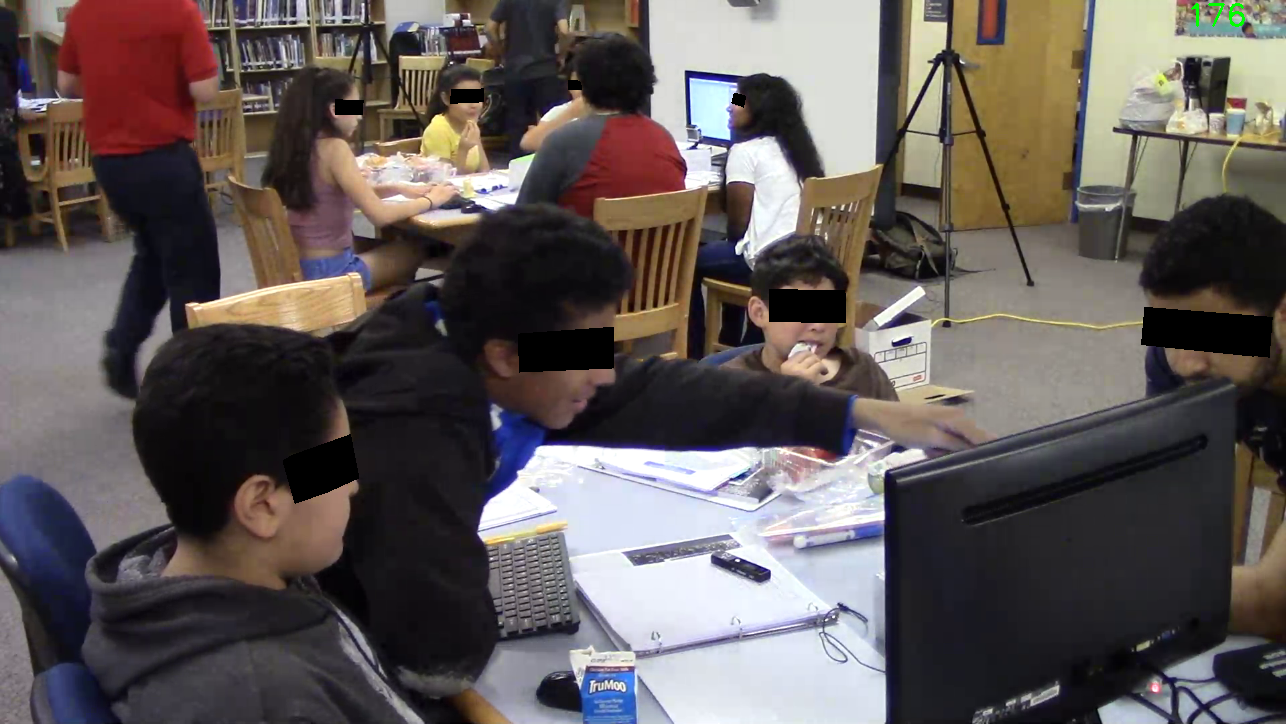}
    \caption{}
    \label{fig:3}
  \end{subfigure}\hfill
  
  \begin{subfigure}{0.3\columnwidth}
    \centering
    \includegraphics[width=\textwidth]{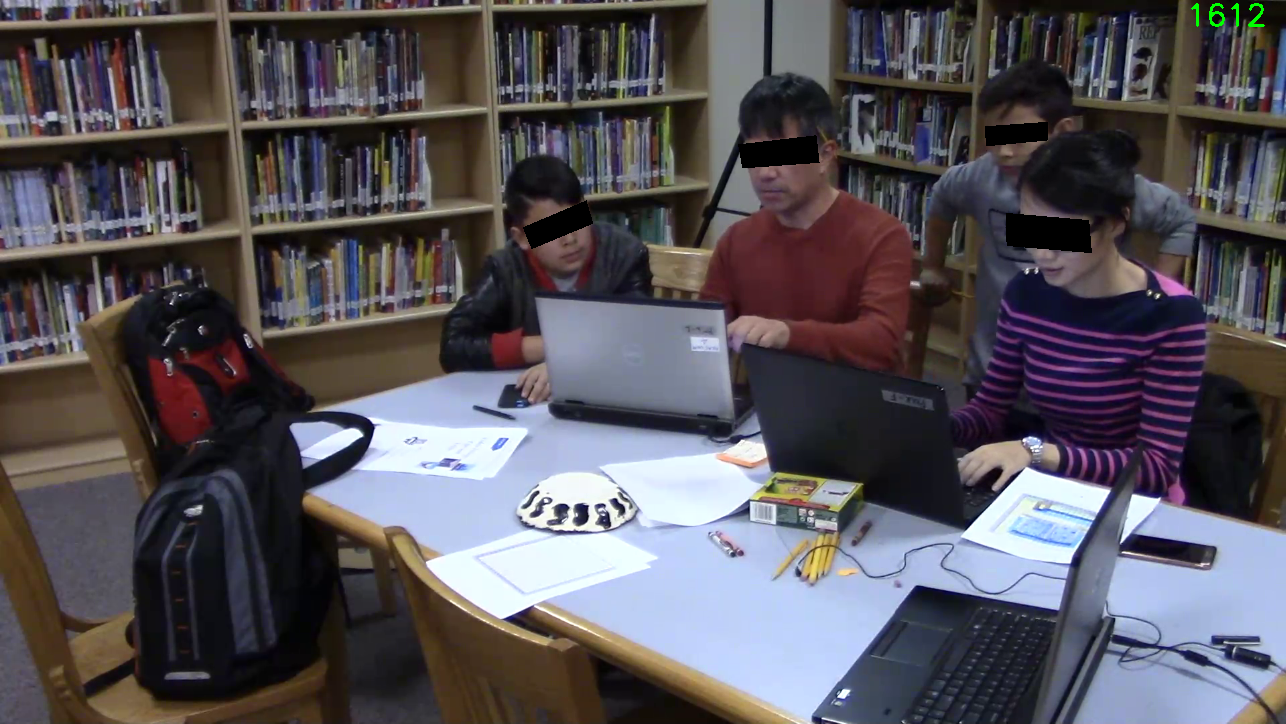}
    \caption{}
    \label{fig:4}
  \end{subfigure}\hfill
  \begin{subfigure}{0.3\columnwidth}
    \centering
    \includegraphics[width=\textwidth]{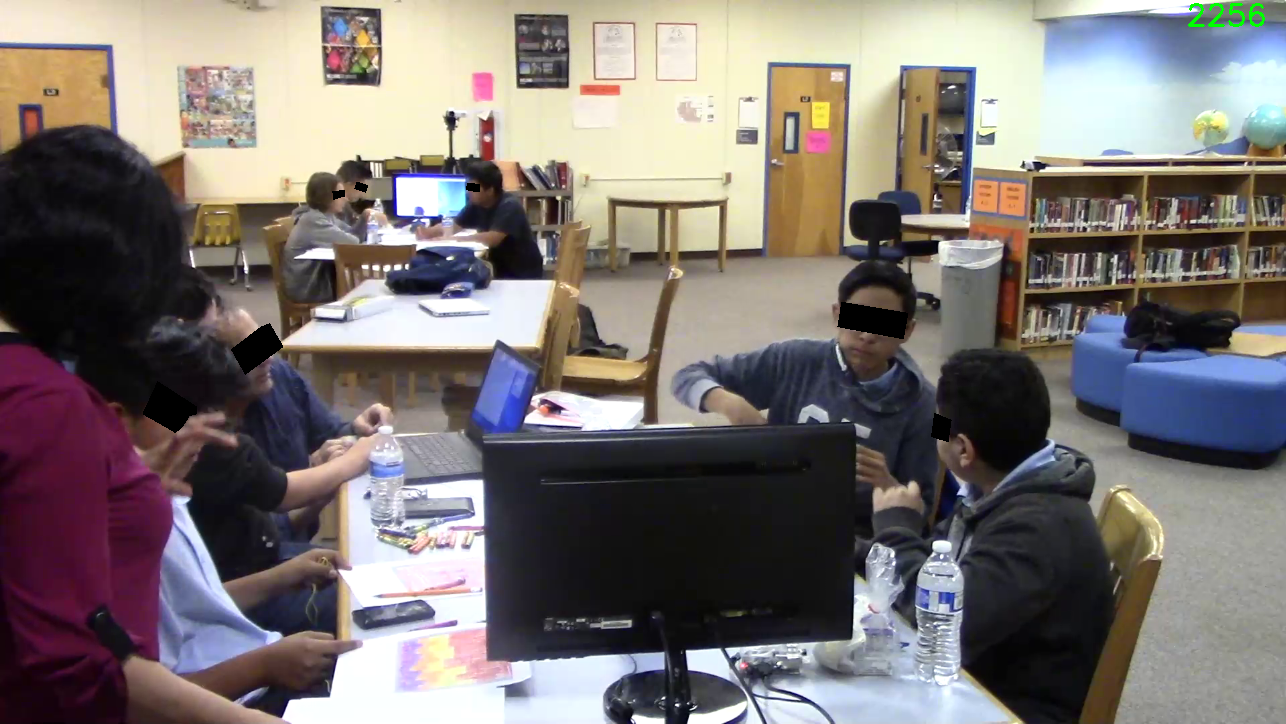}
    \caption{}
    \label{fig:5}
  \end{subfigure}\hfill
  \begin{subfigure}{0.3\columnwidth}
    \centering
    \includegraphics[width=\textwidth]{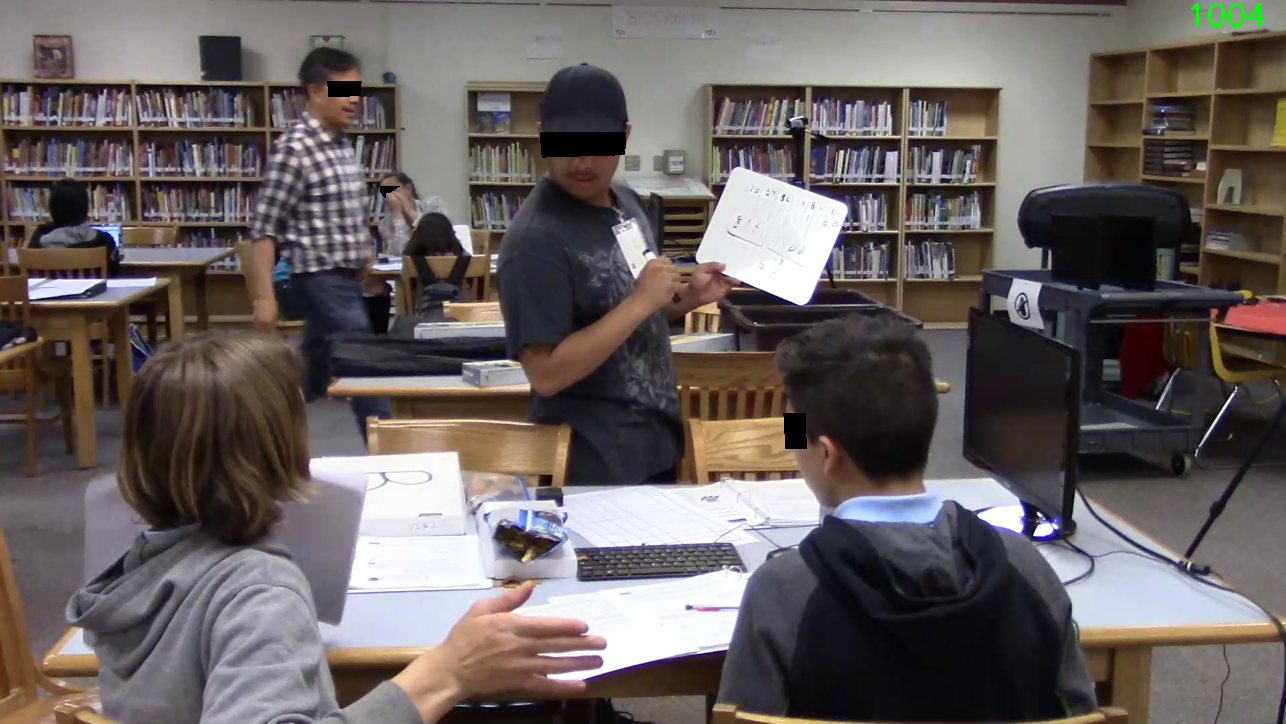}
    \caption{}
    \label{fig:6}
  \end{subfigure}\hfill

  \begin{subfigure}{0.3\columnwidth}
    \centering
    \includegraphics[width=\textwidth]{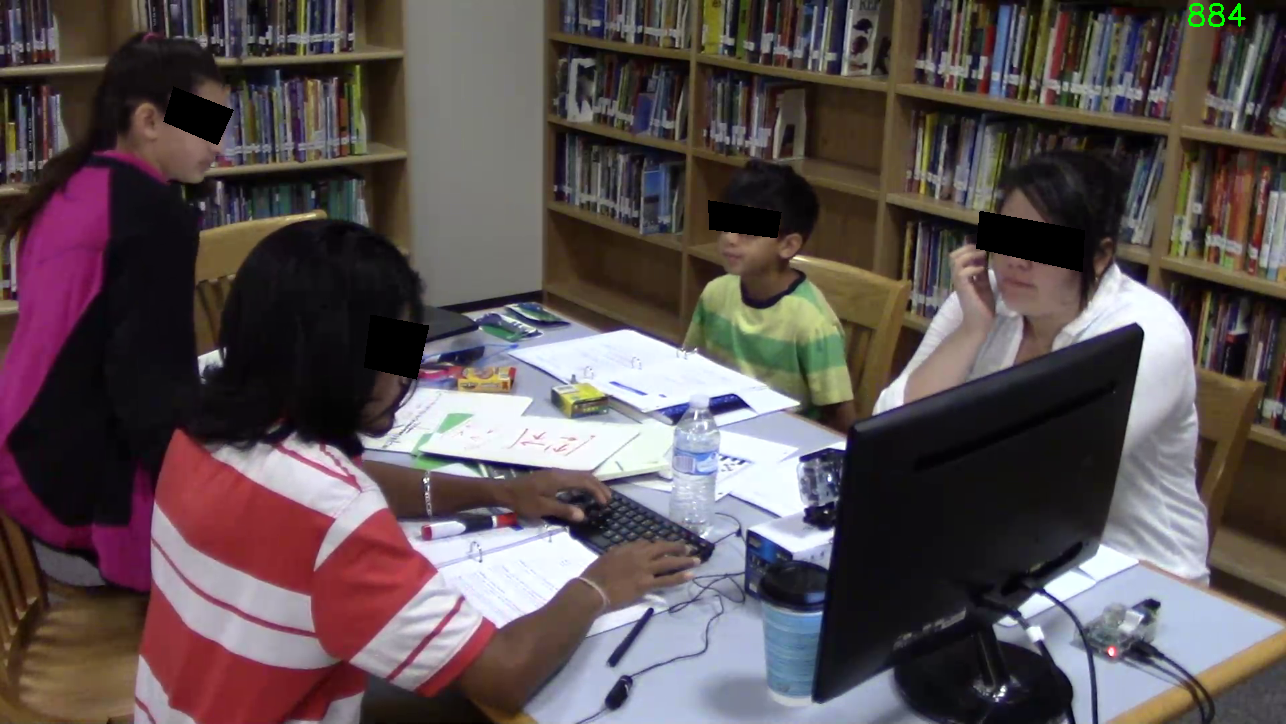}
    \caption{}
    \label{fig:7}
  \end{subfigure}\hfill
  \begin{subfigure}{0.3\columnwidth}
    \centering
    \includegraphics[width=\textwidth]{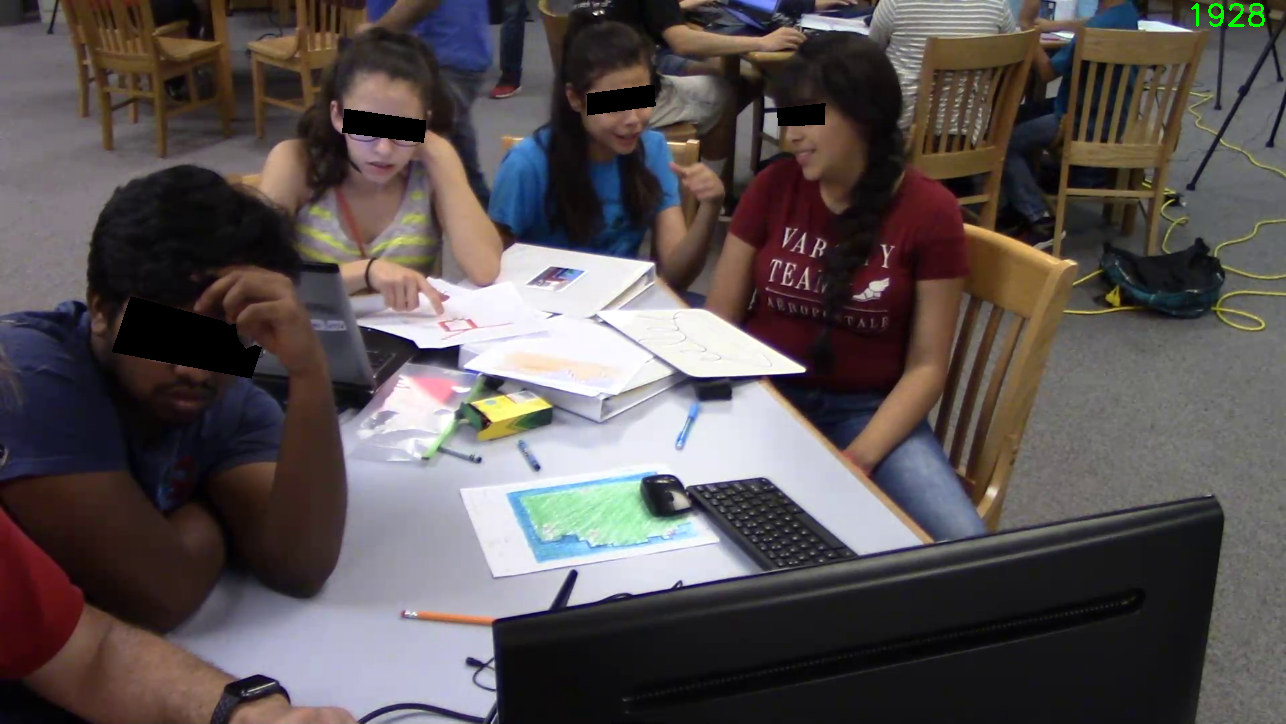}
    \caption{}
    \label{fig:8}
  \end{subfigure}\hfill
  \begin{subfigure}{0.3\columnwidth}
    \centering
    \includegraphics[width=\textwidth]{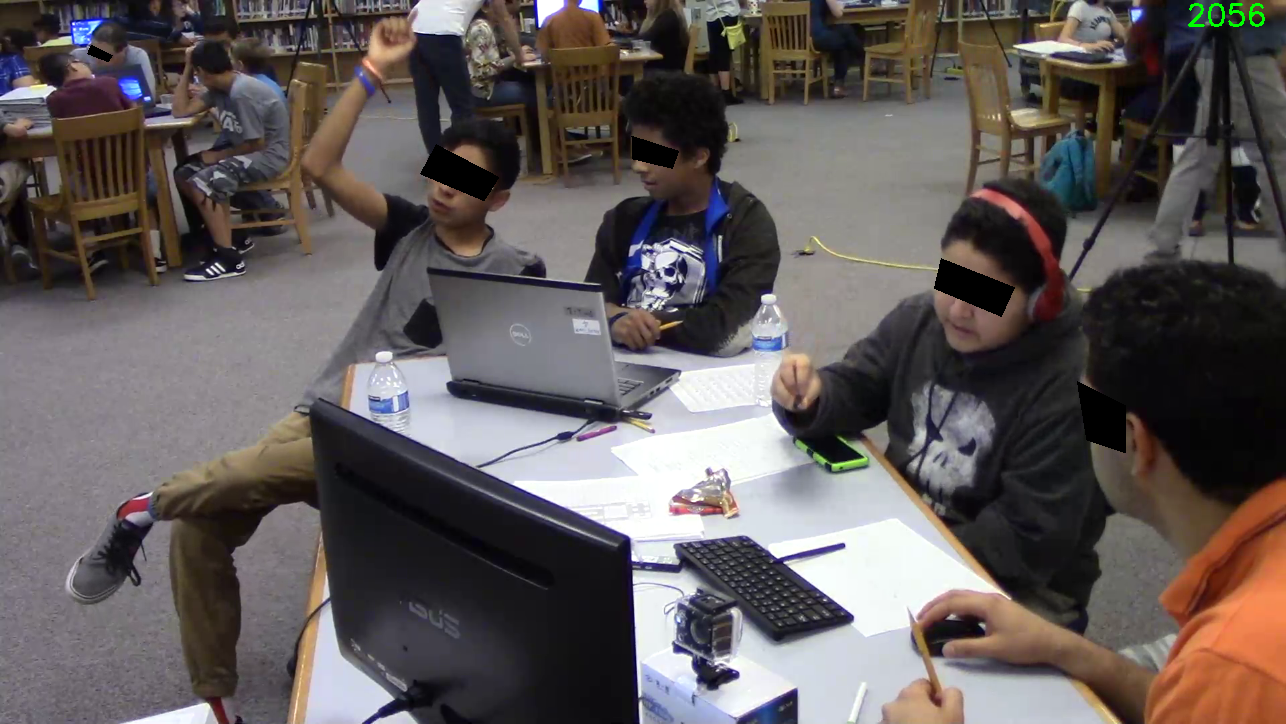}
    \caption{}
    \label{fig:9}
  \end{subfigure}\hfill
  
  \begin{subfigure}{0.3\columnwidth}
    \centering
    \includegraphics[width=\textwidth]{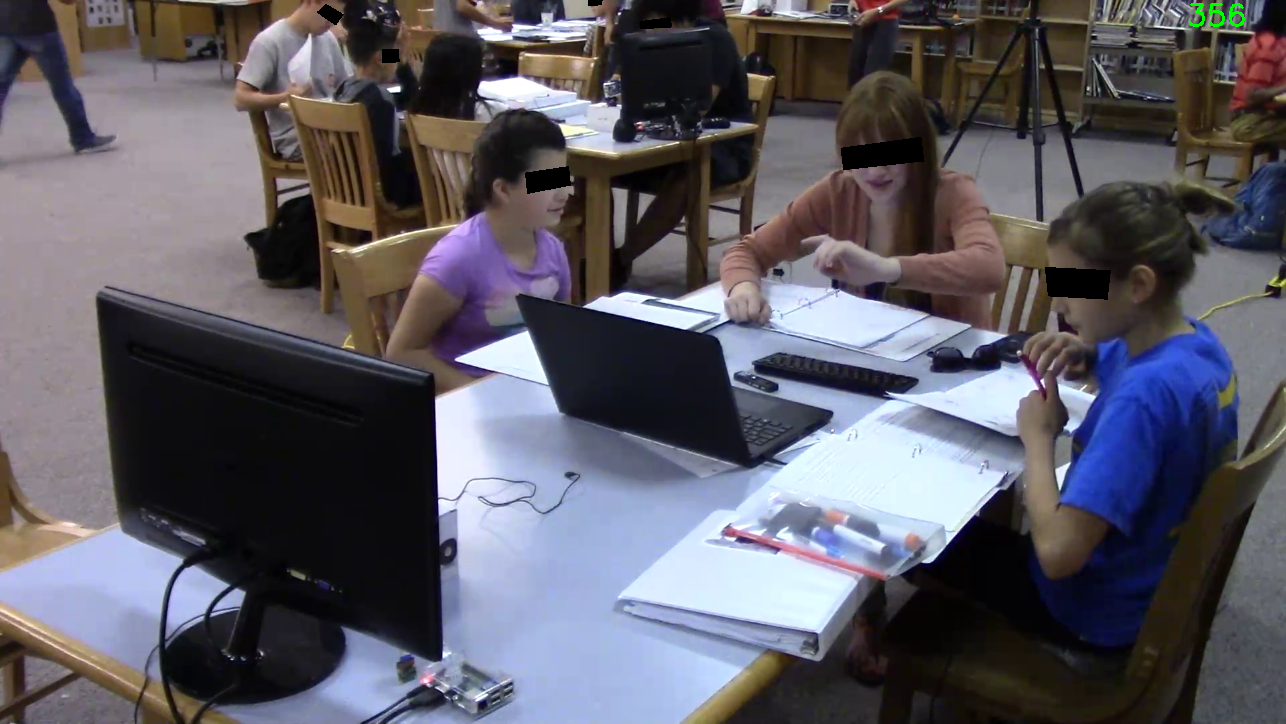}
    \caption{}
    \label{fig:10}
  \end{subfigure}\hfill
  \begin{subfigure}{0.3\columnwidth}
    \centering
    \includegraphics[width=\textwidth]{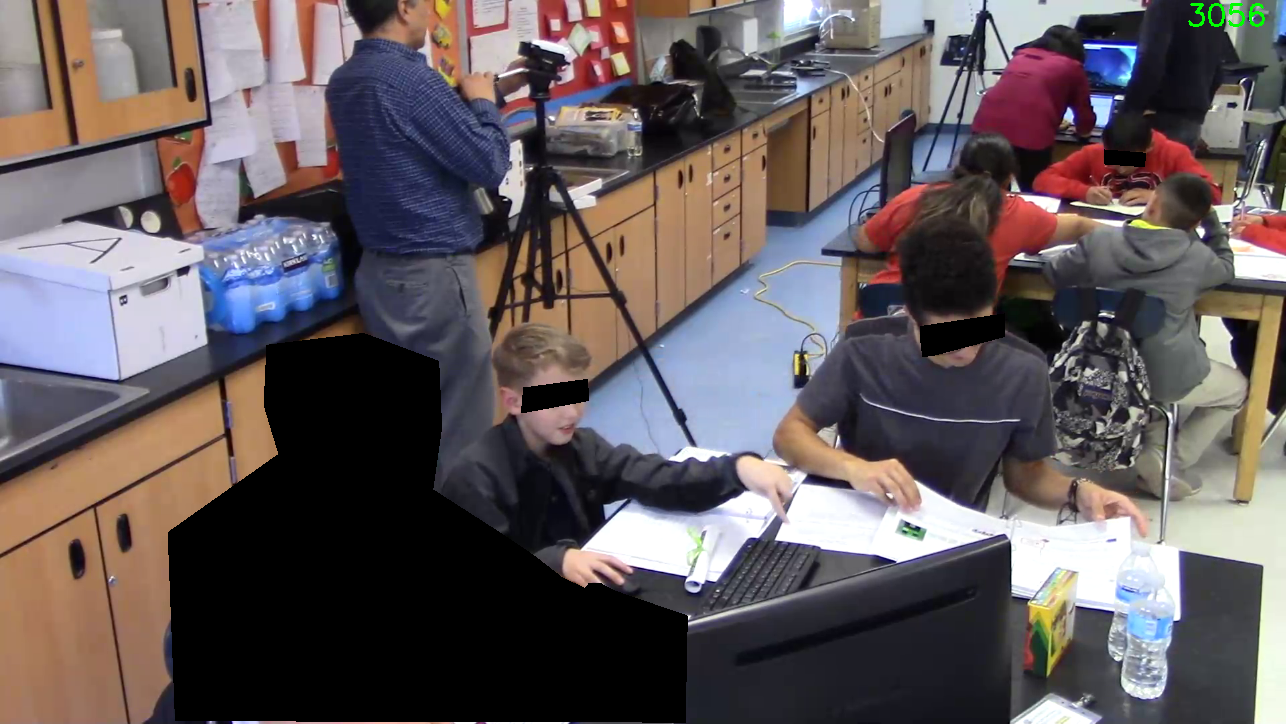}
    \caption{}
    \label{fig:11}
  \end{subfigure}\hfill
  \begin{subfigure}{0.3\columnwidth}
    \centering
    \includegraphics[width=\textwidth]{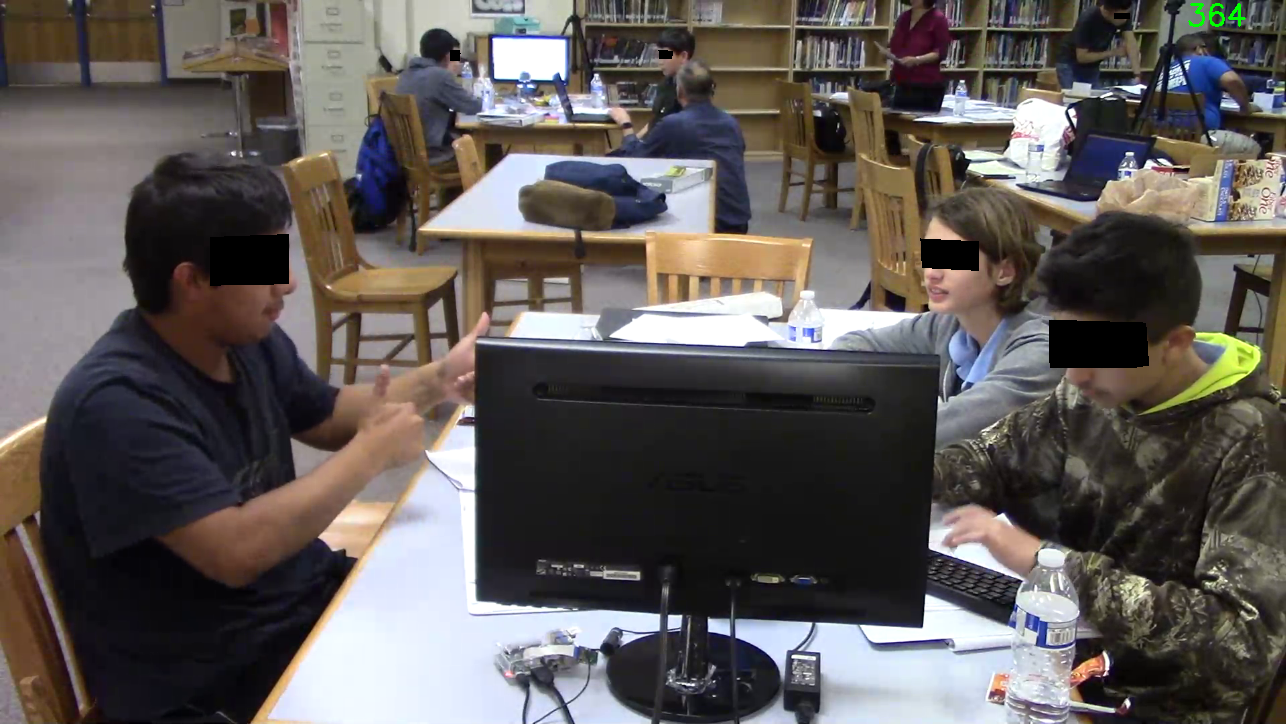}
    \caption{}
    \label{fig:12}
  \end{subfigure}

  \begin{subfigure}{0.3\columnwidth}
    \centering
    \includegraphics[width=\textwidth]{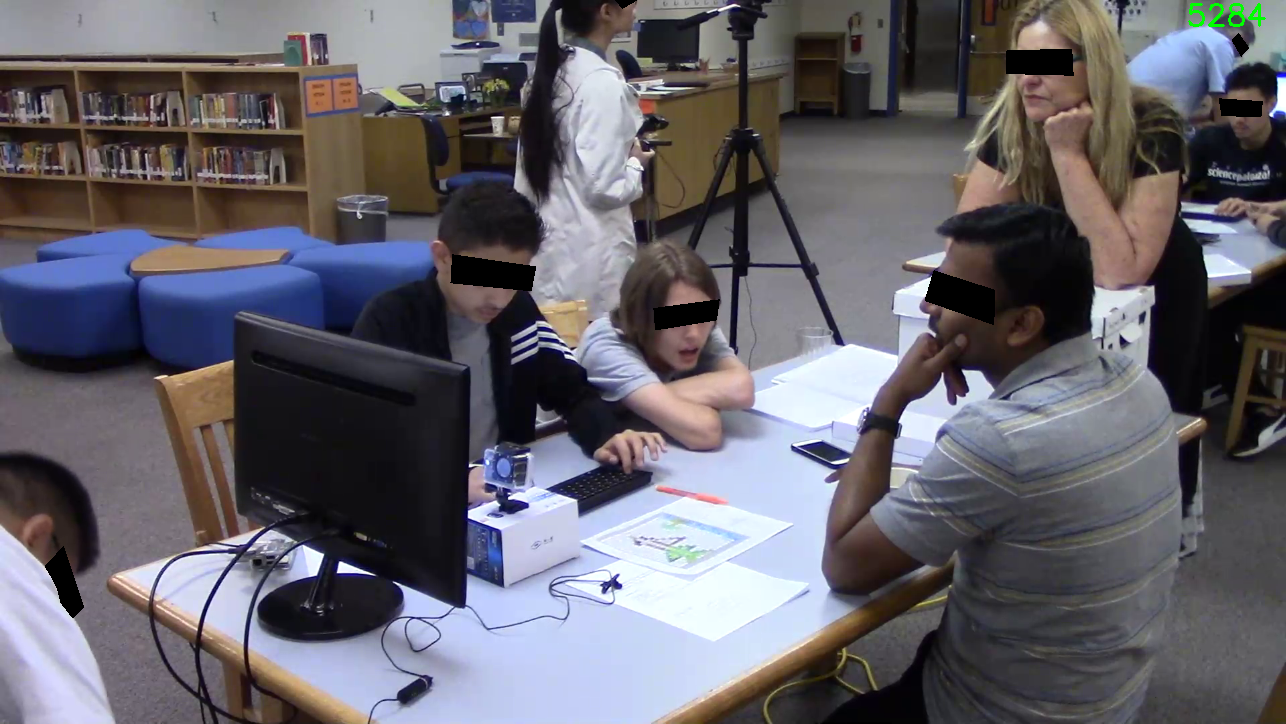}
    \caption{}
    \label{fig:13}
  \end{subfigure}\hfill
  \begin{subfigure}{0.3\columnwidth}
    \centering
    \includegraphics[width=\textwidth]{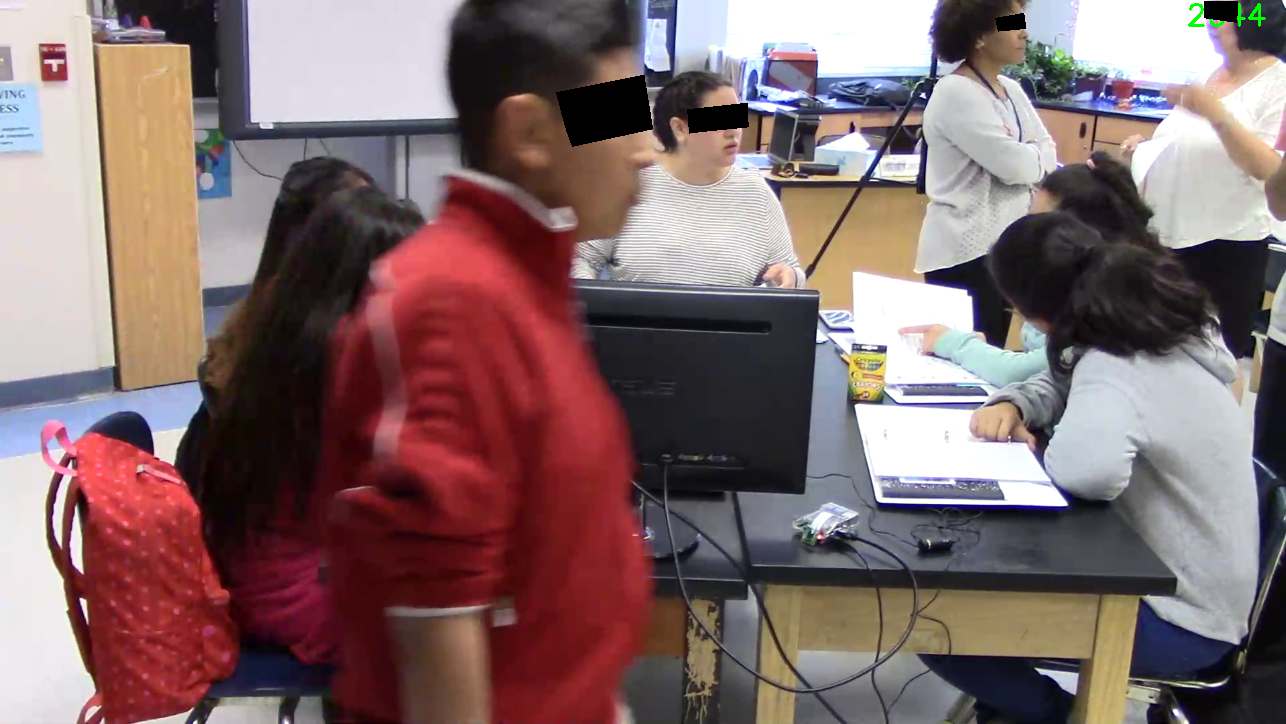}
    \caption{}
    \label{fig:14}
  \end{subfigure}\hfill
  \begin{subfigure}{0.3\columnwidth}
    \centering
    \includegraphics[width=\textwidth]{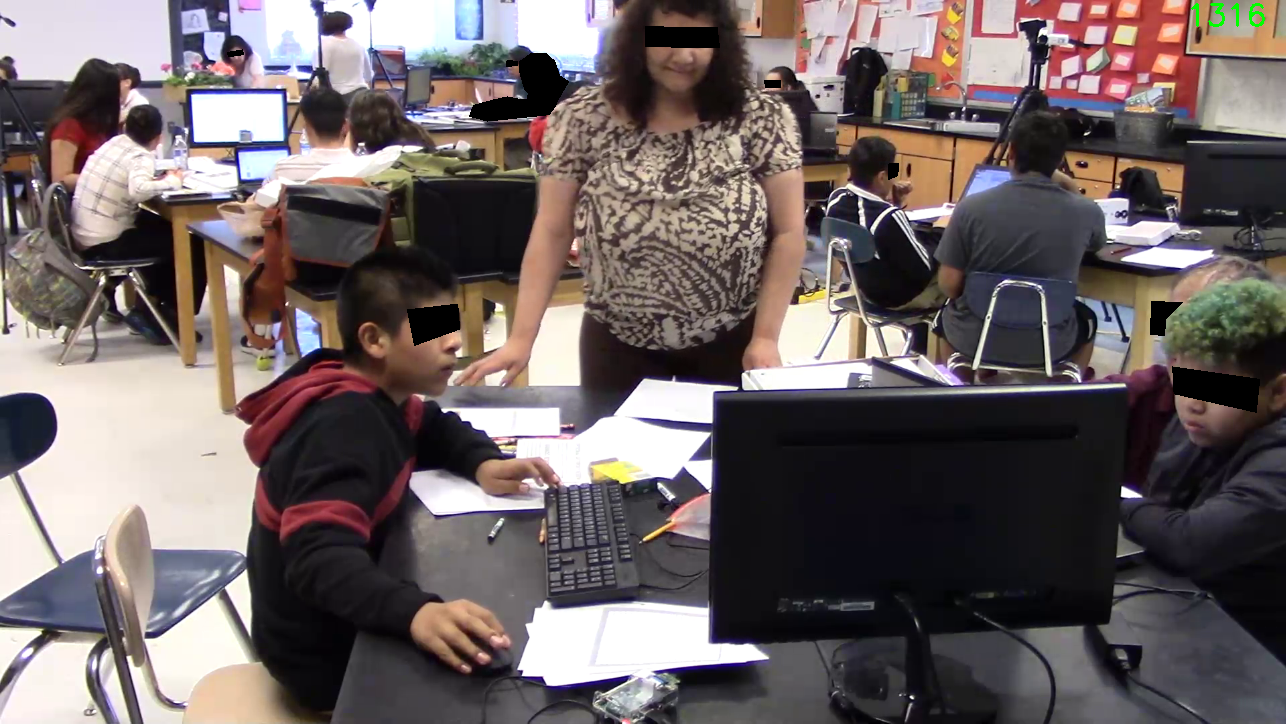}
    \caption{}
    \label{fig:15}
  \end{subfigure} 
  \caption{Frames from different videos showing challenges involved.} 
  \label{fig:dataset challenges}
\end{figure}

We illustrate some of the problems with the dataset in Fig. \ref{fig:dataset challenges}. We have (i) different camera angles in Fig. \ref{fig:3}, \ref{fig:2} and \ref{fig:14} (ii) inconsistent illumination in Fig. \ref{fig:1} and \ref{fig:4} (iii) monitor occluding activity regions in Fig. \ref{fig:12} and \ref{fig:15} (iv) person blocking other people in Fig. \ref{fig:14} (v) activity not associated with primary table of focus in Fig. \ref{fig:12} and \ref{fig:3} (vi) other groups working in background in Fig. \ref{fig:12} and \ref{fig:3} (vii) people walking around in Fig. \ref{fig:14} and \ref{fig:6}. All the problems described above will have an impact on detection results. So there is strong interest in developing methods to analyze videos that can address the challenges of these collaborative learning environments.

\section{Motivation}
A significant focus of the AOLME project is to understand how students learn. Therefore, how students interact with each other, the facilitator and their lessons are of importance.  To illustrate the problem, students learn and interact using their hands by typing, writing and pointing to things. For example, Figs. \ref{fig:4}, \ref{fig:7}, \ref{fig:13}, and \ref{fig:15} show students using keyboard;  Figs. \ref{fig:1} - \ref{fig:2} show writing, Figs. \ref{fig:4}, \ref{fig:7}, \ref{fig:11}, \ref{fig:13} show typing, Figs. \ref{fig:15} shows students using mouse, Figs. \ref{fig:4}, \ref{fig:4}, \ref{fig:8}, \ref{fig:9}, and \ref{fig:10}  show interaction using hands, Figs. \ref{fig:6} and \ref{fig:9} show gestures using hands.

The primary motivation of this thesis is to develop a robust method to detect and track objects in long videos and propose segmented hand regions in videos eliminating the background regions. This thesis will focus on detecting and tracking keyboards and hands in long videos captured in the collaborative learning environment. The thesis also proposes to use characteristics of video to provide a fast and robust system. In the future, the proposed method can easily integrate into an activity detection system providing spatio-temporal instances of typing and writing.

\section{Thesis Statement}
My thesis is that I can develop fast and effective object detection and tracking of hands and keyboards in long videos in collaborative learning environments through the integration of clustering, projections, tracking, and current object detection methods. 

\section{Contributions}
The contributions of this thesis include:
\begin{enumerate}
\item Robust detection and fast-tracking of the keyboard in long videos ($45$ minutes to 90 minutes).
\item Data augmentation study to determine optimal parameters to improve hand detection.
\item A novel method that uses projections and cluster based segmentation to detect hand regions in a video.
\end{enumerate}

\section{\label{section:overview}Overview}
The remainder of the thesis is organized into 5 chapters:
\begin{itemize}
\item \textbf{Chapter 2: Background.} This chapter describes prior work.
\item \textbf{Chapter 3: Dataset.} This chapter describes Dataset Organization and Ground Truth.
\item \textbf{Chapter 3: Methods.} This chapter describes methods for keyboard detection, tracking, hand detection, and proposal regions.
\item \textbf{Chapter 4: Results.} This chapter provides a summary of results for keyboard detection, tracking, hand detection, and proposal regions.
\item \textbf{Chapter 5: Conclusion and future work.} This chapter provides a summary of the thesis and recommendations for future work.
\end{itemize}

\chapter{Background}
This section provides a summary of prior research at ivPCL lab, object detection and tracking methods along with the common datasets used respectively.

\section{ Prior Work}
This thesis derives inspiration and expands on the prior research at the ivPCL lab. In this section, we provide a summary and contrast against the prior work. Table \ref{tab:ivpcl_prior_research} provides the summary.

The thesis extends prior research by developing fast and reliable methods for hand and keyboard detection that work on long videos. For keyboard detection, the thesis integrates fast tracking into the approach. Furthermore, the thesis integrates the use of projections every 12 seconds, cluster-based segmentation, and small area removal to detect hand regions and reject background images. The developed methods are tested on 7 long video sessions.
\newpage
\begin{table}[H]
  \caption{Computer assisted video analysis research related to video object detection and activity detection in the image and video processing and communications lab (ivpcl).}
  \label{tab:ivpcl_prior_research}
  \scalebox{0.79}
  {
  \renewcommand{\arraystretch}{1.5}
	     \begin{tabular}{P{2 cm} P{7.5 cm}  P{7.8 cm}}
	     \hline
	     \textbf{Author} & \textbf{Title} & \textbf{Summary}\\
	     \hline
	     \hline
	     \multicolumn{3}{c}{\textbf{Computer-Assisted Video Analysis Methods}}\\
	     
	     Ulloa, A., et, al. 2021\cite{cerna2021deep} & Deep-learning-assisted analysis of echocardiographic videos improves predictions of all-cause mortality & Analyzes echocardiographic videos to assist cardiologists in predicting one-year all-cause mortality. This framework method successfully increased the sensitivity of cardiologists by 13\%. \\

	     Tapia, L., et, al. 2020 \cite{tapia2020importance} & The Importance of the Instantaneous Phase for Face Detection using Simple Convolutional Neural Networks & Investigates the use of low-complexity image processing system to study  advantages of using AM-FM representations versus raw images. This framework showed significant advantages by reducing training time per epoch with low-complexity architecture at a comparable accuracy using FM images for training. \\
		
		 Esakki, G., et, al. 2020 \cite{esakki2020adaptive} & Adaptive 	Encoding for Constrained Video Delivery in HEVC, VP9, AV1 and VVC Compression Standards and Adaptation to Video Content &
		Provides optimal \texttt{QP} for compressing
		videos for high and acceptable video streaming. For further information please refer to \cite{esakki2020comparative, esakki2016optimal, esakki2017adaptive}. \\
		
		Kent, R. B., st, al. 2020 \cite{kent2020design} & Design,
		Implementation, and Analysis of High-Speed Single-Stage
		N-Sorters and N-Filters &  Provides an implementation in FPGA that can greatly speedup $2\times2$ and $3\times3$ max pooling.\\
		
		Carranza, C., et, al. 2020 \cite{carranza2020fast} & Fast and Scalable 2D Convolutions and Cross-correlations for Processing Image Databases and Videos on CPUs & Proposed a method which maximizes throughput through the use of vector-based memory I/O and optimized 2D FFT libraries that run on all available physical cores. Also shows decomposition of arbitrarily large image into small using overlap-and-add. This approach outperforms Tensorflow for 5×5 kernels and significantly outperforms Tensorflow for 11 × 11 kernels.\\

	     Darsey, C.J., 2018.\cite{darsey2018hand} & Hand Movement Detection in Collaborative Learning Environment Videos & Explores hand movement detection using color and optical flow. This approach used patch color classification, space-time patches of video, and histogram of optical flow. This approach achieved accuracy of 84\%  and ROC AUC of 89\% on video patches from 15 video clips. \\
	     
 	     Jacoby, A. R., et, al. 2017 \cite{jacoby2017context}, \cite{jacoby2018context} & Context-sensitive human activity classification in collaborative learning environments & Explores activity detection of writing, typing and talking. The method was tested on simulated data having 620 video frames for writing, 1050 for typing, and 1755 frames for talking. 	     	  

	     \end{tabular} 
     }
\end{table}

\begin{table}[hbt]
	\scalebox{0.79}
	{
	 \renewcommand{\arraystretch}{1.5}
		\begin{tabular}{P{2 cm} P{7.5 cm}  P{7.8 cm}} 
  	     Eilar, C. W., et, al. 2016 \cite{eilar2016distributed}, \cite{eilar2016distributed1} & Distributed video analysis for the Advancing Out-of-School Learning in Mathematics and Engineering project & Proposes  an open-source, maintainable system for detecting human activity in video datasets. \\

 		Shi, W., et, al. 2016 \cite{shi2018robust}, \cite{shi2016human}	& Robust head detection in collaborative learning environments using AM-FM representations & Focuses on head detection, attention based detection by classifying where faces look, and group interactions based on attention direction detected. This work uses texture by using AM-FM models. This work is still being continued by using  multiple  image representations  to  detect  people  in  specific  regions  of  long-term collaborative  learning  videos. \\
		 
 	     Jatla, V., et, al. 2016 \cite{jatla2016automatic, jatla2019image} & Image processing methods for coronal hole segmentation, matching, and map classification & Explored image processing models that can be used to detect coronal holes automatically. Here an  automated segmentation method has been
		 developed that improves significantly over state of the art models . \\ 
		 \hline
	     \end{tabular} 
}
\end{table}

\section{Object detection}
The goal of object detection methods is to determine where objects are located in an image and determine the category of the object. In this section, I will provide a summary of common object detection datasets and methods. The most commonly used object detection datasets are listed in the table \ref{tab:objdet-dataset}. The most popular object detection algorithms are as described in the table \ref{Common Object Detection Methods}.
\begin{table}[ht!]
  \label{tab:objdet-dataset}
  \caption{Summary of commonly used object detection datasets. These
      datasets contain a large number of classes with images extracted
      from multiple sources. MS-COCO \cite{lin2014microsoft} dataset
      contains a keyboard class, making it of particular interest for
      this research.}

  \scalebox{0.99}
  {
        \renewcommand{\arraystretch}{1.75}
  \begin{tabular}{ P{3cm} P{11cm}}
    \hline
    \textbf{Dataset} & \textbf{Summary} \\
    \hline
    \hline
 MS-COCO \cite{lin2014microsoft} &
    $\bullet$ 330k images($>200K$ labels) \newline
    $\bullet$ 80 object categories \newline
    $\bullet$ 5 captions per image \newline
    $\bullet$ 2,50,000 people with key points, Object segmentation. \newline
    $\bullet$ 1.5 million object instances\\
    
    ImageNet \cite{deng2009imagenet} & 
    $\bullet$ 1,500,000 images with multiple bounding boxes and respective class labels.  \newline 
    $\bullet$ Over 500 images per category     \\
    
    PASCAL VOC 2007 \cite{everingham2007pascal}& 
    $\bullet$ 20 classes, including person, animal, vehicle, and indoor.	\newline
    $\bullet$ Train/validation/test, 9,963 images containing 24,640 annotated objects.	    \\
    
    PASCAL VOC 2012 \cite{everingham2011pascal} & 
    $\bullet$ 20 classes. \newline 
    $\bullet$ The train/val data has 11,530 images containing 27,450 ROI annotated objects and 6,929 segmentations.	 \\
    \hline
  \end{tabular}
  }
\end{table}

\begin{table}[ht!]
  \caption{Summary of the most common object detection methods. These methods are supported officially by deep learning libraries, such as PyTorch, making them easily accessible.}
  \label{Common Object Detection Methods}
  \scalebox{0.79}
  {
        \renewcommand{\arraystretch}{1.75}
	     \begin{tabular}{P{2cm} P{2cm} P{4cm}  P{2.5cm} P{6cm}}
	       \hline
	       \textbf{Author} & \textbf{Method} & \textbf{Summary} & \textbf{Datasets} & \textbf{Results} \\
	       \hline 
	       \hline
	       Redmon,et al. 2018 \cite{redmon2016you}& YOLOv3 &
	       $\bullet$ Single neural network to full image. \linebreak
	        $\bullet$ Fast and accurate  & $\bullet$ PASCAL VOC \newline $\bullet$ MS COCO \cite{lin2014microsoft} & 
	       $\bullet$  YOLOv3 runs significantly faster than other detection methods with comparable performance and with high accuracy. \linebreak
	       $\bullet$ The mAP(mean Average Precision) is 57.9\% on COCO test-dev.\\
	       Shaoqing Ren, et al. 2017 \cite{ren2016faster}& Faster R-CNN & 
	       $\bullet$ End-to-end trained RPN to generate high quality region proposals, which are used by Fast R-CNN for detection  & $\bullet$ PASCAL VOC 2007 \newline $\bullet$ PASCAL VOC 2012 \newline $\bullet$ MSCOCO & 
	       $\bullet$ mAP of 73.2\% on PASCAL VOC 2007 \linebreak
	       $\bullet$ mAP of 70.4\% on PASCAL VOC 2012\\
	       Wei Liu, et al.2017 \cite{liu2016ssd}& SSD & 
	       $\bullet$ End-to-end CNN passing input image through series of convolutional layers, generating bounding boxes. \linebreak
	       $\bullet$ Works well with larger input image  & $\bullet$ PASCAL VOC \newline $\bullet$ MSCOCO  \newline $\bullet$ ILSVRC &
	       $\bullet$ mAP 74.3\% on VOC2007 test \linebreak
	       $\bullet$ On VOC2007 test, operated at 59 FPS with mAP 74.3\%, vs. Faster R-CNN 7 FPS with mAP 73.2\% or YOLO 45 FPS with mAP 63.4\%\\
	       \hline
	     \end{tabular} 
     }
\end{table}			

\begin{figure}	[]
  \includegraphics[width=\columnwidth]{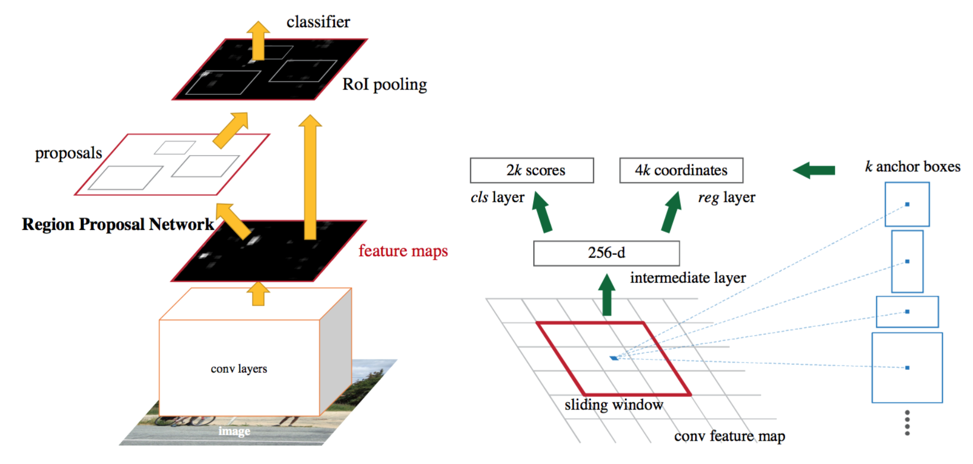}
  \caption{Architecture of Faster R-CNN \cite{ren2016faster}.}
  \label{fig:frcnn}
\end{figure}

\noindent\textbf{Faster R-CNN}\\
Faster R-CNN \cite{ren2016faster} is the modified version of Fast R-CNN. The major difference is that Fast R-CNN uses the selective search for generating Regions of Interest, while Faster R-CNN uses "Region Proposal Network," RPN. RPN takes image feature maps as an input that shares full-image convolutional features with the detection network, thus enabling nearly cost-free region proposals. RPN is a fully convolutional network that continuously generates object bounds and objectness scores at each position. This is trained end-to-end for high-quality region proposals which are further used by Fast R-CNN for detection. An RoI pooling layer is applied on these proposals to bring down these proposals to the same size. These proposals are passed through the FCN  layer, which has softmax and linear regressor at its top to classify and output bounding boxes.  The architecture is shown in Fig.  \ref{fig:frcnn}.

\begin{figure}	[]
  \includegraphics[width=\columnwidth]{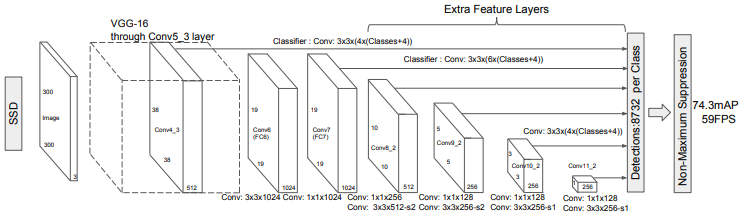}
  \caption{Architecture of SSD \cite{liu2016ssd}.}
  \label{fig:ssd}
\end{figure}

\noindent\textbf{Single Shot Detector} \\
Single Shot Detector (SSD) \cite{liu2016ssd} is a method for detecting objects in images using a single deep neural network. The SSD approach discretizes the output space of bounding boxes into a set of default boxes over different aspect ratios. After discretizing, the method scales per feature map location. The Single Shot Detector network combines predictions from multiple feature maps with different resolutions to naturally handle objects of various sizes. The architecture is shown in Fig. \ref{fig:ssd}.

\begin{figure}	[] 
  \includegraphics[width=\columnwidth]{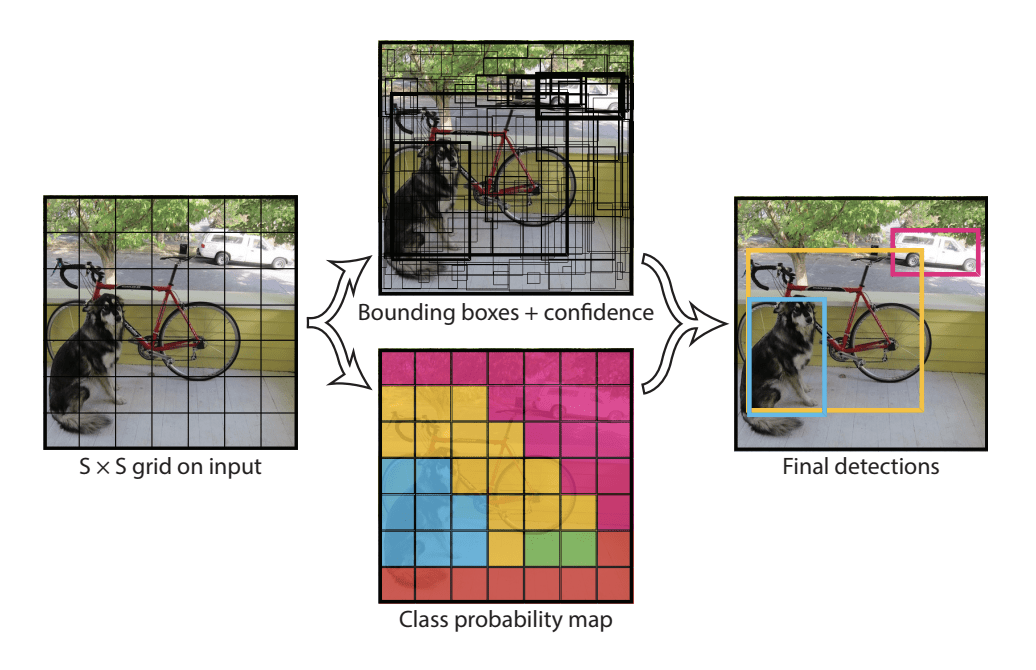}
  \caption{YOLO \cite{redmon2016you}.}
  \label{fig:yolo}
\end{figure}

\noindent\textbf{YOLO} \\
YOLO \cite{redmon2016you} shown in figure \ref{fig:yolo} uses a single neural network trained end-to-end that takes a photograph as input and predicts bounding boxes and class labels for each bounding box directly. This model works by first splitting the input image into a grid of cells, where each cell is responsible for predicting a bounding box if the center of a bounding box falls within the cell. Each grid cell predicts a bounding box involving the x, y coordinate, the width and height, and the confidence. A class prediction is also based on each cell.

\subsection{Object detection libraries}
This section briefly describes the frameworks used in
the thesis to implement object detection. There are multiple github
repositories that provide object detection. In this thesis, we used
the frameworks supported by well established companies in AI research,
such as Facebook \cite{wudetectron2} and OpenMMLab \cite{chen2019mmdetection}.

\noindent\textbf{Detectron2}\\
Detectron2 \cite{wudetectron2} is Facebook AI Research's software
system that implements state-of-the-art object detection algorithms.
Detectron2 is powered by Pytorch deep learning frameworks. It includes
features such as Deeplab, Cascade R-CNN, rotated bounding boxes, etc.
The training  is faster, and models can be exported to Torchscript or
Caffe format for deployment.

\noindent\textbf{MMDetection}\\
MMDetection \cite{chen2019mmdetection} is an open-source object
detection toolbox based on PyTorch. It is a part of the OpenMMLab
project. The toolbox stems from the codebase developed by the MMDet
team, who won the COCO Detection Challenge. The detection framework is divided into different
components so that one can easily use a customized object detection
framework by combining different modules. It supports multiple
state-of-the-art frameworks like Faster R-CNN, YOLO, SSD, Mask R-CNN
etc. Once the data is arranged in a particular COCO format, multiple
frameworks can be trained by just changing configuration files. All
the operations are run on GPUs, so the training speed is comparable
to other popular networks like Detectron2 and Tensorflow.

\section{Object tracking}
Object tracking methods aim to determine the spatial coordinates of an object (initialized previously) in future frames. This section summarizes common object tracking datasets and trackers supported in OpenCV and state-of-the-art algorithms for tracking. The most commonly used datasets for object tracking are listed in table \ref{Common Object Tracking Datasets}. OpenCV object tracking API was introduced in OpenCV 3.0. A total of 8 tracking algorithms are available in OpenCV. They are BOOSTING, MIL, KCF, TLD, MEDIANFLOW, GOTURN, MOSSE, and CSRT. Generally, tracking algorithms are faster than detection; the reason is that the algorithm already knows the object's appearance. All the tracking algorithms are summarized in table \ref{Object Trackers from OpenCV}. In addition, the state-of-the-art tracking algorithms are summarized in table \ref{Sota obj tracking algorithms}.
\begin{table}[h]
  \caption{{Summary of object tracking datasets used to train and test single-target and multi-target tracking.}}
  \label{Common Object Tracking Datasets}
  \resizebox{\textwidth}{!}{
      \renewcommand{\arraystretch}{1.75}
	     \begin{tabular}{ P{0.25\textwidth} P{0.75\textwidth} }
	       \hline
	       \textbf{Dataset} & \textbf{Summary} \\
	       \hline
	       \hline
	       MOT17 \cite{milan2016mot16}& 
	       $\bullet$ Mutli-target tracking. \newline
	       $\bullet$ 42 sequences with crowded scenarios, camera motions, and weather conditions. \\
	       
	       MOT20 \cite{dendorfer2020mot20} & 
	       $\bullet$ 8 new sequence depicting very crowded challenging scenes.  \\

	       TrackingNet \cite{muller2018trackingnet} &
	       $\bullet$ A Large-Scale Dataset and Benchmark for Object Tracking in the Wild. \newline	
	       $\bullet$ $>30K$ Video Sequences. \newline
	       $\bullet$ $>14M$ Bounding Boxes. \newline
	       $\bullet$ Diversity ensured by YouTube  \\

	       VOT2018 \cite{kristan2018sixth}&
	       $\bullet$ 235 sequences, carefully selected to obtain a dataset with long sequences containing many target disappearances from LTB35. \newline
	       $\bullet$ Twenty sequences were obtained from the UAVL20 \cite{mueller2016benchmark}, six sequences were taken from Youtube, six sequences were generated from the omnidirectional view generator AMP [96] to ensure many target disappearances.\newline
	       $\bullet$ Sequence resolutions range between 1280×720 and 290×217. The dataset contains 14687 frames, with 433 target disappearances. \newline
	       $\bullet$ Each sequence contains on average 12 long-term target disappearances, each lasting on average 40 frames.\\
	       \hline
	     \end{tabular} }
\end{table}

\noindent
\textbf{Boosting Tracker \cite{grabner2006real}:} \\
This tracker is based on AdaBoost's online edition. There is no special reason to use these trackers since there are more advanced trackers such as KCF, MIL. The tracking performance is mediocre and does not reliably track failure. It is not real time. It fails with random and fast movements. \\
\textbf{MIL Tracker \cite{babenko2009visual}:}\\ 
The perfomance is better compared to BOOSTING, since it does not drift much. It does good job under partial occlusion. It fails with random and fast movements. It is not real time.\\
\textbf{KCF Tracker \cite{henriques2014high}:} \\
Accuracy and speed are both better than previous two trackers. It reports tracking failure. It does not recover from full occlusion and fast movements. It is real time. This is the best peforming algorithm that was used in this Thesis. \\
\textbf{TLD Tracker \cite{kalal2011tracking}:} \\
This tracker decomposes the task into (short term) tracking, learning, and detection. The tracker follows the object from frame to frame. The detector localizes all appearances that have been observed so far and corrects the tracker if necessary. It works under occlusion. It is not real time. It fails with fast and random movements. \\
\textbf{MEDIANFLOW Tracker \cite{kalal2010forward}:} \\
This tracker tracks the object both in forward and backward direction. It reliably reports tracking failure. It works well when there is no occlusion. It fails under large motions. It  is real time.  \\
\textbf{GOTURN Tracker \cite{held2016learning}:} \\
Out of all the trackers, this is the only one based on Convolutional Neural Networks. It does track pretty well when the object is in training set. The tracker has a hard time tracking part of an object. It is not real time. \\
\textbf{MOSSE Tracker \cite{bolme2010visual}:} \\
It is the fastest of all the available trackers. It is easy to implement. It fails with random movements. But due to its speed, on a performance scale it lags behind. \\ 
\textbf{CSRT Tracker \cite{lukezic2017discriminative}:} \\
It  uses the spatial reliability map for adjusting the filter support to the part of the selected region from the frame for tracking. It is not real time.

\begin{table}[!h]
\centering
  \caption{{Summary of object tracking methods available in OpenCV. These
  are typically fast methods that do not use deep learning.}}
  \label{Object Trackers from OpenCV}
    \resizebox{\textwidth}{!}{
    \renewcommand{\arraystretch}{1.75}
	     \begin{tabular}{ P{0.25\textwidth} P{0.75\textwidth}  }
	       \hline
	       \setlength{\extrarowheight}{20pt}
	       \textbf{Method} & \textbf{Summary} \\
	       \hline
	       \hline
	       Boosting\cite{grabner2006real} & 
	           \tabitem Old and Performance is mediocre \newline
	           \tabitem Does not know when tracking is failed.\\
	       
	       Multiple Instance Learning \cite{babenko2009visual} & 
	       \tabitem Performance is pretty good, performs well under partial occlusion. \newline
	       \tabitem Tracking failure is not reported reliably. \newline
	       \tabitem Does not recover from full occlusion. \\
	       
	       Kernelized Correlation Filters \cite{henriques2014high}& 
	        \tabitem Accuracy and Speed are both better than MIL. \newline
	        \tabitem Reports tracking Failure better than MIL and BOOSTING. \newline
	        \tabitem Does not recover from full occlusion \\

	       Tracking, Learning and Detection \cite{kalal2011tracking}&
	        \tabitem Best under occlusion over multiple frames and over scale changes. \newline
	        \tabitem Lots of false position.			\\

	       Median Flow \cite{kalal2010forward} & 
	        \tabitem Excellent tracking failure reporting. Works very well when the motion is predictable and there is no occlusion. \newline
	        \tabitem Fails under large motion.			\\

	       GO TURN \cite{held2016learning}&
	        \tabitem Fails on tracking objects that are not in the training set. \newline
	        \tabitem Hard time tracking part of the object.\\

	       MOSSE \cite{bolme2010visual} &
	        \tabitem Operates at higher fps (450 and even more). \newline
	        \tabitem Easy to implement and accurate\\

	      DCF-CSR \cite{lukezic2017discriminative}& 
		  \tabitem Operates at 25fps  \newline
		  \tabitem Gives High Accuracy\\
		  \hline
	     \end{tabular} 
     }
\end{table}

\begin{table*}[!h]
  \caption{{Summary of State-of-the-Art Object tracking algorithms.}}
  \label{Sota obj tracking algorithms}
      \resizebox{\textwidth}{!}{
      \renewcommand{\arraystretch}{1.75}
	     \begin{tabular}{P{2cm} P{3cm} P{9cm}P{3cm}}
	       \hline
	       \textbf{Author} & \textbf{Method} & \textbf{Summary} & \textbf{Datasets} \\
	       \hline 
   	       \hline
	       Qiang Wang,et al. 2019 \cite{wang2019fast}& Fast Online Object Tracking and Segmentation, A Unifying Approach
	       SiamMask & 
	       $\bullet$ 3rd best model for Visual Object Tracking on YouTube-VOS \newline
	       $\bullet$ Single Object Tracking \newline
	       $\bullet$ Produces segmentation masks and bounding boxes at 55fps \newline
	       $\bullet$ Real time and fastest \newline& VOT-2016, VOT-2018, DAVIS-2016, DAVIS-2017 \\
	       
	       Yifu Zhang , et al. 2020 \cite{zhang2020fairmot}& A Simple Baseline for Multi-Object Tracking & 
	       $\bullet$ SOTA for Multi-Object Tracking on MOT16 \newline
	       $\bullet$ First among all online trackers \newline
	       $\bullet$ Operates at 30fps  \newline
	       $\bullet$ Anchor-free object detection to reduce ambiguity, parallel branch for Re-ID features & 2DMOT15, MOT16, MOT17, MOT20\\
	       \hline
	     \end{tabular}
     }
\end{table*}

\section{Uniqueness of AOLME Dataset} 
AOLME is different from other typical datasets in the way that it primarily includes long videos which are over an hour. Each video has multiple activities but not limited to typing, talking, eating, and writing. Many challenges impact the results like occlusion, multiple camera angles, illumination issues, multiple people performing the same activity, fast and random movements, people moving across the videos, and activities in the background. 

\chapter{Dataset}
The Advancing Out-of-school Learning in Mathematics and Engineering
(AOLME) project is an after-school program collaboratively implemented
by the Department of Electrical and Computer Engineering and
the Department of Language, Literacy, and Sociocultural Studies.
AOLME generated a large amount of multimedia data that includes around 2200 hours of group
interactions, monitor data, and screen recordings.

A group in the AOLME setting has three to five students, one facilitator, and a co-facilitator. The interactions between the group were recorded using a video camera. The videos have
$1920\times1080$ resolution and $30$ or $60$ frames per second.

In this chapter, we give a detailed description of AOLME data
terminology and organization. Following this, we describe the ground truth generation to train object detectors for detecting hands and
keyboards in group interactions.

\section{Organization of Dataset}
\begin{figure}[!b]
  \centering
  \includegraphics[width=\textwidth]{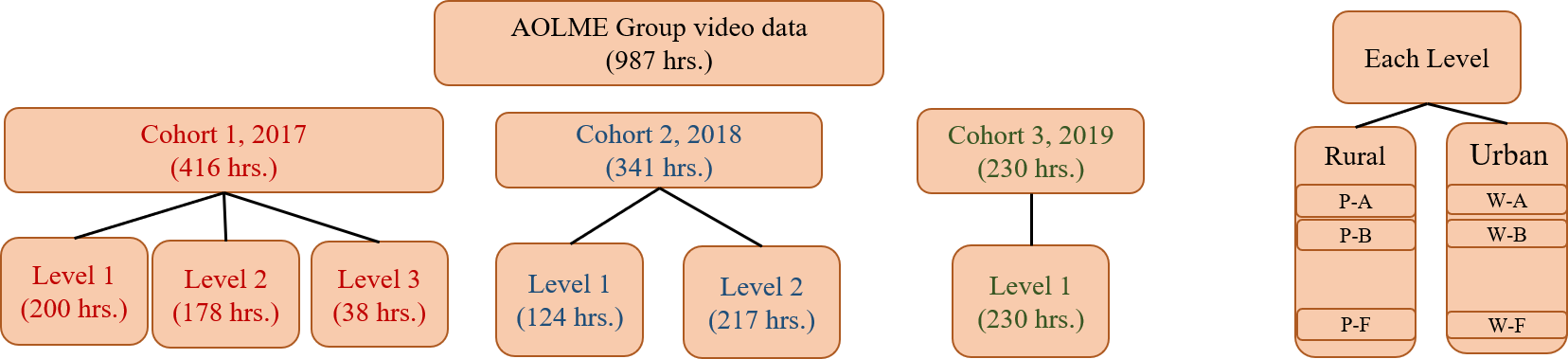}
  \caption{AOLME Dataset Organization.}
  \label{Dataset Organization}
\end{figure}

Fig. \ref{Dataset Organization} shows the organization of the AOLME Dataset. Group video data is collected across three years which are named Cohorts. Each Cohort is again organized into levels based on Spring, Summer, and Fall. Each Level again has two schools: Rural and Urban.

For the dataset, we have:
\begin{itemize}
\item Each school has 3 to 7 groups.
\item Each group has 3 to 5 members.
\item Typically, per level a group does 10 to 12 sessions.
\item A session lasts anywhere between 45 minutes to 90 minutes.

The videos are organized using the following naming convention:
  \begin{itemize} 
  \item \textbf{C1L1P-A}  : Cohort 1, Level 1, Rural, Group A (Includes all sessions from Group)
  \item \textbf{C1L1P-A, Mar02} : Cohort 1, Level 1, Rural, Group A (Single Session)
  \end{itemize}
\end{itemize}

\section{Ground Truth}
\subsection{Generation of Ground Truth}
To generate ground truth, MATLAB Video Labeler 2018Rb was used. Each video was reviewed in three second segments. Within each segment, a spatiotemporal bounding box was used to mark each activity. The list of activities included: typing/no-typing and writing/no-writing. For each activitiy, we also stored the anonymized student, the start time, and activity duration.
\begin{figure}[!h]
  \includegraphics[width=\columnwidth]{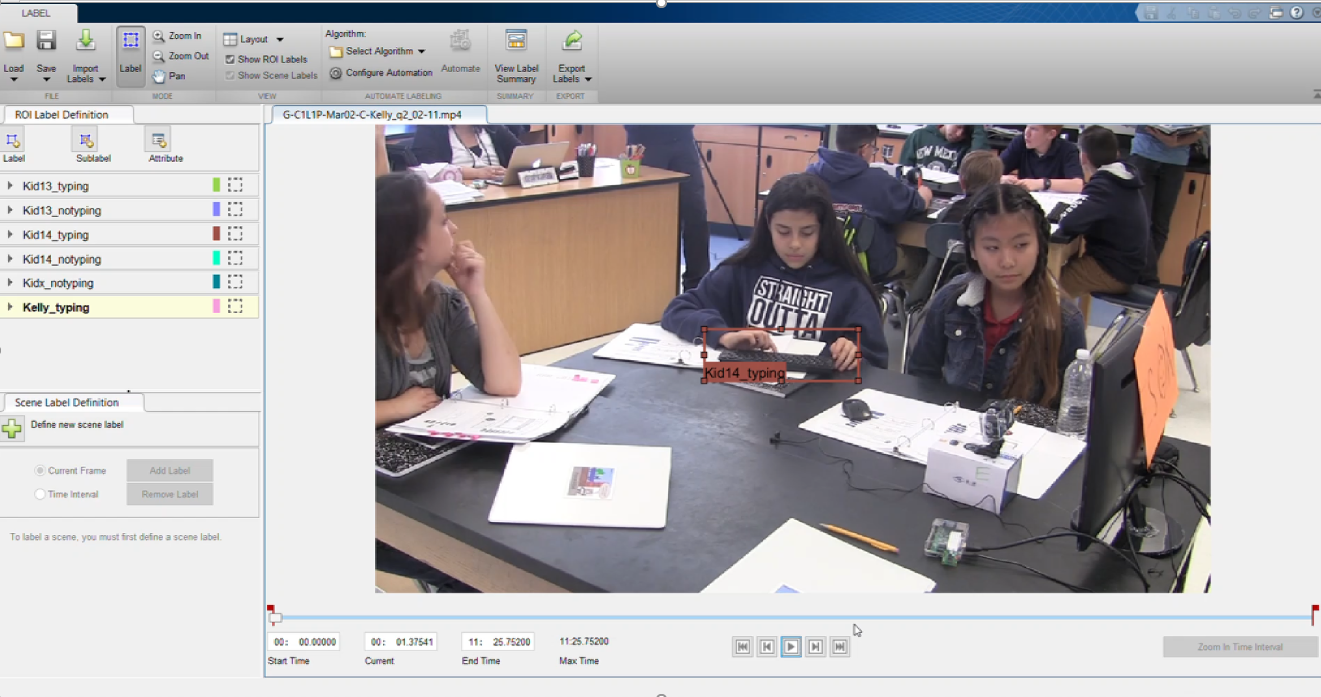}
  \caption{Ground truth labeling process.}
  \label{fig:}
\end{figure}
\begin{figure}[!h]
  \includegraphics[width=\columnwidth]{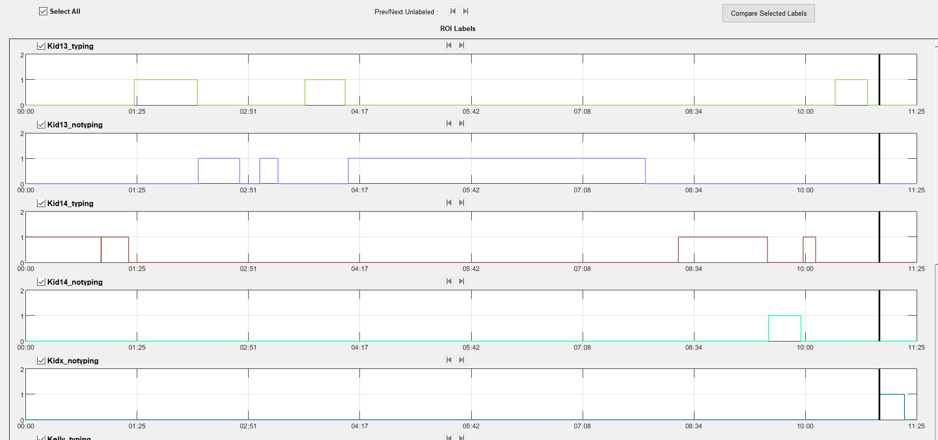}
  \caption{Visualizing ground truth using video labeler tool.}
  \label{fig:gt}
\end{figure}

The ground truth datasets are summarized in Tables \ref{tab:typing-dataset} and \ref{tab:writing-dataset}. A total of 83 hours and 49 hours of video data was analyzed for labeling typing and writing activities, respectively.
\begin{flushleft}
  \begin{table}[!h]
    \caption{
      { Typing and no typing ground truth on AOLME group videos. These videos
	are of 848x480 resolution. Playback time typically ranges from 11 to 16 minutes at 30 or 60 FPS. We use boldface to identify video sessions where ground truth was provided over the entire session. }
    }
    \label{tab:typing-dataset}
    \scalebox{0.63}
	     {
	       \begin{tabular}{l l l l l l l l}
		 \multicolumn{8}{c}{\textbf{Cohort 1, Level 1}}\\
		 \multicolumn{8}{c}{\textbf{Rural}}\\
		 Group & Dates & \# videos proc. & Total Videos & Hours proc. & Total hours & Typing & No Typing \\
		 A   & Feb 16, Feb 25, Mar 02, Mar 09 & 7 & 59 & 1.65 & 17.79 & 0.26 & 0.63\\
		 & Apr 06, Apr 13, Apr 20\\
		 B   & Mar 02, Mar 09, Mar 30, Apr 06 & 17 & 61 & 3.69 & 14.13 & 0.45 & 1.03 \\
		 & Apr 27, May 04, May 06, May 11\\
		 C   & Feb 16, Feb 25, Mar 02, Mar 09 & 35 & 68 & 8.13 & 14.05 & 1.24 & 1.32 \\
		 & \textbf{Mar 30, Apr 13}, Apr 20, \textbf{May 04}\\
		 D   & Mar 09, Apr 06, Apr 13, Apr 20 & 9 & 14 & 3.77 & 6.42 & 0.5 & 0.04\\
		 E   & Feb 25, \textbf{Mar 02}	     & 9 & 23 & 1.61 & 4.01	& 0.4 & 0.67\\
		 ~\\
		 \multicolumn{8}{c}{\textbf{Urban}}\\
		 A   & Feb 21, Feb 28, Mar 07, Mar 28 & 6 & 36 & 1.3 & 7.05 & 0.4 & 0.45\\
		 & Apr 25\\
		 B   & May 06 & 1 & 14 & 0.27 & 2.89 & 0.11 & 0.03\\
		 C   & Feb21  & 1 & 9 & 0.19 & 1.38 & 0.04 & 0.11\\
		 D   & Feb28  & 1 & 8 & 0.15 & 1.48 & 0.05 & 0.03\\
		 ~\\
		 \hline
		 \textbf{Total}   & &35 & 292 & 20.76 &69.2 &3.45 & 4.31\\
		 \hline
		 \\
		 \\
		 \multicolumn{8}{c}{\textbf{Cohort 2, Level 1}}\\
		 \multicolumn{8}{c}{\textbf{Rural}}\\

		 Group & Dates & \# videos proc. & Total Videos & Hours proc. & Total hours & Typing & No Typing \\
		 B & \textbf{Feb 23} & 6 & 6 & 1.81 & 1.80 & 0.21 & 1.37\\
		 C & \textbf{Apr 12} & 6 & 6 & 1.94 & 1.94 & 0.01 & 1.18\\
		 D & \textbf{Mar 08} & 5 & 5 & 1.78 & 1.78 & 0.31 & 1.08 \\
		 ~\\
		 \multicolumn{8}{c}{\textbf{Urban}}\\
		 A & Apr 10 & 2 & 5 & 0.79 & 1.54 & 0.02 & 0.0\\
		 B & \textbf{Feb 27} & 4 & 4 & 1.39 & 1.39 & 0.41 & 0.83\\
		 ~\\
		 \hline
		 \textbf{Total}          &     &23 & 26 & 7.7 &8.44 &0.959 & 3.45\\
		 \hline
		 \\
		 \\
		 \multicolumn{8}{c}{\textbf{Cohort 3, Level 1}}\\
		 \multicolumn{8}{c}{\textbf{Rural}}\\
		 Group & Dates & \# videos proc. & Total Videos & Hours proc. & Total hours & Typing & No Typing \\
		 C & \textbf{Apr 11} & 5 & 5 & 1.78 & 1.78 & 0.31 & 1.08\\
		 D & \textbf{Feb 21} & 5 & 5 & 1.77 & 1.77 & 0.15 & 1.08 \\
		 ~\\
		 \multicolumn{8}{c}{\textbf{Urban}}\\
		 A & \textbf{Mar 19} & 4 & 4 & 1.35 & 1.35 & 0.18 & 1.11\\
		 ~\\
		 \hline
		 \textbf{Total}          &     &14 & 14 & 4.9 &4.9 &0.64 & 3.34\\
		 \hline
	       \end{tabular}
	     }
  \end{table}
\end{flushleft}

\begin{table}[!h]
  \caption{
    {Writing and no writing ground truth on AOLME group videos. These videos
      are of 848x480 resolution. Playback time typically ranges from 11 to 16 minutes
      at 30 or 60 FPS. We use boldface to identify video sessions where ground truth was provided over the entire session.}
  }
  
  \label{tab:writing-dataset}
  \scalebox{0.61}
	   {
	     \begin{tabular}{l l l l l l l l}
	       \multicolumn{8}{c}{\textbf{Cohort 1, Level 1}}\\
	       \multicolumn{8}{c}{\textbf{Rural}}\\
	       Group & Dates & \# videos proc. & Total Videos & Hours proc. & Total hours & Writing & No Writing \\
	       B & \textbf{Mar 02} & 9 & 9 & 1.5 & 1.5 & 0.5 & 2.85\\
	       C   & Feb 16, Feb 25, Mar 09, \textbf{Mar30} & 38 & 69 & 8.86 & 15.07 & 2.22 & 7.69\\
	       & Apr 06, \textbf{Apr 13}, Apr 20,May04, May11\\
	       D   & Mar 02, Mar09, Mar30, Apr 06, Apr 13 & 18 & 29 & 7.03 & 9.45 & 1.08 & 0\\
	       &  Apr 20\\
	       E & \textbf{Mar 02} & 8 & 8 & 1.42 & 1.42 & 0.87 & 3.61\\
	       ~\\
	       \multicolumn{8}{c}{\textbf{Urban}}\\
	       Group & Dates & \# videos proc. & Total Videos & Hours proc. & Total hours & Writing & No Writing \\
	       A   & Feb 14, Feb 21, Feb 28, Apr 04 & 13 & 30 & 2.33 & 5.49 & 0.56 & 0\\
	       ~\\
	       \hline
	       \textbf{Total}          &        &86 & 149 & 21.12 & 32.93 & 5.23 & 14.15\\
	       \hline
	       \\
	       \\
	       \multicolumn{8}{c}{\textbf{Cohort 2, Level 1}}\\
	       \multicolumn{8}{c}{\textbf{Rural}}\\
	       Group & Dates & \# videos proc. & Total Videos & Hours proc. & Total hours & Writing & No Writing \\
	       B & \textbf{Feb 23} & 6 & 6 & 1.80 & 1.80 & 0.15 & 2.83\\
	       C & \textbf{Apr 12} & 6 & 6 & 1.95 & 1.95 & 0.97 & 1.23\\
	       D & \textbf{Mar 08} & 5 & 5 & 1.78 & 1.78 & 0.88 & 0.01\\
	       E & \textbf{Apr 12} & 6 & 6 & 1.85 & 1.85 & 0.33 & 2.16\\
	       
	       \multicolumn{8}{c}{\textbf{Urban}}\\
	       A & Feb 20, Apr 10 & 3 & 9 & 1.09 & 2.57 & 0.34 & 0.0\\
	       ~\\
	       \hline
	       \textbf{Total}          &     &26 & 32 & 8.47 &9.95 &2.67 & 6.23\\
	       \hline
	       \\
	       \\
	       \multicolumn{8}{c}{\textbf{Cohort 3, Level 1}}\\
	       \multicolumn{8}{c}{\textbf{Rural}}\\
	       Group & Dates & \# videos proc. & Total Videos & Hours proc. & Total hours & Writing & No Writing \\
	       C & \textbf{Apr 11} & 5 & 5 & 1.78 & 1.78 & 0.84 & 2.74\\
	       D & Feb14, \textbf{Feb21} & 6 & 9 & 1.78 & 2.94 & 0.14 & 1.15\\
	       \multicolumn{8}{c}{\textbf{Urban}}\\
	       D & \textbf{Mar19} & 4 & 4 & 1.35 & 1.35 & 0.16 & 0.19\\
	       \hline
	       \textbf{Total}          &        &15 & 18 & 4.91 & 6.07 &1.14 & 4.08\\
	       \hline
	     \end{tabular}
	   }
\end{table}

\subsection{Testing Dataset} 
All the sessions handpicked by the College of Education are used for testing.  These include 13 different sessions from Urban and Rural schools, which include 37 students and 10 facilitators. The total length of these videos is over 21 hours. The dataset is summarized in Table \ref{Testing Dataset}.
\begin{table}[H]
  \centering
  \caption{{AOLME Testing dataset.}}
  \label{Testing Dataset}
  \begin{tabular}{ |l | l| l| l | p{3cm}|}
    \hline
    Cohort & Level & Date & Group & School \\
    \hline 
    1 & 1 & Mar-02 & B & Rural \\
    1 & 1 & Mar-30 & C & Rural \\
    1 & 1 & Apr-06 & C & Rural \\
    1 & 1 & Apr-13 & C & Rural \\
    1 & 1 & Mar-02 & E & Rural \\
    2 & 1 & Feb-23 & B & Rural \\
    2 & 1 & Apr-12 & C & Rural \\
    2 & 1 & Mar-08 & D & Rural \\
    2 & 1 & Apr-12 & E & Rural \\
    2 & 1 & Feb-27 & B & Urban \\
    3 & 1 & Apr-11 & C & Rural \\
    3 & 1 & Feb-21 & D & Rural \\
    3 & 1 & Mar-19 & D & Urban \\
    
    \hline
  \end{tabular}	
\end{table}	

\begin{figure}[ht!]
  \centering
  \begin{subfigure}[t]{0.49\textwidth}
    \centering
    \includegraphics[height=1.2in]{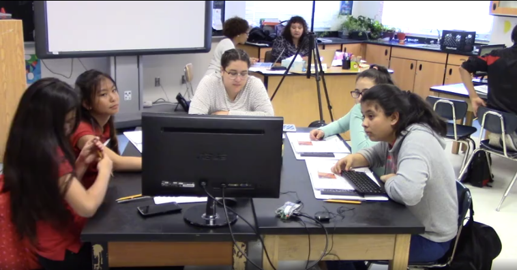}
    \caption{}
  \end{subfigure}%
  ~ 
  \begin{subfigure}[t]{0.49\textwidth}
    \centering
    \includegraphics[height=1.2in]{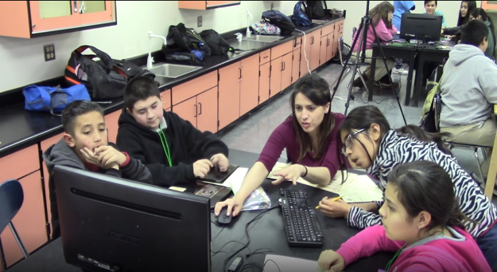}
    \caption{}
  \end{subfigure}
  \begin{subfigure}[t]{0.49\textwidth}
    \centering
    \includegraphics[height=1.2in]{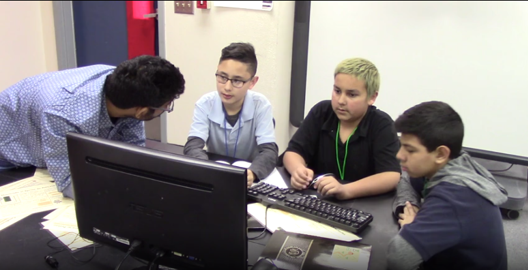}
    \caption{}
  \end{subfigure}%
  ~ 
  \begin{subfigure}[t]{0.49\textwidth}
    \centering
    \includegraphics[height=1.2in]{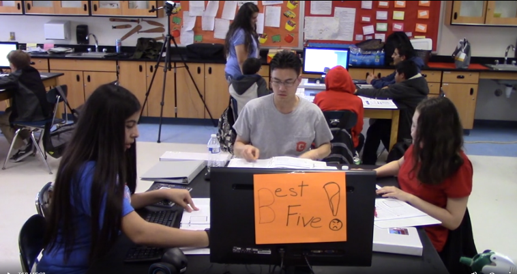}
    \caption{}
  \end{subfigure}
  \begin{subfigure}[t]{0.49\textwidth}
    \centering
    \includegraphics[height=1.2in]{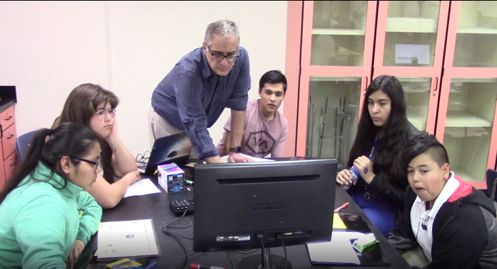}
    \caption{}
  \end{subfigure}%
  ~ 
  \begin{subfigure}[t]{0.49\textwidth}
    \centering
    \includegraphics[height=1.2in]{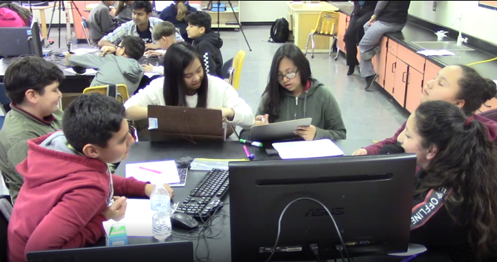}
    \caption{}
  \end{subfigure}
  \caption{Some of the images representing test set.}
\end{figure}

\clearpage
\section{Datasets for keyboard and hand detection}
\subsection{Dataset for keyboard detection}
Two Frames and corresponding bounding boxes are extracted
for every minute from typing-no typing ground truth
and exported as CSV files. Each image is of size 858$\times$480
pixels. Train and Test splits are done following the standard
of different sessions. The dataset used for keyboard detection is as shown in the table {Dataset for keyboard detection}.

\begin{table}[H]
	\caption{{Dataset for keyboard detection. }}
	\label{Dataset for keyboard detection}
	\centering
	\begin{tabular}{ |l | l| l| l |}
		\hline
		& \textbf{No. of Groups} & \textbf{No. of Sessions} & \textbf{No. of Images} \\
		\hline 
		\textbf{Train} & 9 & 33 & 700 \\
		\textbf{Validation} & 4 & 4 & 100\\
		\textbf{Test} & 6 & 7 & 648 \\
		\hline
	\end{tabular}
	
\end{table}

\subsection{Dataset for hand detection}
\begin{figure}	[ht!]
	\includegraphics[width=\columnwidth]{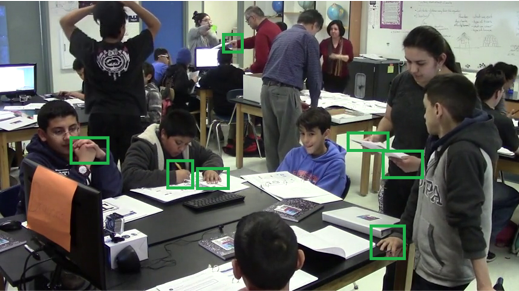}
	\caption{Sample from hand detection dataset.}
	\label{fig:hand_det_sample}
\end{figure}

Here as writing ground truth does not include all hands in the image, makesense.ai (Free open-source and Online labeling tool) \cite{make-sense} is used to generate ground truth. A total of 718 images were used here, with all the hands marked as shown in figure \ref{fig:hand_det_sample}. The dataset used for hand detection is shown in table \ref{Dataset for hand detection}.
\begin{table}[h!]
	\caption{{Dataset for hand detection. }}
	\label{Dataset for hand detection}
	\begin{tabular}{ |P{2.5cm } | P{2.5cm }| P{2.5cm }| P{2.5cm } | p{2.5cm }|}
		\hline
		& \textbf{No. of Groups} & \textbf{No. of Sessions} & \textbf{No. of Images} & \textbf{No. of Hand Instances} \\
		\hline 
		\textbf{Train} & 9 & 33 & 305 & 1803 \\
		\textbf{Validation} & 4 & 4 & 100 & 714 \\
		\textbf{Test} & 6 & 7 & 313 & 2031 \\
		\hline
	\end{tabular}
	
\end{table}	

\subsection{Testing Dataset}
Training, Validation, and Testing are mutually exclusive with respect to sessions. Throughout all the experiments, the dataset used for testing is from sessions listed in the table \ref{testing_datset}.
\begin{table}[hbt]
	\centering
	\caption{Dataset for Testing.}
	\label{testing_datset}
	\begin{tabular}{|l|l| }
		\hline
		\textbf{Group Video} & \textbf{Date} \\
		\hline
		C1L1P-C 	& Mar 30  5\\
		C1L1P-C		& Apr 13  5\\
		C1L1P-E	 	& Mar 02  \\
		C2L1P-B		& Feb 23\\
		C2L1P-D		& Mar 08  \\
		C3L1P-C 	& Apr 11 \\
		C3L1P-D 	& Mar 19  \\
		\hline
	\end{tabular}
\end{table}

\chapter{Methodology}
This chapter provides summary of methods for keyboard detection and tracking, hand detection and optimal data augmentation parameters.

\section{Keyboard detection and tracking}
\begin{figure}[hbt]
  \includegraphics[width=\textwidth]{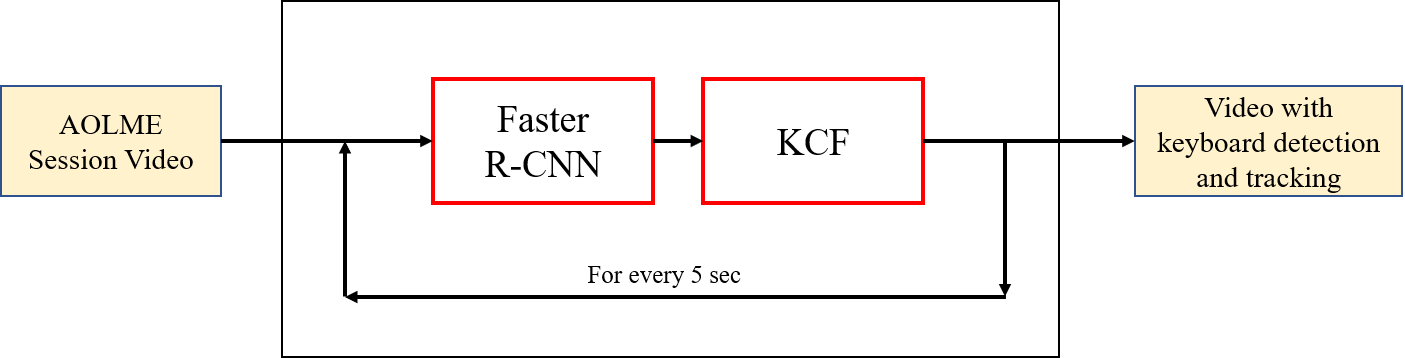}
  \caption{System for keyboard detection and tracking.}
  \label{fig:top_level_diag_typing}
\end{figure}

A top-level diagram for the keyboard detection and tracking system is shown in Figure
\ref{fig:top_level_diag_typing}.  Generally, tracking algorithms are
faster than detecting an object on every video frame. During tracking, the search region is restricted for reasonable motions. The thesis considers an adaptive system where object detection is performed once every n frames, followed by object tracking in the remaining frames.

I implemented a combined network of detection and fast-tracking
such that the accuracy is the same or better than detection alone while being
significantly faster. The fast-tracking is achieved by combining the fast tracker, KCF, selected as the fastest and best performing tracking methods implemented in OpenCV. 

\begin{figure}[!h]	
    \centering
    \includegraphics[width=\textwidth]{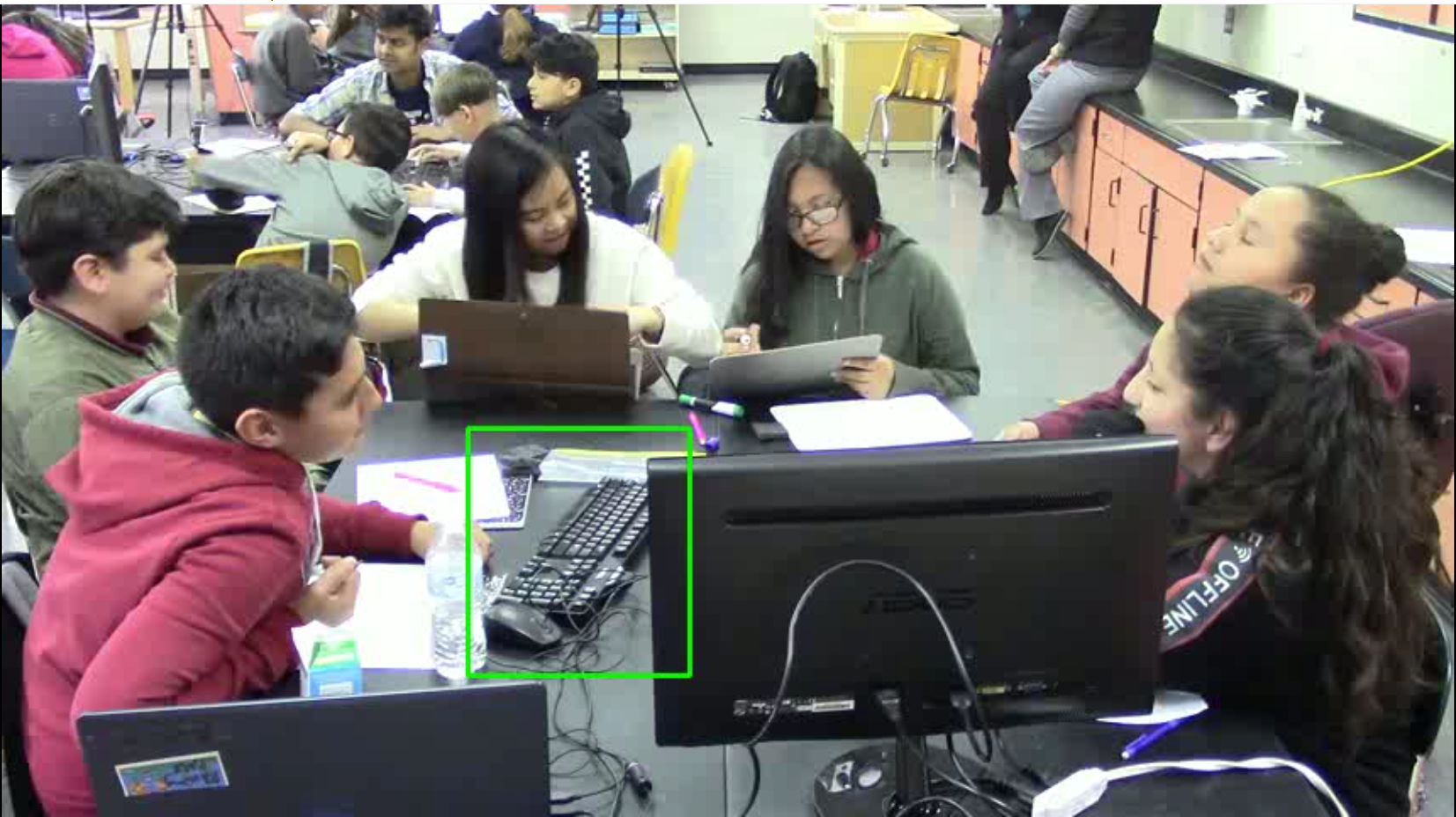}
    \caption{An example of keyboard detection.}
    \label{fig:kb_det}
\end{figure}

\begin{figure}[ht!]
    \centering
    \includegraphics[width=\textwidth]{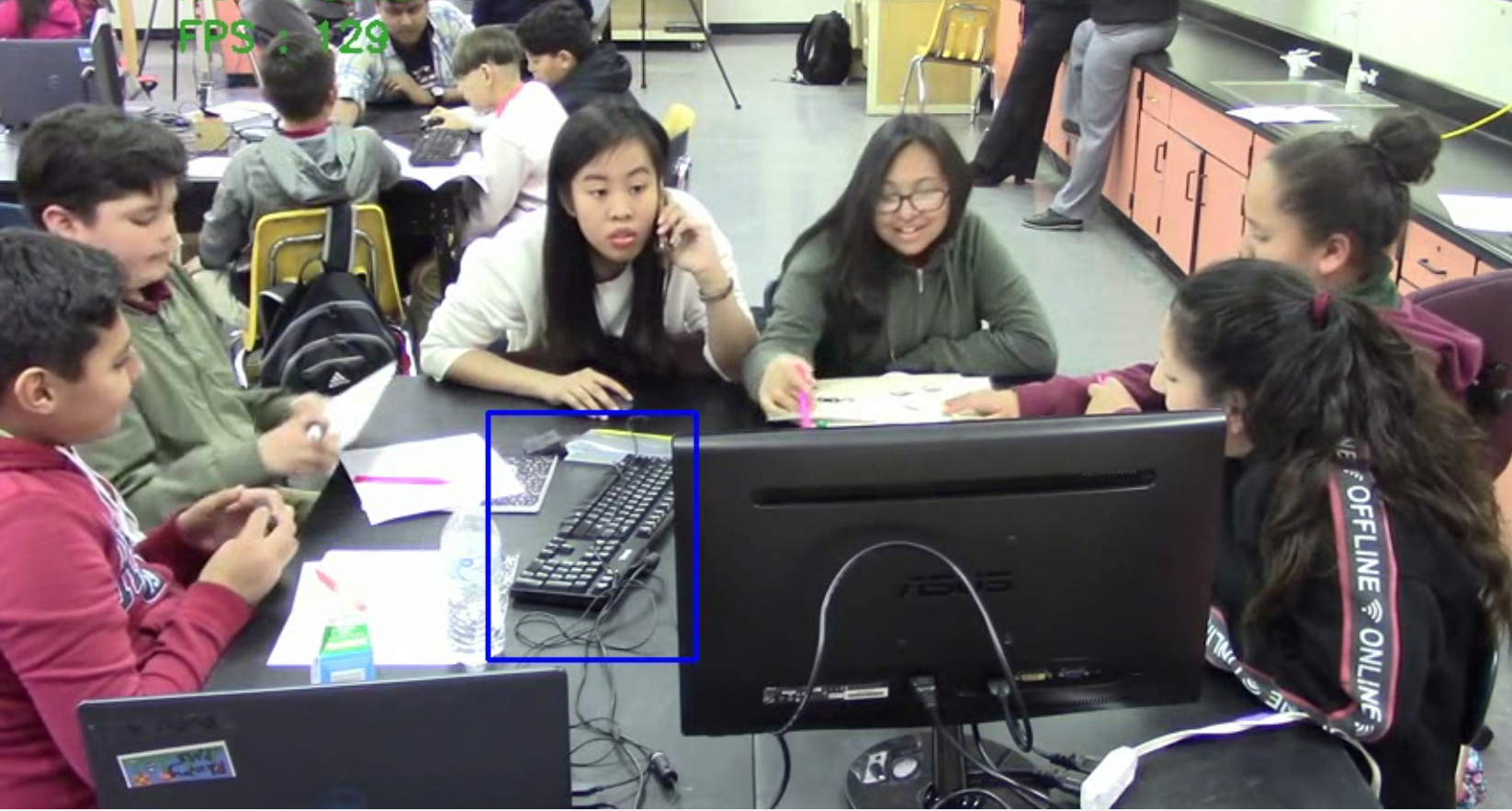}
    \caption{An example of keyboard tracking.}
    \label{fig:kb_trck}
\end{figure}

A session video is given as input, and the object of interest is
detected for the first frame, and it is tracked for the next 5-second
interval. Then, it is again re-initialized with the detection after every
5-second interval. This process is repeated until the end of the
video. Fig. \ref{fig:kb_det} shows an example from keyboard detection
and Fig. \ref{fig:kb_trck} shows an example from tracking.

\section{Top-level diagram for hand regions}
\begin{figure}	[!ht]
	\centering
	\includegraphics[width=\textwidth]{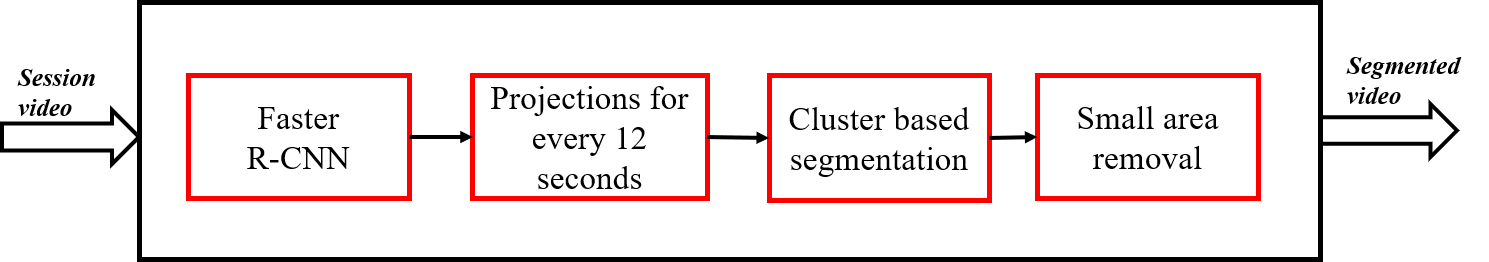}
	\caption{Hand region detection.}
	\label{fig:top_level_diag_writing}
\end{figure}

The top-level diagram for generating hand regions is as shown in Fig. \ref{fig:top_level_diag_writing}. It has five different steps. It takes a session video as input and produces the output video with bounding boxes around potential instances of activities involving hand movement. \\

\noindent
\textbf{Components of Top-level diagram for hand regions}:
\begin{enumerate}
\item \textbf{Object Detection} \\
Input Video is passed through an Object Detector (Faster-RCNN) to detect hands throughout the video for every second, as shown in figure \ref{fig:det}. 
\begin{figure}	[!h]
	\centering
	\includegraphics[width=\textwidth]{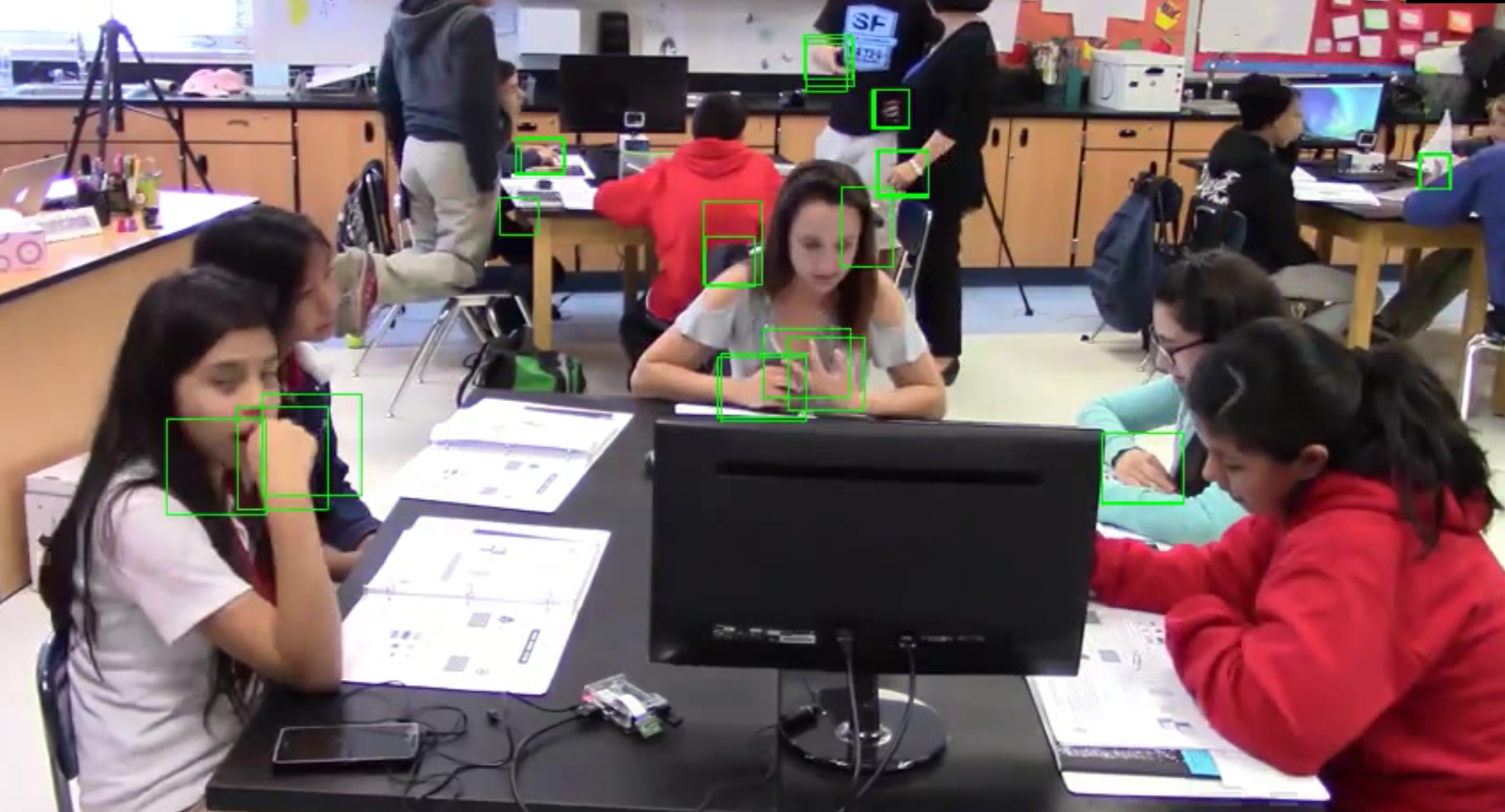}
	\caption{Video frame showing hands detected.}
	\label{fig:det}
\end{figure}

\item \textbf{Projection for every 12 seconds} \\
 All the detections for every 12 seconds are added up and projected on to an image as shown in Fig. \ref{fig:proj}. 
\begin{figure}[!h]
	\centering
	\includegraphics[width=\textwidth]{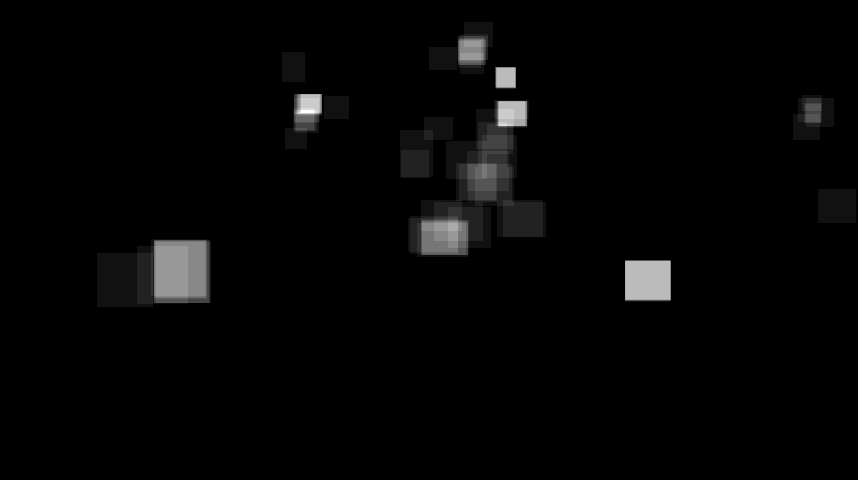}
	\caption{Video frame showing projected detections for 12-seconds interval}
	\label{fig:proj}
\end{figure}   
 
\item \textbf{Cluster based segmentation} \\ 
ISODATA \cite{ball1965isodata}  thresholding is used to automatically find a threshold value for a given grayscale image. ISODATA is an unsupervised, iterative process for computing the minimum Euclidean distance when assigning each candidate cell to a cluster.  \\
A variety of clustering techniques i.e., Isodata, Li, Mean, Minimum, Otsu, Triangle, Yen \cite{van2014scikit} as shown in figure \ref{fig:diff_thresholding} were tried. Over an exploratory dataset, ISODATA gave the best over-segmentation results by capturing the writing regions using the minimum number of clusters. It is as shown in figure \ref{fig:clustered}.
\begin{figure}	[!h]
	\centering
	\includegraphics[width=\textwidth]{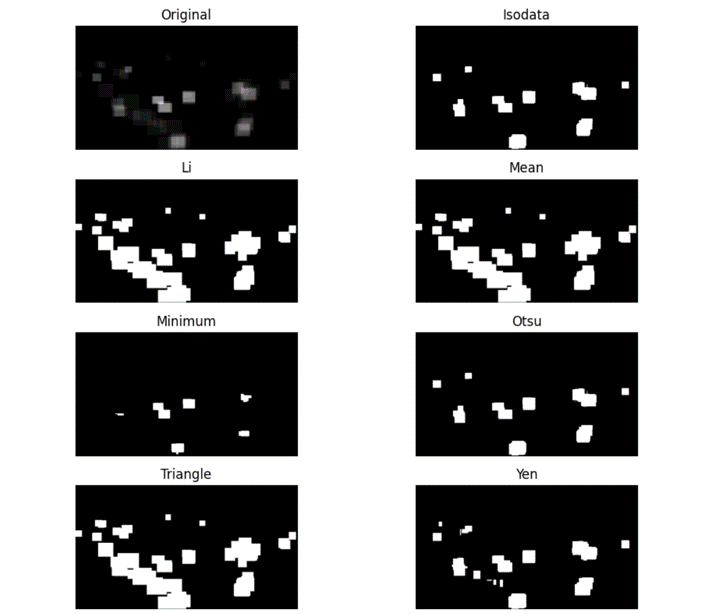}
	\caption{picture showing different clustering techniques.}
	\label{fig:diff_thresholding}
\end{figure} 
\begin{figure}	[!h]
	\centering
	\includegraphics[width=\textwidth]{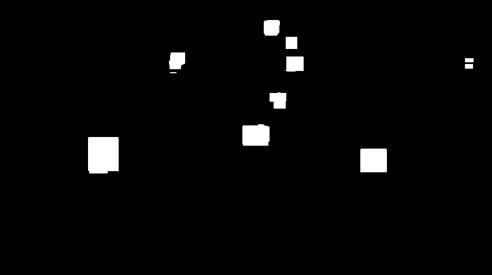}
	\caption{Video frame showing isodata clustering.}
	\label{fig:clustered}
\end{figure}

\item \textbf{Small area removal} \\
  All the regions whose areas are smaller than the smallest ground truth regions are removed as shown in Fig. \ref{fig:labeled_img}.
  
\begin{figure}	[!h]
	\centering
	\includegraphics[width=\textwidth]{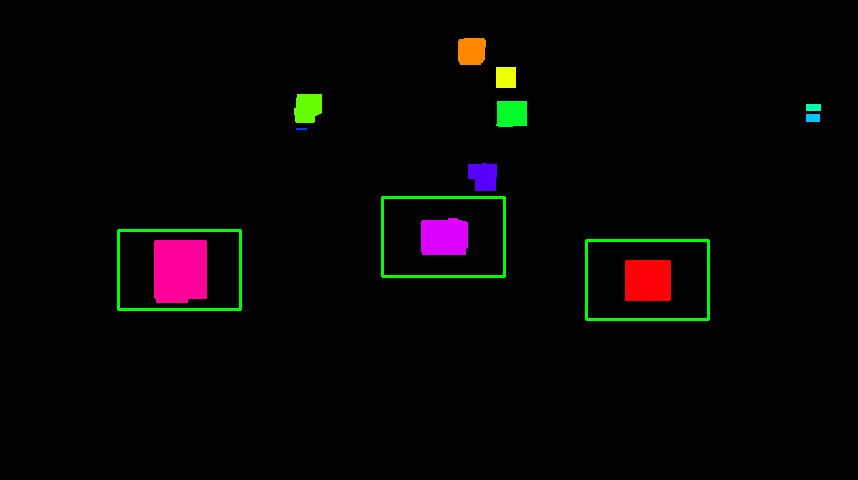}
	\caption{Video frame showing removal of smaller regions.}
	\label{fig:labeled_img}
\end{figure}   
  
\item \textbf{Segmented video} \\ 
Bounding boxes are drawn around remaining regions with the median width and height from ground truth regions as shown in Fig. \ref{fig:output}.

\begin{figure}	[!h]
	\centering
	\includegraphics[width=\textwidth]{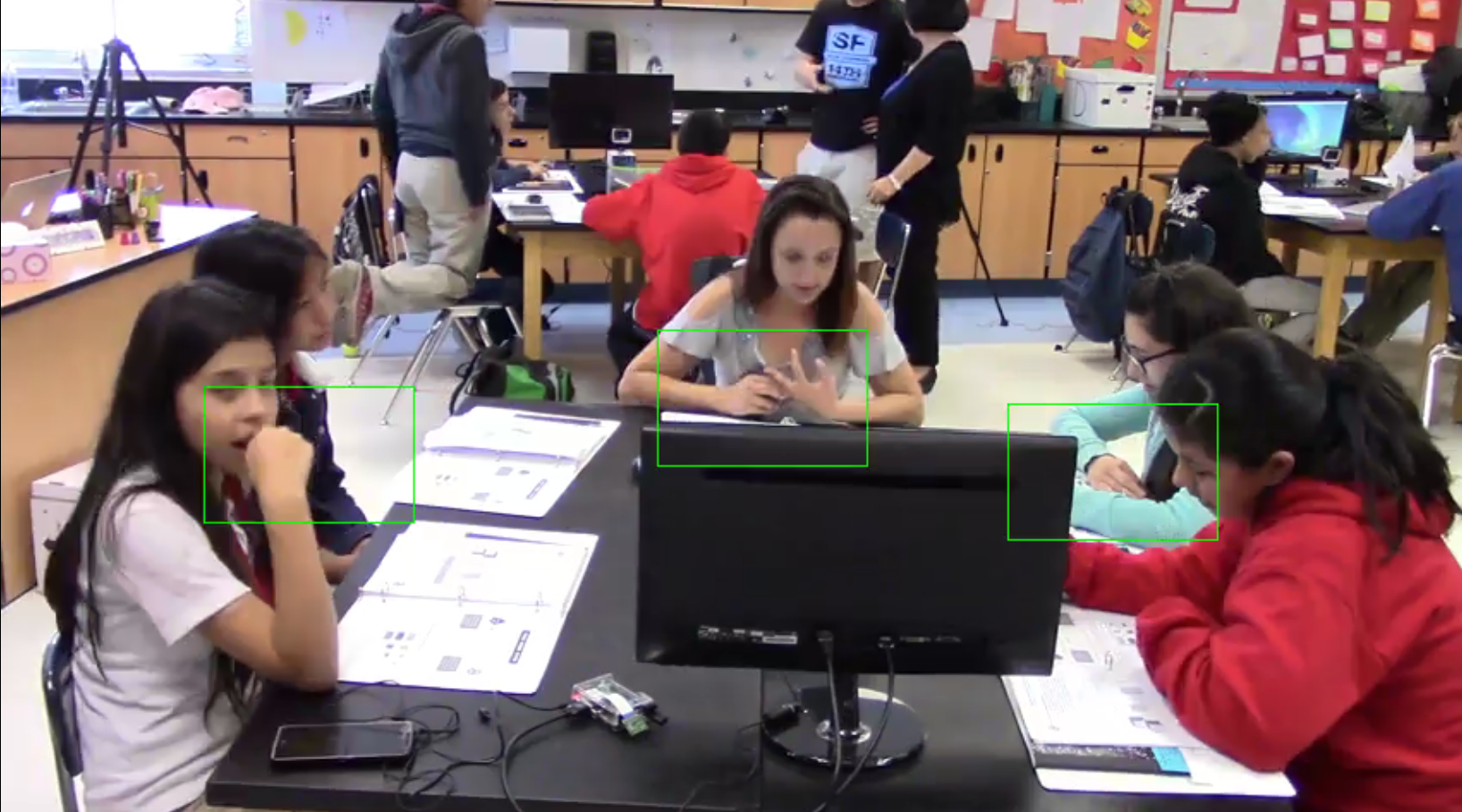}
	\caption{Video frame with bounding boxes around hand activity regions.}
	\label{fig:output}
\end{figure}

\end{enumerate}

\section{Optimal data augmentation study}
    A data augmentation study was performed here to improve hand detection results. First, an optimal range of angles for shearing, rotation, and pixels to be translated are determined separably. Once all the optimal values are found, an optimal study of probabilities to apply these augmentations is performed. Once all the optimal values are found, finally, the model is trained with optimal data augmentation parameters.

\subsection{Experimental Setup} 
To perform all these experiments, the library used is MMDetection \cite{chen2019mmdetection}. The detection algorithm used is Faster R-CNN. The study was done on RTX 5000 GPU with a learning rate of 0.001. The number of epochs used here is 12, as recommended. The mini-batch size is two images with two workers.\\

\noindent
\textbf{Randomized affine transformations} \\
\textbf{Affine transformation.} Transformation of an image such that parallel lines in an image remain parallel after the transformation. Scaling, translation, rotation shearing are all examples of affine transformations. \\
\noindent
\textbf{Rotate angle:} \\
The image and the corresponding bounding box are rotated with the angle $\theta$. A Positive angle rotates it counter clockwise while the negative angle rotates the image clockwise. \\
\noindent
\textbf{Translation pixels:} \\
The image and the corresponding bounding box are translated horizontally towards left and right based on the values. \\
\noindent
\textbf{Shear angle:} \\
Shearing slides one edge of an image along the X or Y axis, creating a parallelogram. Horizontal shear slides an edge along the X direction and the corresponding bounding box too. Similarly, vertical shear slides an edge along the Y direction. Shear angle specifies the number of degrees to shear the images. In the experiments, horizontal shearing is used. \\

The optimal values for shear angle, rotate angle and translation pixels are obtained using the pseudo code shown in Fig. \ref{alg:optimal_ranges}. For Shear Angle optimization, the values of  $\Theta$ are {2$^\circ$, 4$^\circ$, 8$^\circ$, 16$^\circ$, 32$^\circ$.} For Rotate Angle Optimization, values of $\Theta$ are {2$^\circ$, 4$^\circ$, 8$^\circ$, 16$^\circ$, 32$^\circ$.} For translation pixels, values of $\Theta$ are {2, 4, 8, 16, 32, 64, 128, 256, 512, 800}.

\begin{figure}[H]
	\begin{algorithmic}    		
		\FOR{each $\theta$ in list($\Theta$)}
		\STATE Augment all the training images with the $\theta$.
		\STATE Train the model with the augmented dataset.
		\STATE	 Test on validation set.
		\STATE Record validation accuracy.
		\ENDFOR
		\STATE \textit{\textbf{Repeat}} the above procedure untill validation accuracy starts decreasing.
		\STATE \textit{\textbf{Repeat}} the same procedure for all the transformations, shear, rotate and translation.
		\STATE \textit{\textbf{List}} all the optimal ranges for each of the transformation.
	\end{algorithmic}
	\caption{Pseudo code for finding optimal ranges for each of the affine transform.}
	\label{alg:optimal_ranges}
\end{figure}

Once the optimal ranges are determined separably, an optimal probability was determined using the pseudo code shown in \ref{alg:optimal_prob}.

\begin{figure}[H]
	\begin{algorithmic}   
		\FOR{each p in list(P)}
		\FOR{image in images with p}
		\STATE	Apply random horizontal flips with p.
		\STATE	Apply Scaling of [0.8,1.2] with p.
		\STATE	Apply random shear angle transforms between optimal range with p.
		\STATE	Apply random rotate angle transforms between optimal range with p.
		\STATE	Apply random horizontal translate pixel transforms between optimal range with p.
		\ENDFOR
		\STATE Train the model with this dataset.
		\STATE Test on validation set and record validation accuracies.
		\ENDFOR
		\STATE \textit{\textbf{Note}} the p which gave best validation accuracy. 
	\end{algorithmic}
	\caption{Pseudo code for finding optimal probability.}
	\label{alg:optimal_prob}
\end{figure}

After obtaining optimal ranges for affine transforms and optimal probability, the model is trained with these parameters. So the optimization is performed for probability, p, and range of $\theta$. \\

\chapter{Results}
This chapter summarizes detection, tracking and projection results. We first present keyboard
detection and tracking, followed by hand detection and projections.
\section{Keyboard Detection and Tracking Results} 
This section describes the results of keyboard detection, tracking and the combined system of detection and tracking.

\subsection{Keyboard detection results}
\begin{figure}[!ht]	
  \begin{subfigure}{0.49\columnwidth}
    \centering
    \includegraphics[width=\textwidth]{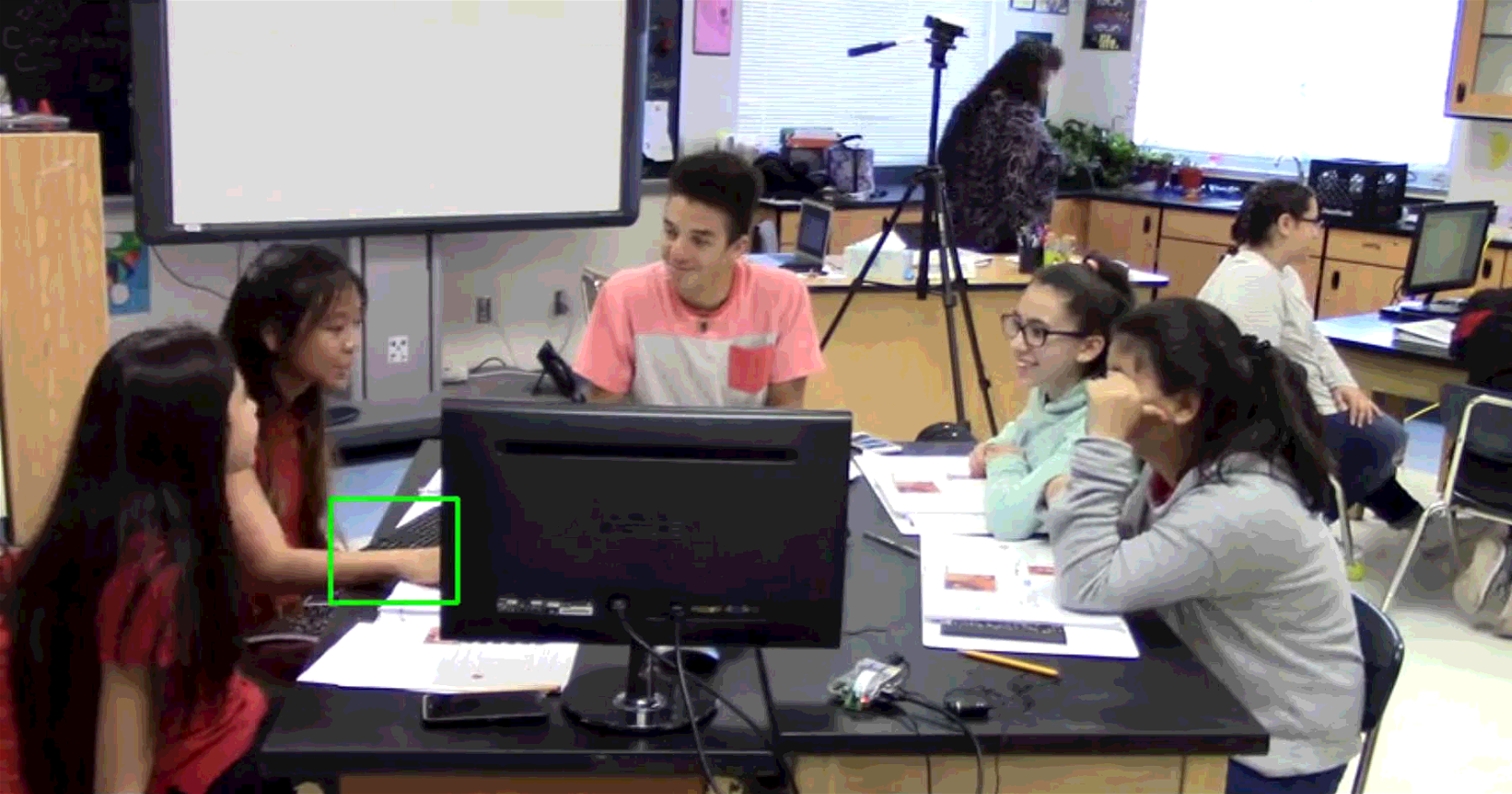}
    \caption{Successful detection with partial occlusion.}
    \label{fig:}
  \end{subfigure}
  \begin{subfigure}{0.49\columnwidth}
    \centering
    \includegraphics[width=\textwidth]{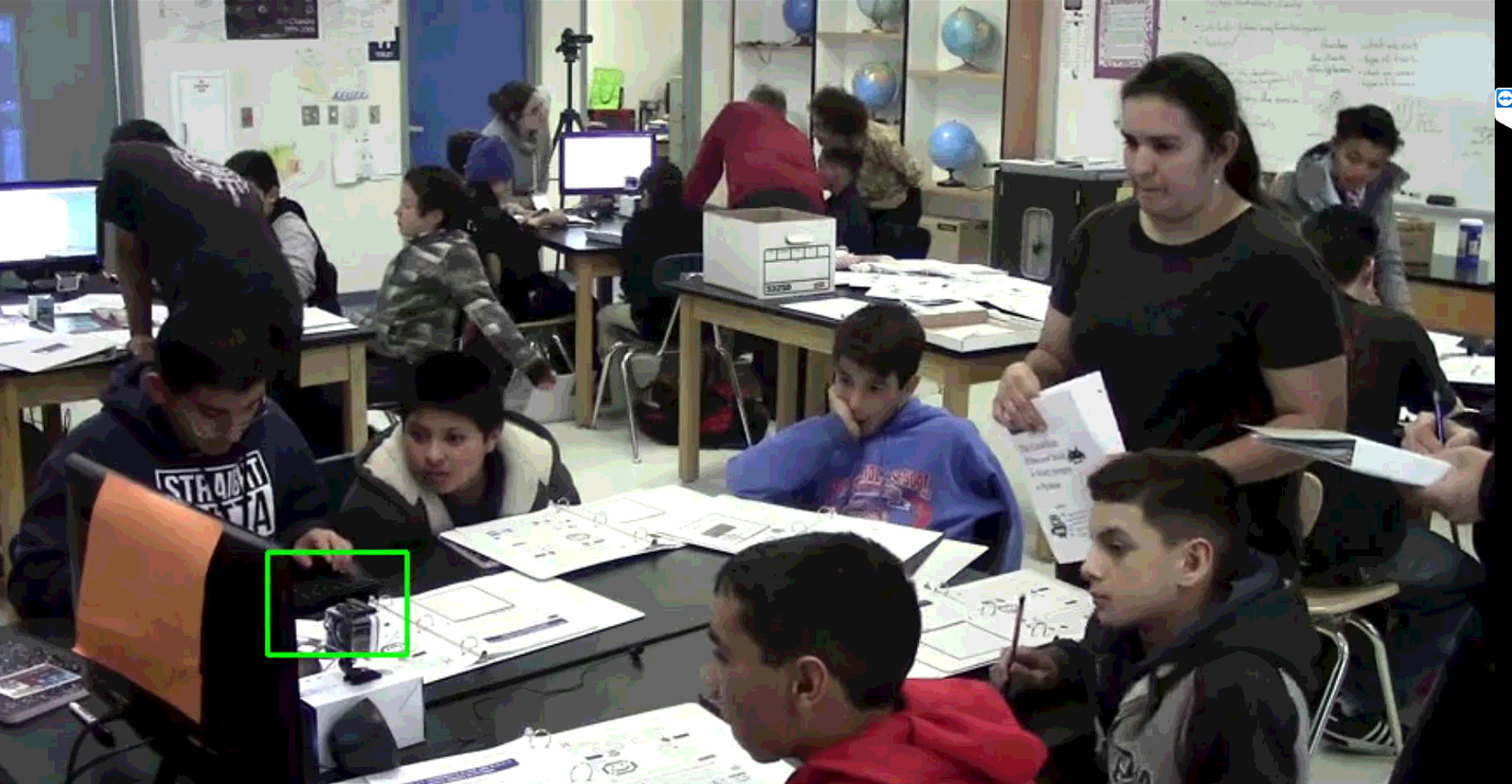}
    \caption{Successful detection with partial occlusion.}
    \label{fig:}
  \end{subfigure}
  
    \begin{subfigure}{0.49\columnwidth}
    \centering
    \includegraphics[width=\textwidth]{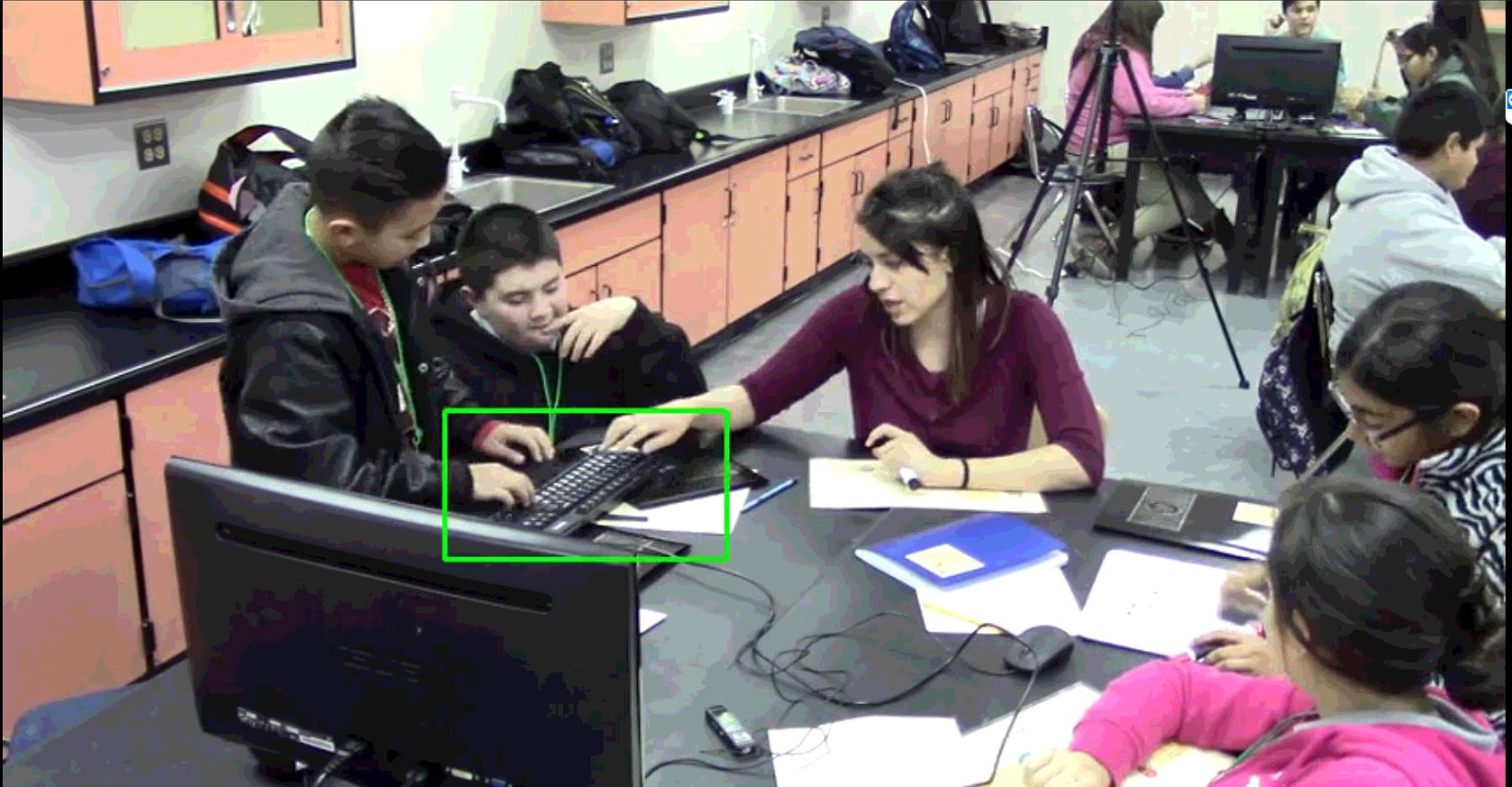}
    \caption{Successful detection with hands.}
    \label{fig:}
  \end{subfigure}
\begin{subfigure}{0.49\columnwidth}
    \centering
    \includegraphics[width=\textwidth]{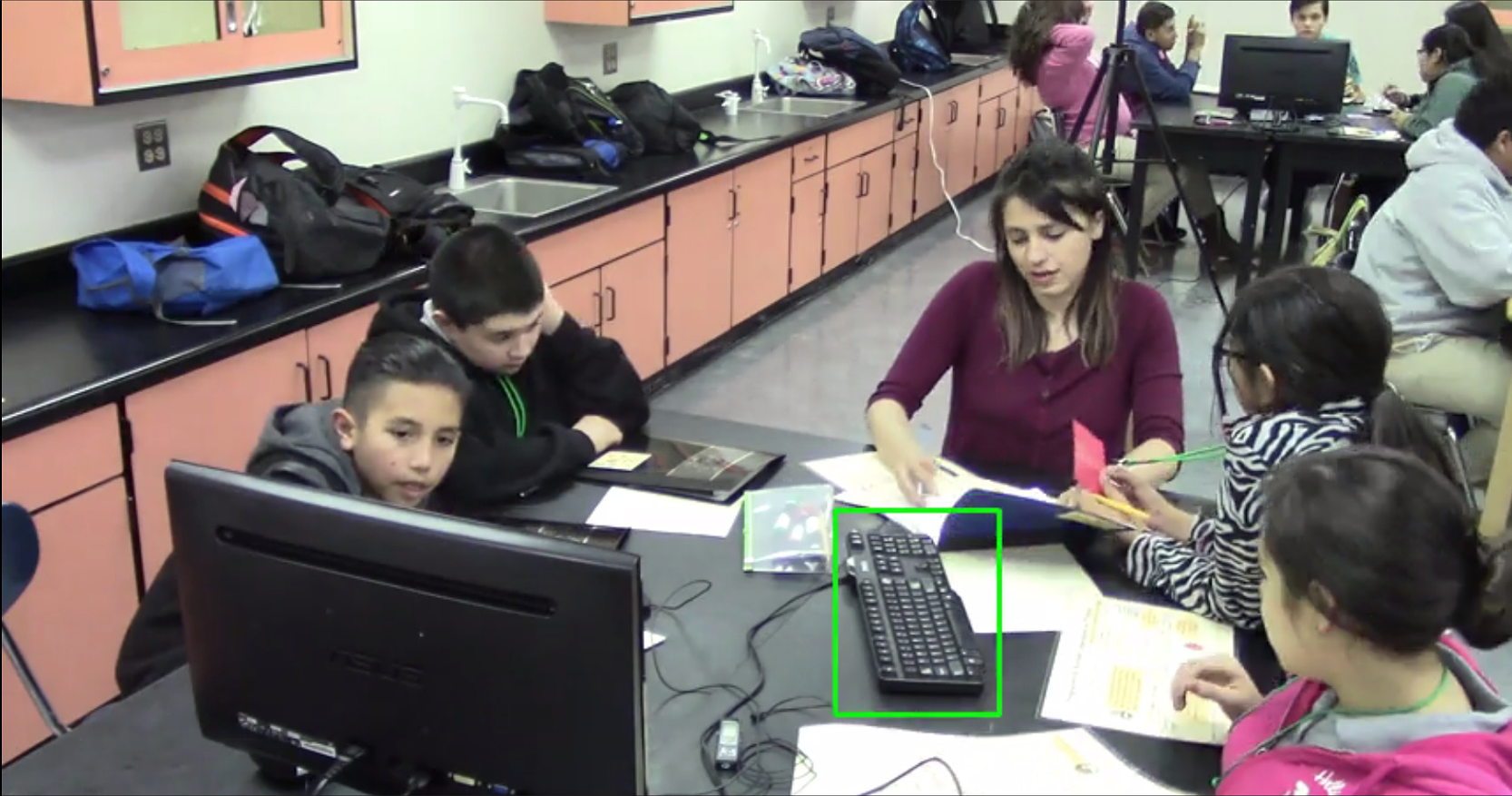}
    \caption{Successful detection without hands.}
    \label{fig:}
  \end{subfigure}
  \caption{Successful keyboard detections.}
  \label{fig: kb_success_results}
\end{figure}

\begin{figure}[!ht]
  \begin{subfigure}{0.50\columnwidth}
    \centering
    \includegraphics[width=\textwidth]{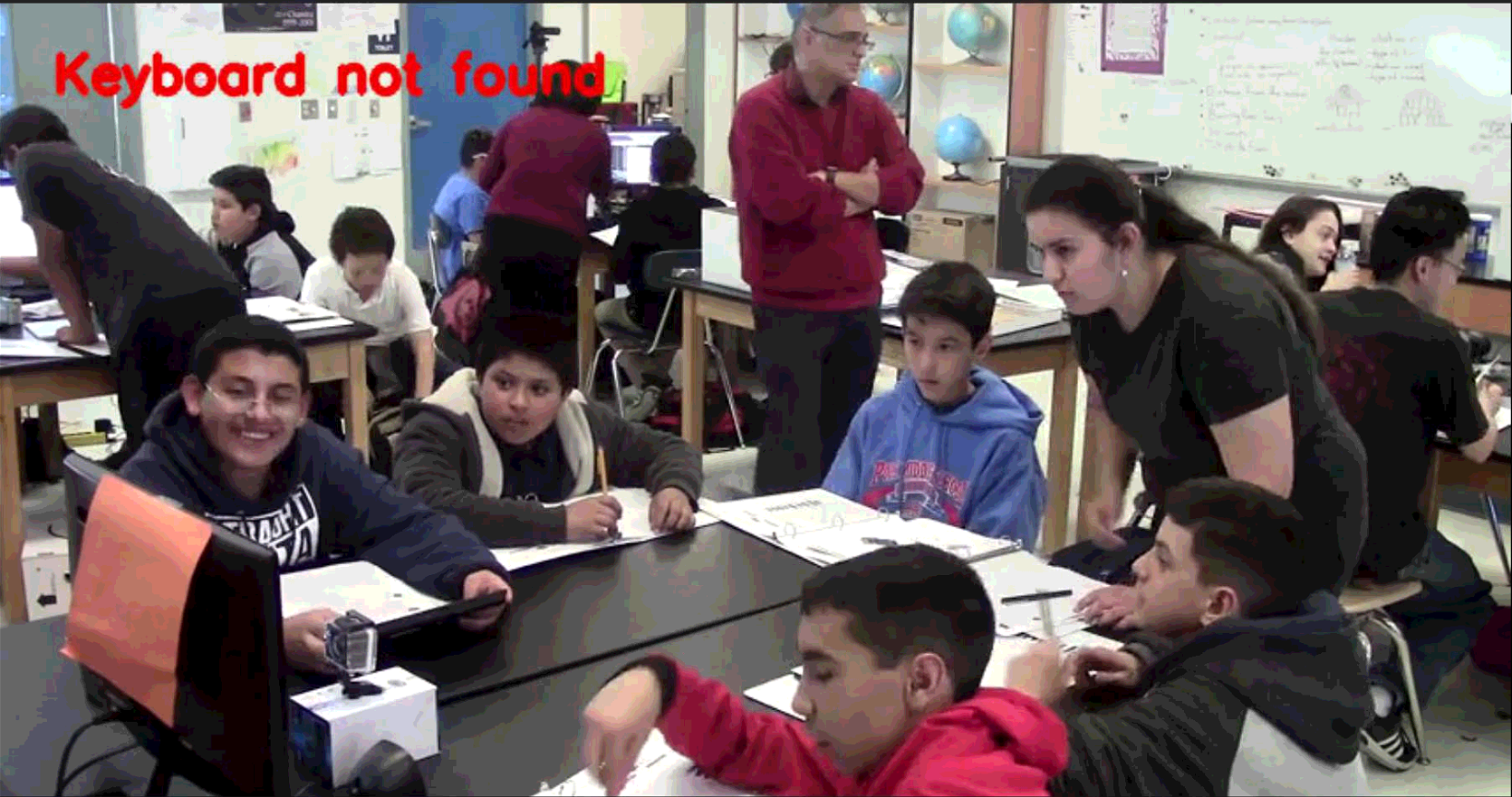}
    \caption{Failure to detect keyboard due to lack of visible keys.}
    \label{fig:fn}
  \end{subfigure}
  \begin{subfigure}{0.49\columnwidth}
    \centering
    \includegraphics[width=\textwidth]{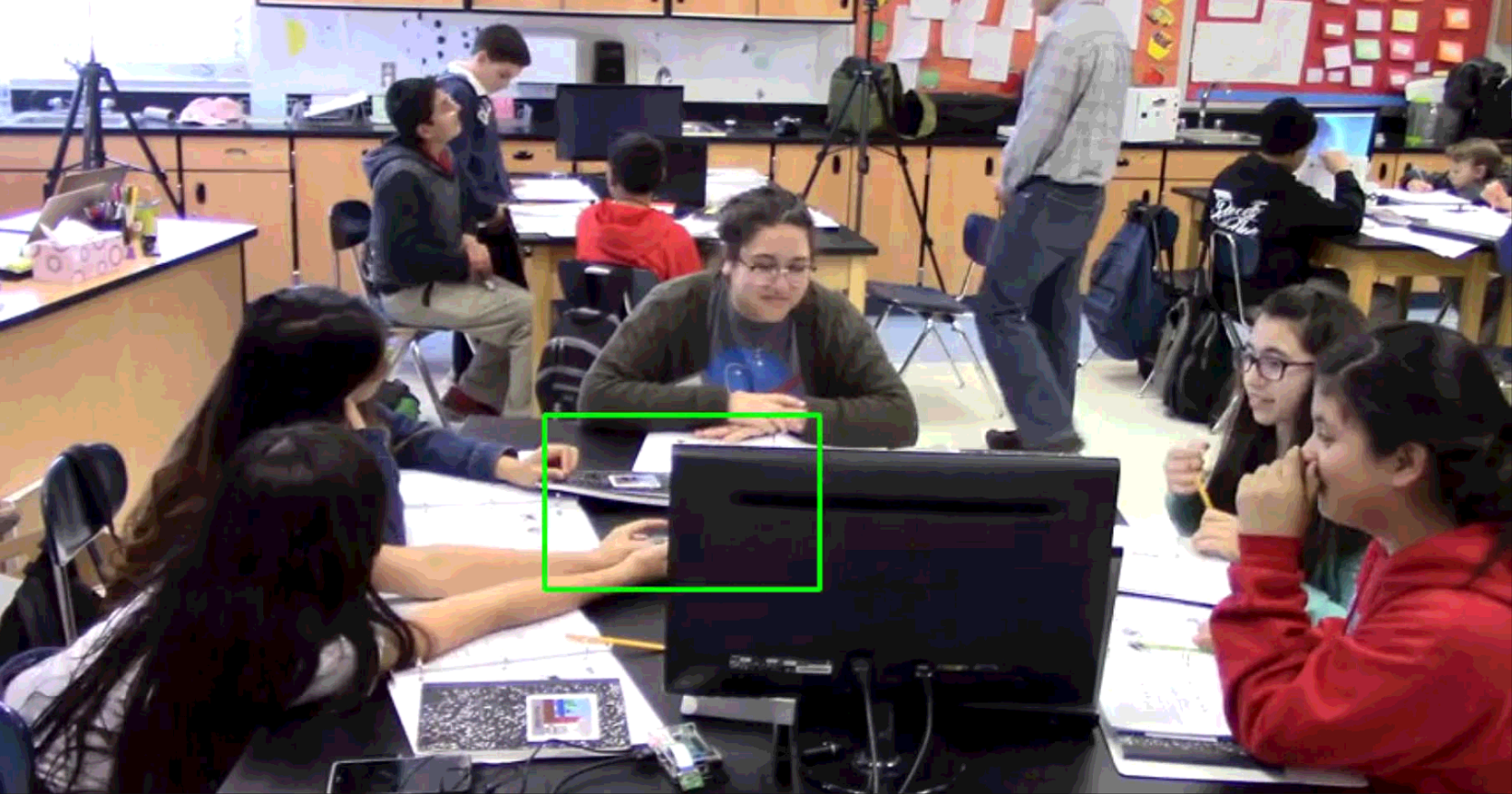}
    \caption{False positive due to similar looking book.}
    \label{fig:}
  \end{subfigure}
  
  \begin{subfigure}{0.50\columnwidth}
    \centering
    \includegraphics[width=\textwidth]{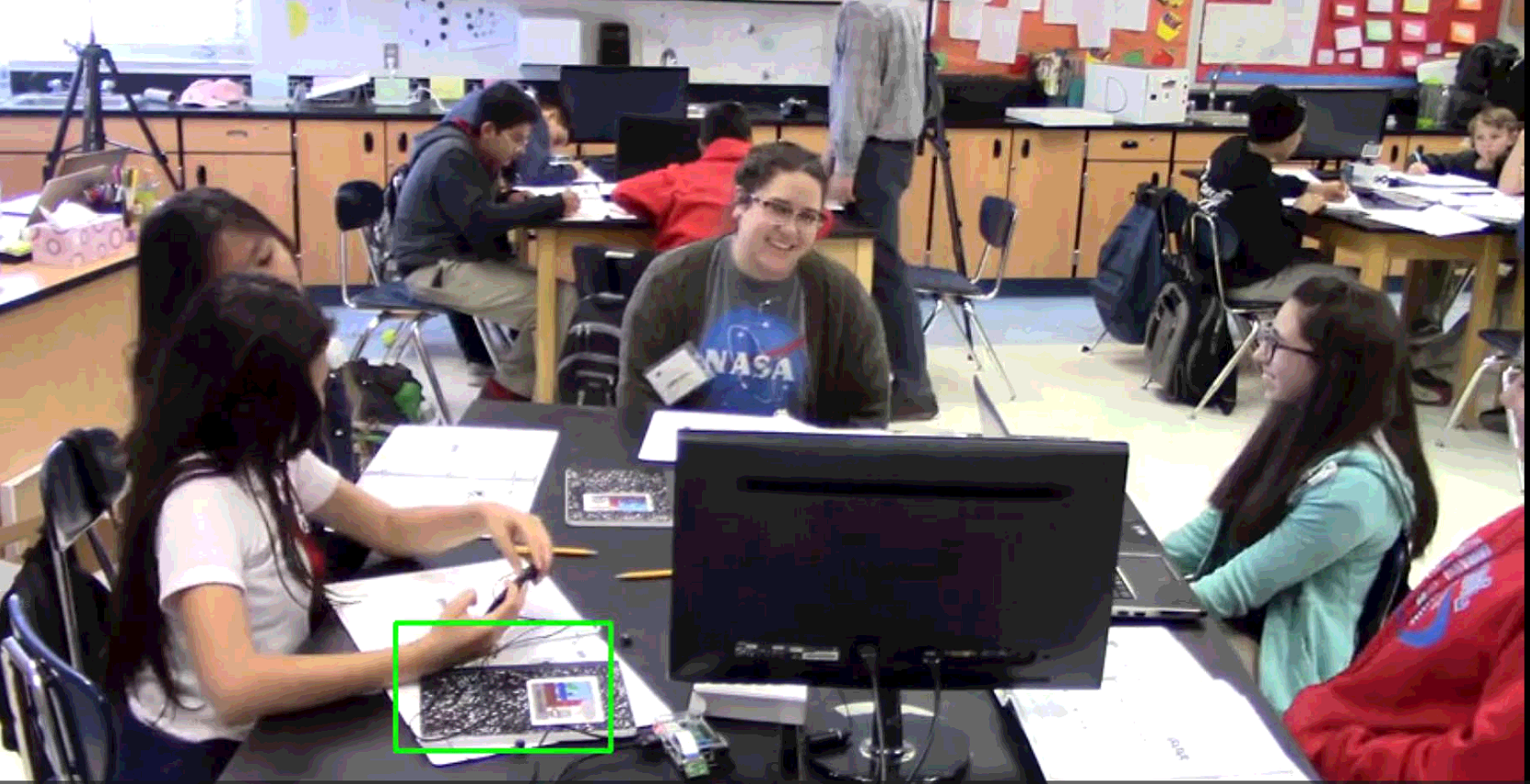}
    \caption{False positive due to similar looking book.}
    \label{fig:}
  \end{subfigure}
   \begin{subfigure}{0.49\columnwidth}
    \centering
    \includegraphics[width=\textwidth]{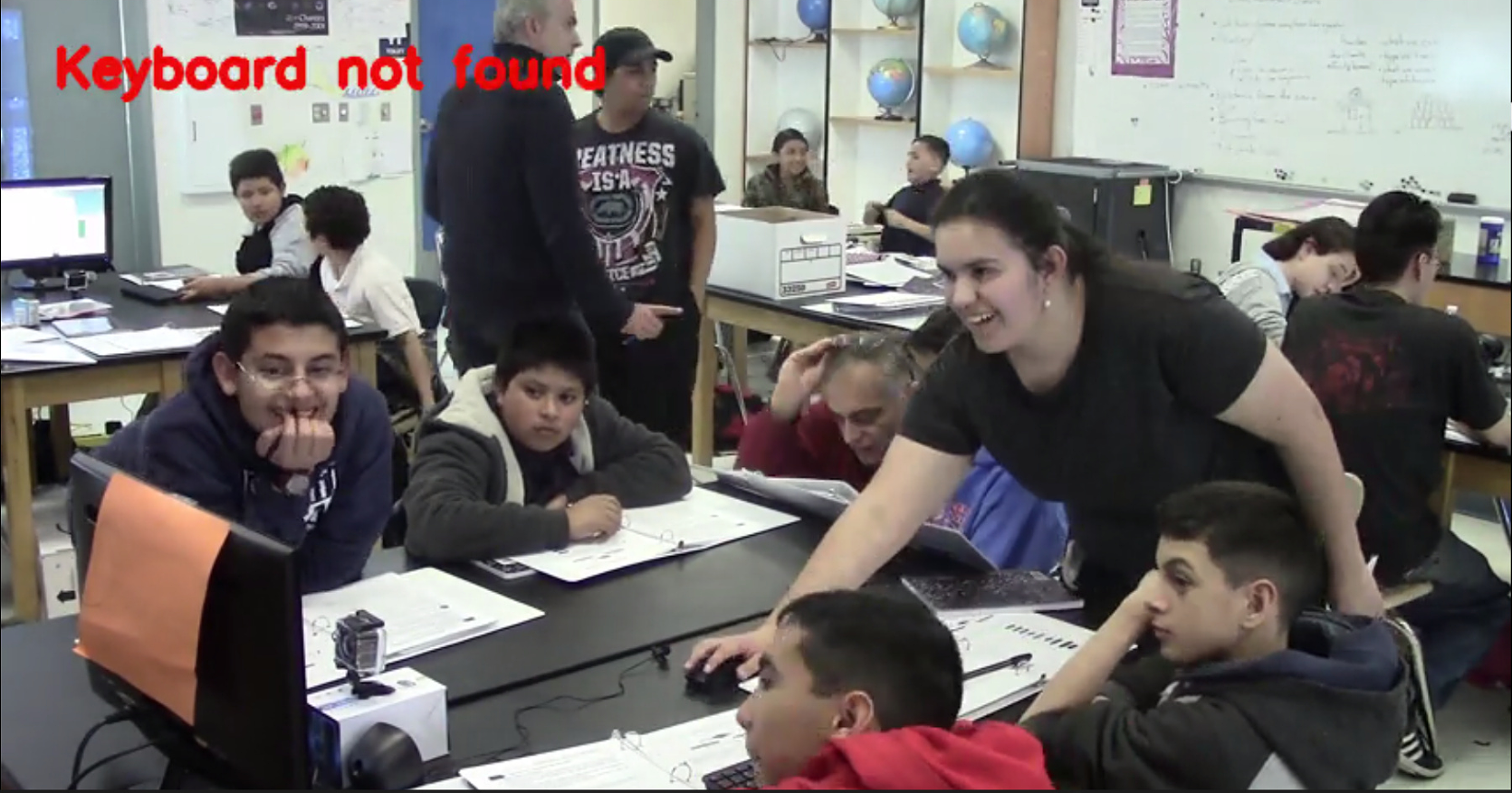}
    \caption{Failure due to occlusion.}
    \label{fig:}
  \end{subfigure}
  \caption{Some of the test results showing failure of keyboard detection}
   \label{fig: kb_failure_results}
\end{figure}

For keyboard detection, library named Detectron2 \cite{wudetectron2} is used. Detectron2 is Facebook AI Research's next-generation library that provides state-of-the-art detection and segmentation algorithms. The method used for keyboard detection is Faster R-CNN, which is pre-trained on COCO \cite{lin2014microsoft} detection dataset. The re-training time for 300 iterations is about 20 minutes. We were able to achieve a very high AP of 0.92 for keyboard detection. However, the drawback with detection alone is that the inference time for video is more. 
Results for keyboard detection are shown in table \ref{results:keyboard detection}. An AP of 0.92 was achieved at an IOU of 0.5. AP @IoU=0.5 represents the model has used threshold of 0.5 to remove unnecessary boxes. Similarly, range 0.50:0.95 represents AP and AR are averaged over multiple IoU values. Specifically, we use 10 IoU thresholds of .50:.05:.95. Area = small represents small objects for area  $< 32^2 $. AR @ [maxDets=1] and
[maxDets=10] mean the maximum recall given 1 detection per image and 10
detections per image correspondingly.
Some of the success and failure cases are shown in Fig. \ref{fig: kb_success_results} and \ref{fig: kb_failure_results}, respectively. The failure cases mainly show the misclassification of the book as a keyboard.
\begin{table*}[hbt]
  \centering
  \caption{Average precision (AP) and Average recall (AR) at different IoU ratios for keyboard detection. We are able to achieve a very high
  AP (0.92) for keyboard detection.}
  \label{results:keyboard detection}
  \begin{tabular}{| l | l|}
    \hline
    \textbf{Metric} & \textbf{Value}\\
    \hline
    \hline
    AP @[IoU=0.50:0.95 |area = all | maxDets = 100]    & 0.614\\
    AP @[IoU=0.50 |area = all | maxDets = 100] 	    &  \textbf{0.922} \\
    AP @[IoU=0.75 |area = all | maxDets = 100]		    & 0.708\\
    AP @[IoU=0.50:0.95 |area = small | maxDets = 100]  & -1.000\\
    AP @[IoU=0.50:0.95 |area = medium | maxDets = 100] & 0.611\\
    AP @[IoU=0.50:0.95 |area = large | maxDets = 100]  & 0.620\\
    AR @[IoU=0.50:0.95 |area = all | maxDets = 1] 		& 0.659\\
    AR @[IoU=0.50 |area = all | maxDets = 10] 			& 0.676\\
    AR @[IoU=0.75 |area = all | maxDets = 100]			& 0.676\\
    AR @[IoU=0.50:0.95 |area = small | maxDets = 100]  & -1.000\\
    AR @[IoU=0.50:0.95 |area = medium | maxDets = 100] & 0.670\\
    AR @[IoU=0.50:0.95 | area = large | maxDets = 100]  & 0.679\\
    \hline
  \end{tabular}
\end{table*}

\subsection{Tracking results} \label{tracking_res}
To determine how well the trackers perform with AOLME video, I experimented with calculating accuracy and speeds, respectively. To conduct the experiments, the system used is Intel Xeon CPU ES-2640, 16 cores per node at 2.6 GHz with RAM of 64 GB and GPU, Nvidia Tesla K40M. The video I used is of 23.45 minutes @ 30fps with a resolution of 858$\times$480. All the trackers are initialized with the same object and tracked throughout the end of the video. All the trackers failed with fast movements of the object. However, MOSSE, KCF, and Median flow proved to be real-time with 500, 159, 223 fps, respectively. Table \ref{tab:tracking} shows the speed of all the trackers on AOLME video.

\begin{table*}[!h]
  \centering
  \caption{Trackers performance using hardware system described in \ref{tracking_res} and OpenCV 4.0. Video under consideration is
   23.45 minutes long $858\times480$ @ 30 fps.}
  \label{tab:tracking}
  \begin{tabular}{|l|l| l|l|}
    \hline
    \textbf{Method} & \textbf{FPS} & \textbf{Comments} &\textbf{Accuracy}\\
    \hline
    Boosting & 17 & Not Real Time & Not Accurate \\\
    MIL 	 & 11 & Not Real Time & Not Accurate \\
    Mosse	 & 500 & \textbf{Real time, Fastest} &  \textbf{Accurate}\\
    Median Flow & 223 & \textbf{Real Time} & Not Accurate\\
    TLD & 21 & Not Real Time & Not Accurate\\
    KCF & 159 & \textbf{Real Time} &  \textbf{Accurate}\\
    GOTURN &  $<$7 & Not Real Time & Not Accurate\\
    CSRT & 25 & Not Real Time & Not Accurate\\
    \hline
  \end{tabular}
\end{table*}

\begin{figure}	[hbt]
  \includegraphics[width=\columnwidth]{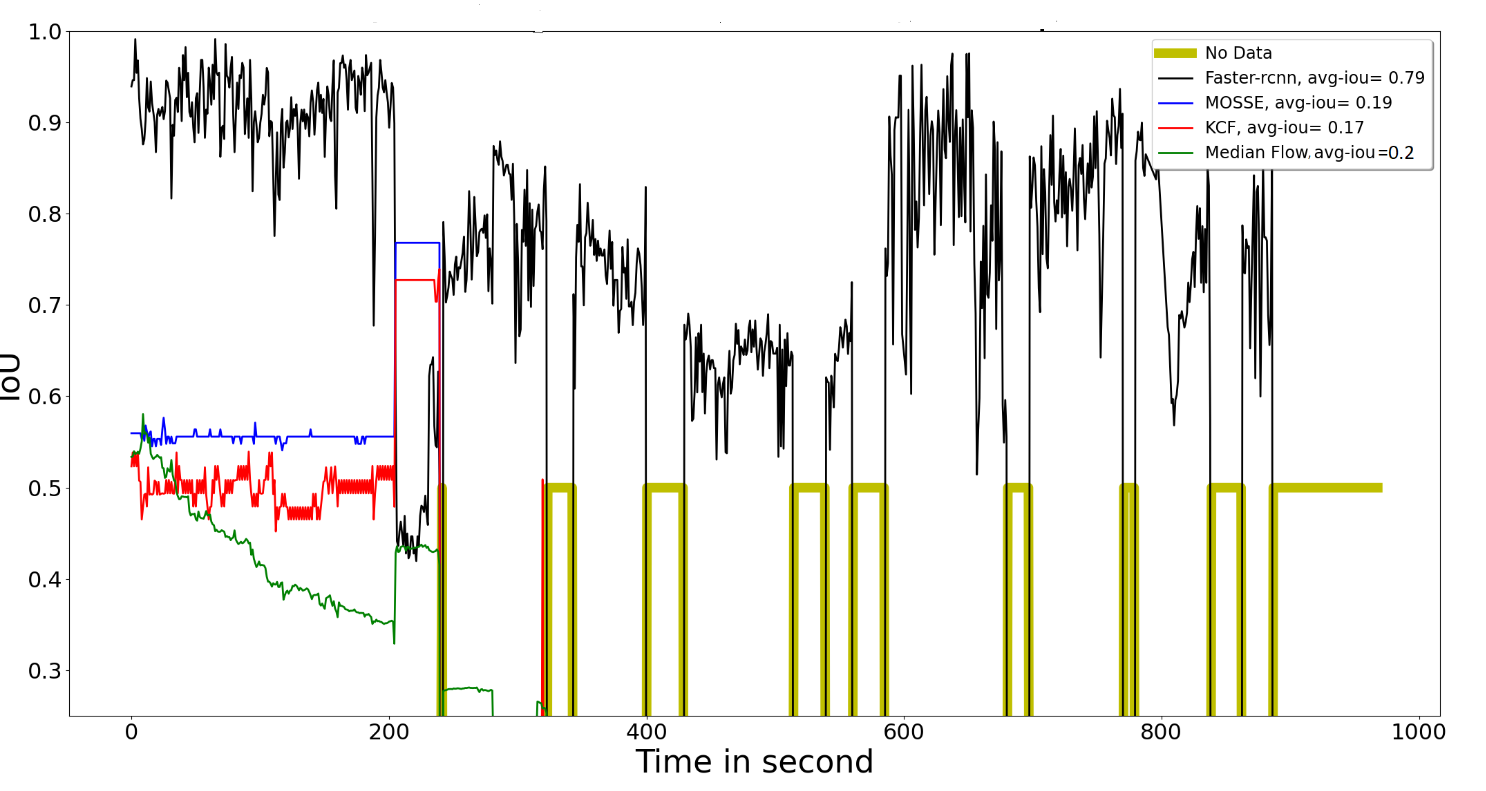}
  \caption{Detection vs fast-trackers.}
  \label{fig:det_vs_track}
\end{figure}

Figure \ref{fig:det_vs_track} shows the performance (IoU ratios) of detector and all the fast trackers across a video. The IoU ratios are plotted for every second throughout the video. No-Data represents there is no ground truth. This figure shows that all the trackers failed with fast movements of the object. The video under consideration for the figure is 16 minutes long, having resolution of $858\times480$ @ 30 fps.  

\subsection{Combination of detection and tracking results}
\begin{figure}	[hbt]
  \includegraphics[width=\columnwidth]{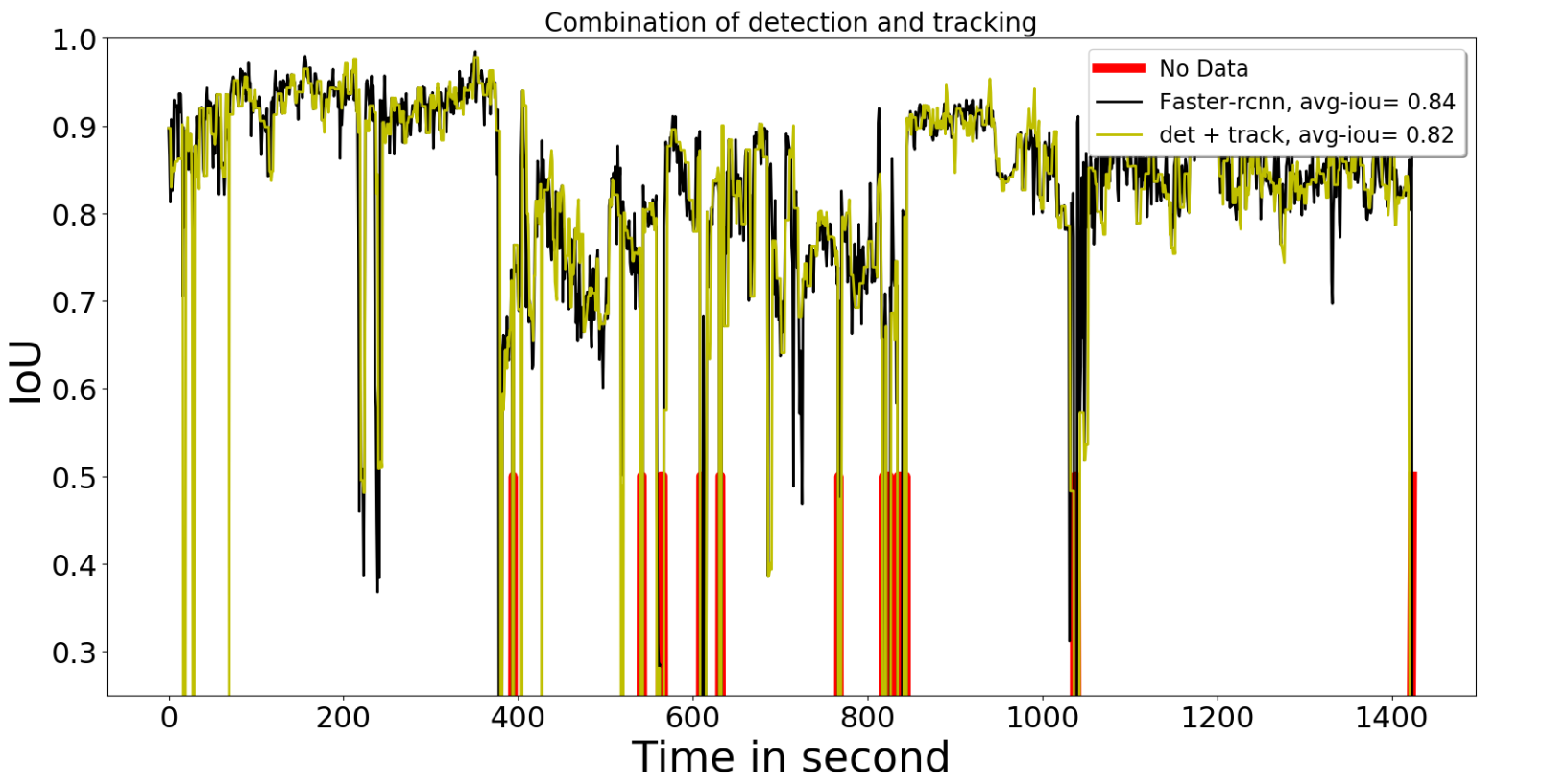}
  \caption{Combination of detection and tracking. Video under consideration is 23.45 minutes long having resolution of
  	$858\times480$ @ 30 fps.}
  \label{fig:det_and_track}
\end{figure}

We present the results of keyboard detection (every 5 seconds) followed by tracking. Figure \ref{fig:det_and_track} shows the results of (i) detection alone and (ii) a combination of detection and tracking. This method was able to achieve almost the same accuracy with a significantly faster running time. The fast tracker used to achieve this is KCF.
For running detection every second on a 23-minute video, the detection alone performed at 4.7 times the real-time rate, while the combined algorithm performed at 21 times the real-time rate, maintaining the same avg-IoU ratio. Fig. \ref{fig:det_and_track} shows that detection alone is able to achieve an average IoU of 0.84 for a video and the combined system achieved an average IoU of 0.82.

\clearpage
\section{Hand detection, optimal data augmentation and projection results} 
This section summarizes the results of hand detection, data augmentation optimization, and projections. 
\subsection{Hand detection results}
For hand detection, the library called MMDetection \cite{chen2019mmdetection} is used. To run these experiments, the training time for 12 epochs is 16 minutes. The dataset used for training is shown in table \ref{Dataset for hand detection}. The hand detector was able to achieve  Average precision (AP) of $0.72$ at 0.5 intersection over union (IoU). AP @ IoU=0.50 represents the model has used 0.5 threshold value to remove unnecessary bounding boxes.
\comment{
AP @ IoU=0.50 represents the model has used 0.5 threshold value to remove unnecessary bounding boxes. AR represents Average Recall. Area small represents score based on smaller objects, similarly large represents model has given AP/AR score based on larger objects in the data.
\begin{table*}[hbt]
  \centering
  \caption{hand detection results}
  \label{results:hand detction}
  \begin{tabular}{|l|l|}
    \hline
    \textbf{Metric} & \textbf{Value}\\
    \hline
    \hline
    AP @[IoU=0.50:0.95 |area = all | maxDets = 100]    & 0.243\\
    AP @[IoU=0.50 |area = all | maxDets = 100] 	    & \textbf{0.722}\\
    AP @[IoU=0.75 |area = all | maxDets = 100]		    & 0.081\\
    AP @[IoU=0.50:0.95 |area = small | maxDets = 100]  & 0.109\\
    AP @[IoU=0.50:0.95 |area = medium | maxDets = 100] & 0.269\\
    AP @[IoU=0.50:0.95 |area = large | maxDets = 100]  & 0.081\\
    AR @[IoU=0.50:0.95 |area = all | maxDets = 1] 		& 0.372\\
    AR @[IoU=0.50 |area = all | maxDets = 10] 			& 0.372\\
    AR @[IoU=0.75 |area = all | maxDets = 100]			& 0.372	\\
    AR @[IoU=0.50:0.95 |area = small | maxDets = 100]  & 0.276	\\
    AR @[IoU=0.50:0.95 |area = medium | maxDets = 100] & 0.390	\\
    AR @[IoU=0.50:0.95 | area = large | maxDets = 100]  & 0.400	\\
    \hline
  \end{tabular}
\end{table*}
}

\subsection{Optimal data augmentation parameters study}
\begin{figure}	[!ht]
  \includegraphics[width=\columnwidth]{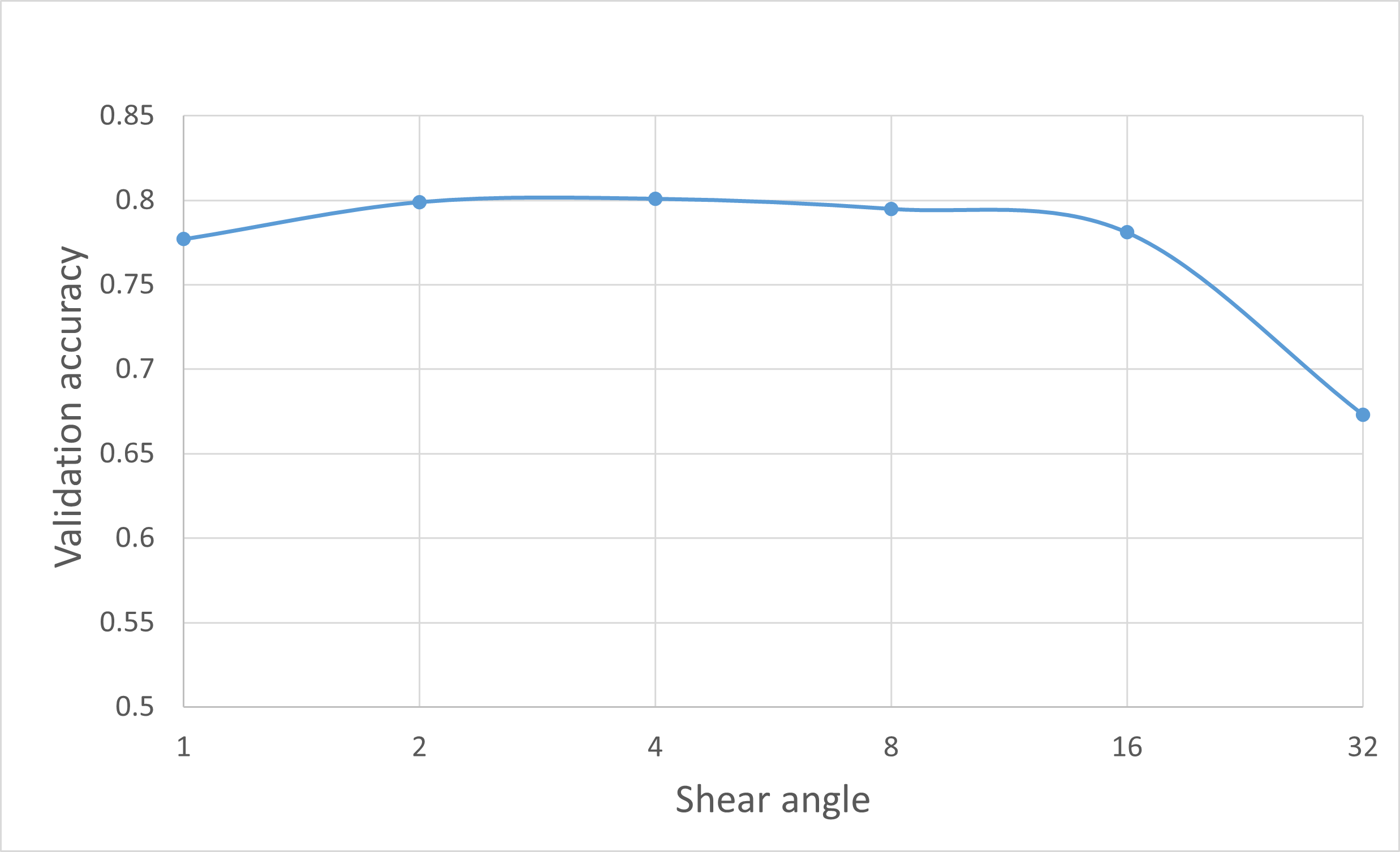}
  \caption{Shear angle optimization.}
  \label{fig:shear}
\end{figure}

\noindent
\textbf{Shear Angle Optimization}  \\
All the training images are augmented with each of the below angles, and maximum validation accuracies are recorded accordingly. 
The angles used for experiments are  {2$^{\circ}$, 4$^{\circ}$, 8$^{\circ}$, 16$^{\circ}$, and 32$^{\circ}$} . The validation accuracy started to decrease after 8$^{\circ}$. The best validation accuracy is achieved when angle is 4$^{\circ}$. So the range considered is (-3,3). The plot is shown in Fig. \ref{fig:shear}.

\begin{figure}	[!ht]
  \includegraphics[width=\columnwidth]{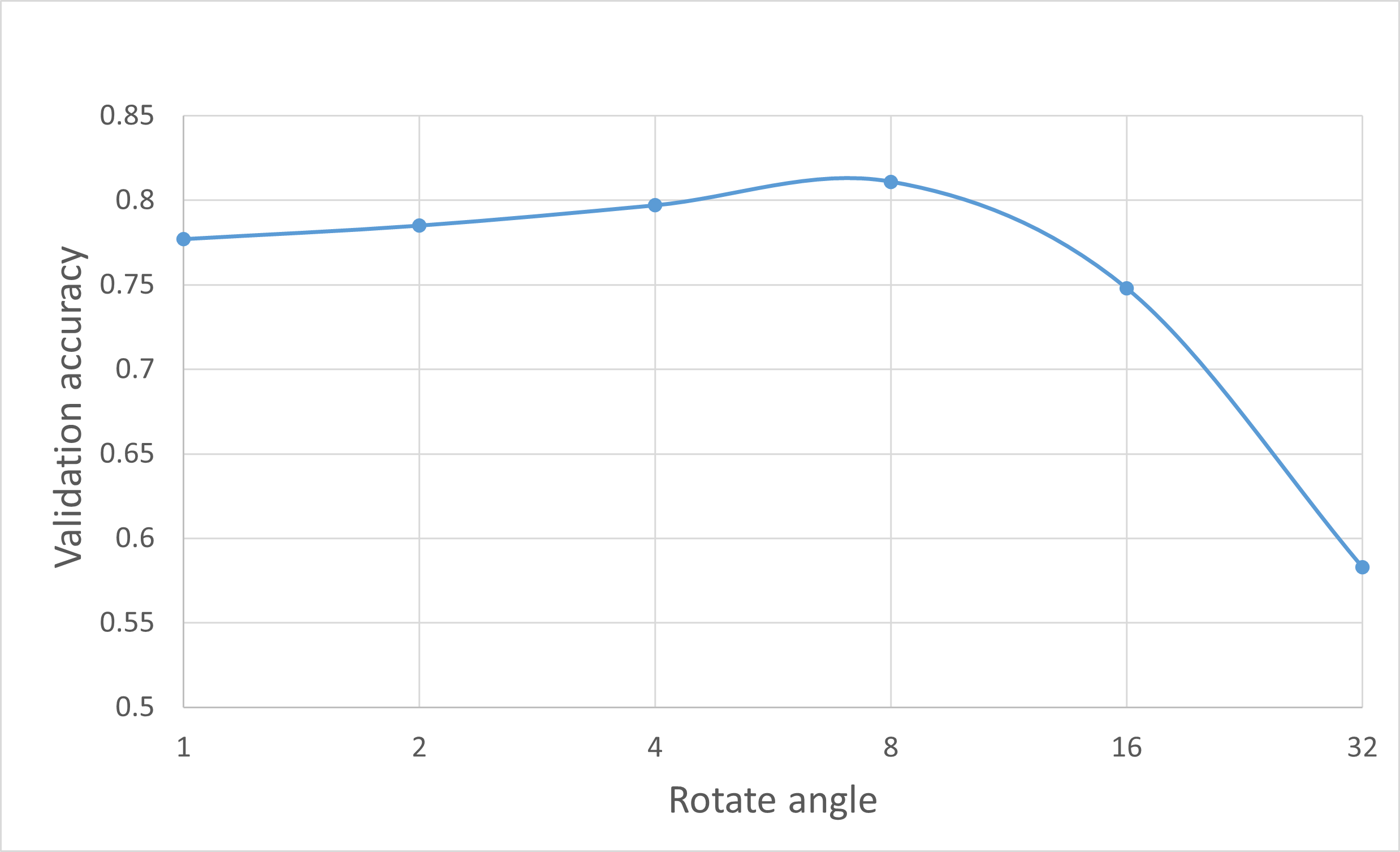}
  \caption{Rotate angle optimization.}
  \label{fig:rotate}
\end{figure}

\noindent
\textbf{Rotate Angle Optimization}  \\
All the training images are augmented with each of the below angles and Maximum Validation accuracies are recorded accordingly. 
The angles used for experiments are  {2$^{\circ}$, 4$^{\circ}$, 8$^{\circ}$, 16$^{\circ}$, and 32$^{\circ}$} . The validation accuracy started to decrease after 8 $^{\circ}$. The best validation accuracy is achieved when angle is 8 $^{\circ}$. So the range considered is (-7,7). The plot is shown in figure \ref{fig:rotate}.

\begin{figure}	[hbt]
  \includegraphics[width=\columnwidth]{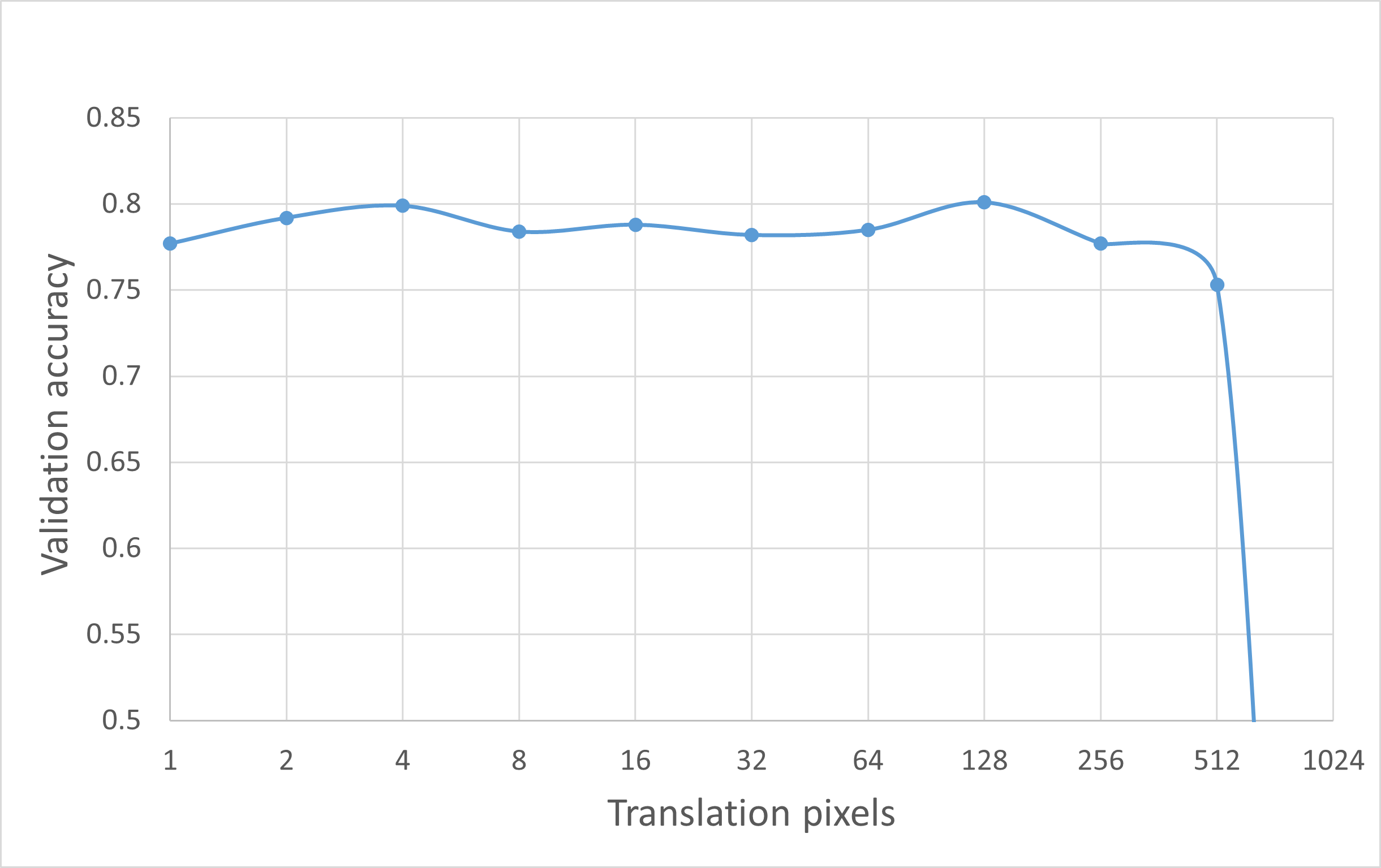}
  \caption{Translation pixels optimization.}
  \label{fig:translate}
\end{figure}

\noindent
\textbf{Translation Pixels Optimization}  \\
All the training images are horizontally translated with each of the below numbers of pixels and Maximum Validation accuracies are recorded accordingly. 
Pixels used for Translation are {2, 4, 6, 8, 16, 32, 64, 128, 256, 512, 800} . The best validation accuracy started to decrease after 128 pixles. The best validation accuracy is when the image is translated untill 128 pixels.  The plot is shown in figure \ref{fig:translate}. So the optimal range considered is (-20,20).

\noindent
\textbf {Probability Optimization} \\
\textbf{Custom Augmentation}
MMDetection does not support random uniform sampling. So to support random uniform sampling, custom classes for shear, rotate and translate were implemented. For each augmentation, a random uniform value is selected from the optimal ranges for shear, rotate, and translation. The value is made negative with half the probability.

To find the optimal probability, five different augmentations are used. The order of these augmentations are listed in table \ref{Augmentations used for probability study}.
\begin{table}[hbt]
  \centering
  \caption{{Augmentations used for probability study }}
  \label{Augmentations used for probability study}
  \begin{tabular}{ |l | l| }
    \hline
    Data augmentation method & Parameter range \\
    \hline 
    H-flip & \\
    Rescale & [0.8,1.2]  \\
    Shear & [-3,3] \\ 
    Rotate & [-7,7]\\
    Translate & [-20,20]\\
    \hline
  \end{tabular}	
\end{table}

\begin{figure}	[hbt]
	\includegraphics[width=\columnwidth]{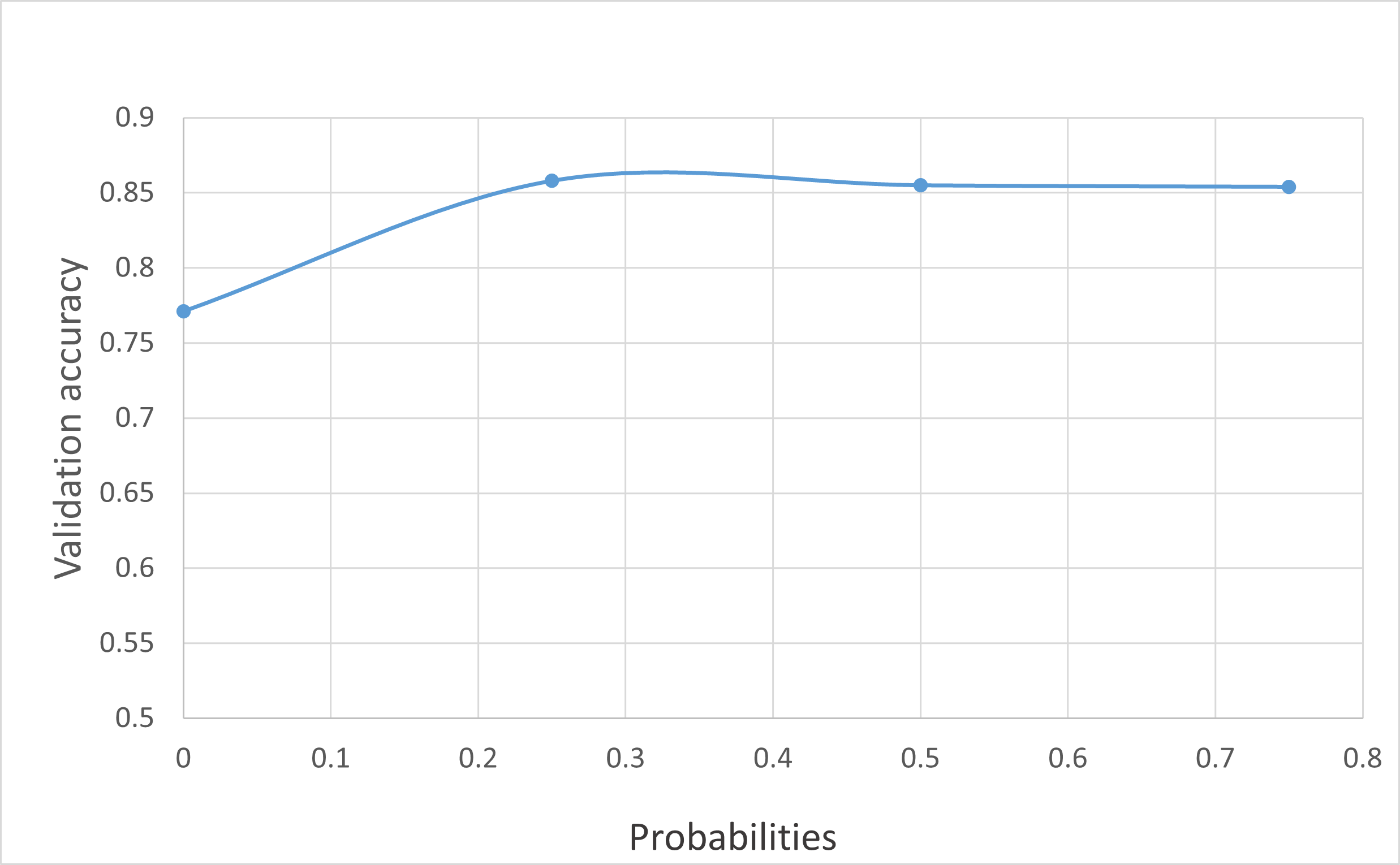}
	\caption{Validation Accuracy with Probability = 0.5.}
	\label{fig:probability study}
\end{figure}

The probabilities considered for the experiments are P =\{0, 0.25, 0.5, 0.75, 1.0\}. The figure for all the probabilities considered is shown in Fig. \ref{fig:probability study}. The best validation accuracy is with probability 0.5. 

\begin{figure}	[hbt]
  \includegraphics[width=\columnwidth]{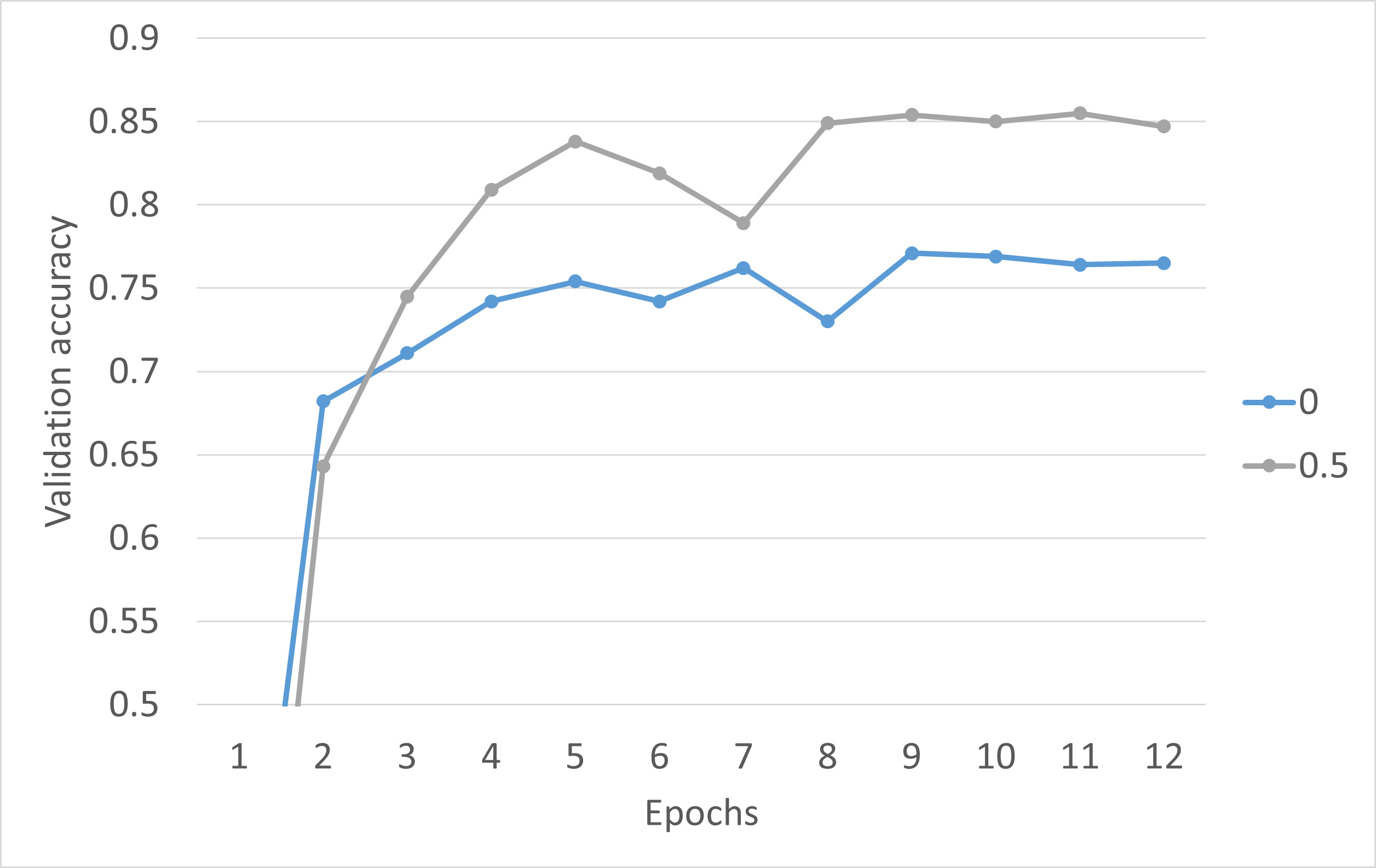}
  \caption{Probability Study.}
  \label{fig:opt_res}
\end{figure}

Now that all the optimal parameters are obtained, model is trained with the optimized dataset, the improvement of validation accuracies of no-augmentation vs all the five augmentation applied with 0.5 probability is shown in Fig. \ref{fig:opt_res}

\subsection{Improvement of Hand Detection Results using Optimal Data Augmentation} 
 An AP of 0.81 was achieved by training the model with optimized data augmentation parameters. The training time remains the same as the size of dataset is not changed.
 \comment{
\begin{table*}[hbt]
  \centering
  \caption{Improvement of hand detection results using data augmentation}
  \label{results:hand_det_with_aug}
  \begin{tabular}{|l|l|}
    \hline
    \textbf{Metric} & \textbf{Value}\\
    \hline
    \hline
    AP @[IOU=0.50:0.95 |area = all | maxDets = 100]     &    0.299\\
    AP @[IOU=0.50 |area = all | maxDets = 100] 	        & 	\textbf{0.801}\\
    AP @[IOU=0.75 |area = all | maxDets = 100]		    &	0.107\\
    AP @[IOU=0.50:0.95 |area = small | maxDets = 100]   & 	0.136\\
    AP @[IOU=0.50:0.95 |area = medium | maxDets = 100]  & 	0.312\\
    AP @[IOU=0.50:0.95 |area = large | maxDets = 100]   & 	0.350\\
    AR @[IOU=0.50:0.95 |area = all | maxDets = 1] 		& 	0.404\\
    AR @[IOU=0.50 |area = all | maxDets = 10] 			& 	0.404 \\
    AR @[IOU=0.75 |area = all | maxDets = 100]			& 	0.404\\
    AR @[IOU=0.50:0.95 |area = small | maxDets = 100]   & 	0.298\\
    AR @[IOU=0.50:0.95 |area = medium | maxDets = 100]  & 	0.423\\
    AR @[IOU=0.50:0.95 | area = large | maxDets = 100]  & 	0.350\\
    \hline
  \end{tabular}
\end{table*}
}
Table \ref{tab:aug_res} summarizes the final results of no augmentation vs optimal augmentation. An improvement of AP of  11\% and 9\% on the validation and testing set, respectively, was achieved. Here the word best means the epoch, which gave maximum validation accuracy. 
\begin{table}[hbt]
  \centering
  \caption{Validation and Testing Accuracies for multiple probabilities.}
  \label{tab:aug_res} 
  \begin{tabular}{ |l | l| l | l| l| l| l| }
    \hline
    \textbf{Data Split} & \textbf{Model} & \textbf{No-aug} & \textbf{P=0.25} & \textbf{P =0.5} & \textbf{P=0.75} & \textbf{P=1.0}\\
    \hline 
    \textbf{Val} & Best & 0.77 & 0.86 & \textbf{0.86} & 0.85 & 0.84 \\ 
    & Last & 0.76 & 0.85 & \textbf{0.86} & 0.84 & 0.82 \\
    \textbf{Test} & Best & 0.75 & 0.80 & \textbf{0.80} & 0.79 & 0.78 \\
    & Last & 0.71 & 0.80 & \textbf{0.81} & 0.78 & 0.76 \\
    \hline
  \end{tabular}	
\end{table}

\begin{figure}[!h]
  \begin{subfigure}{0.49\columnwidth}
    \centering
    \includegraphics[width=\textwidth]{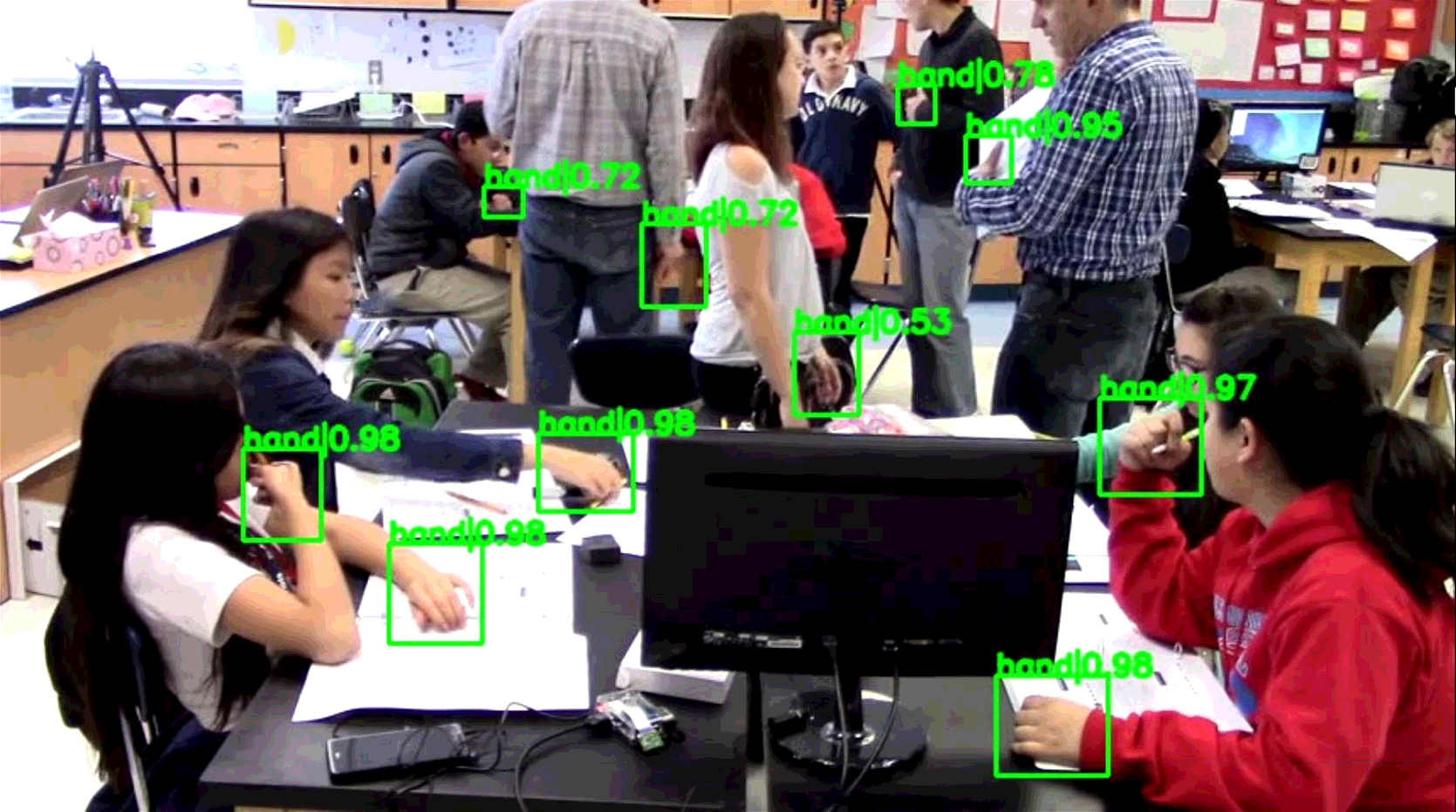}
    \caption{}
  \end{subfigure}\hfill
  \begin{subfigure}{0.49\columnwidth}
    \centering
    \includegraphics[width=\textwidth]{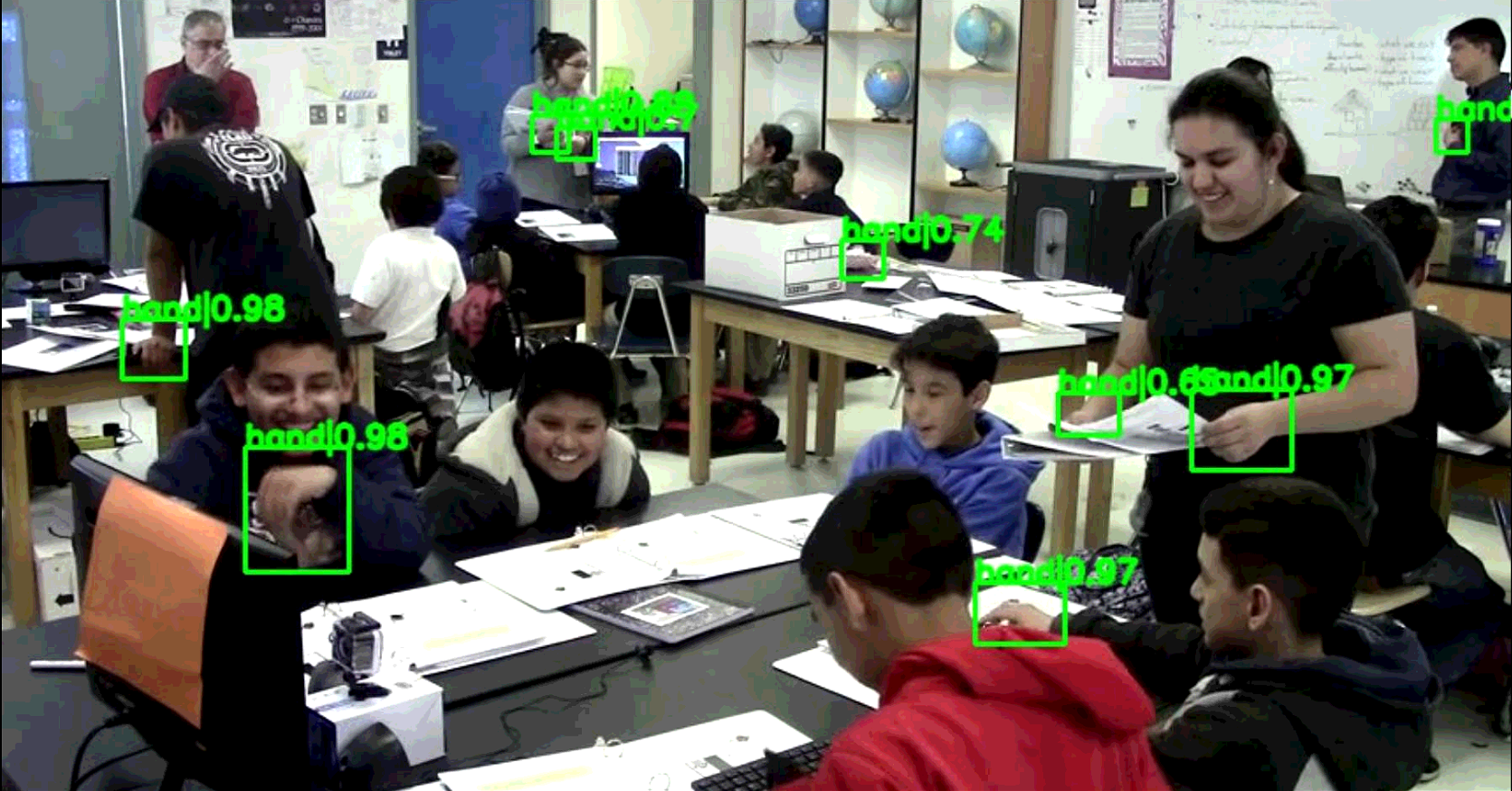}
    \caption{}
    \label{fig:}
  \end{subfigure}
  \begin{subfigure}{0.49\columnwidth}
    \centering
    \includegraphics[width=\textwidth]{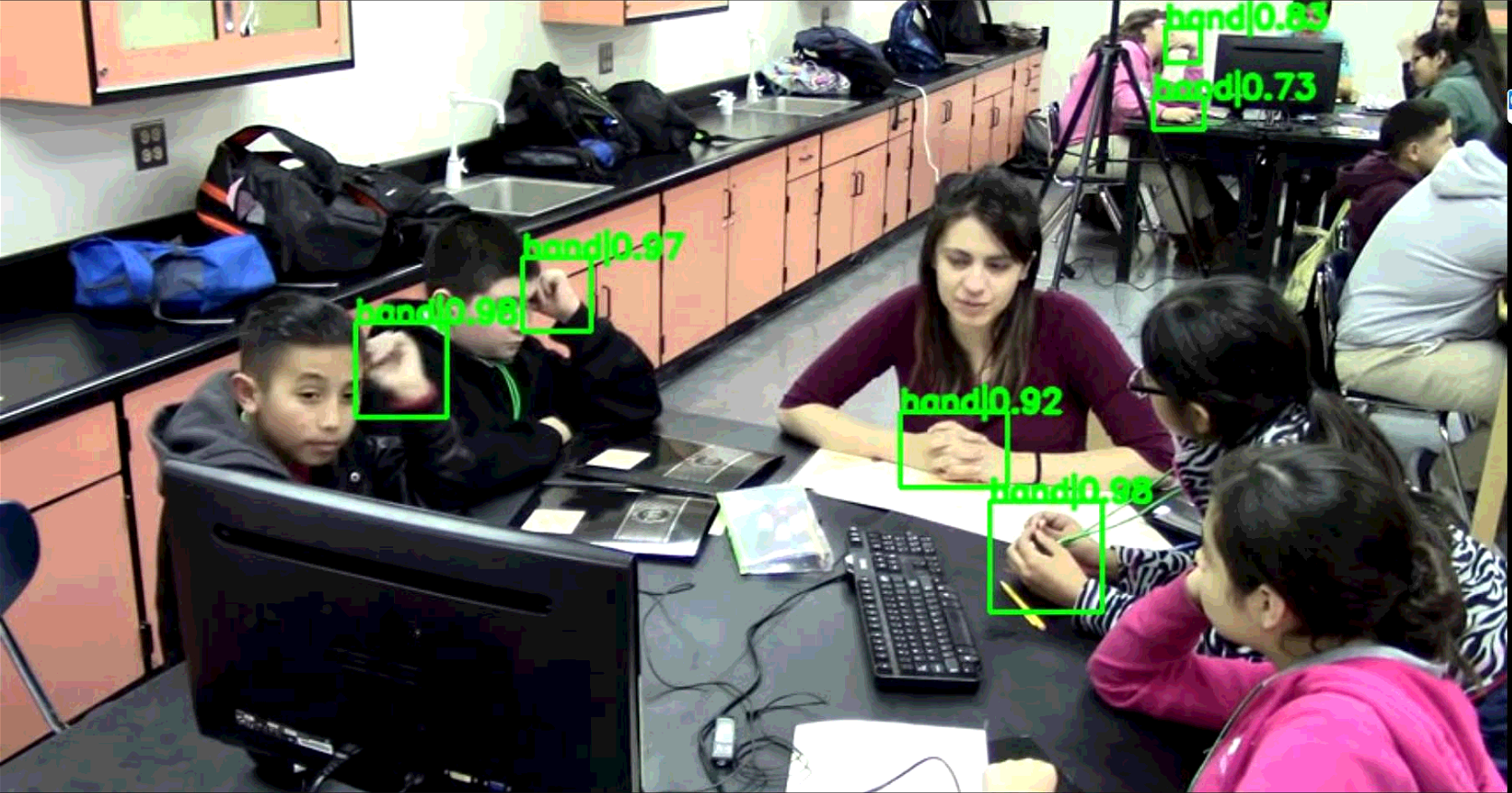}
    \caption{}
    \label{fig:}
  \end{subfigure}
  \begin{subfigure}{0.49\columnwidth}
    \centering
    \includegraphics[width=\textwidth]{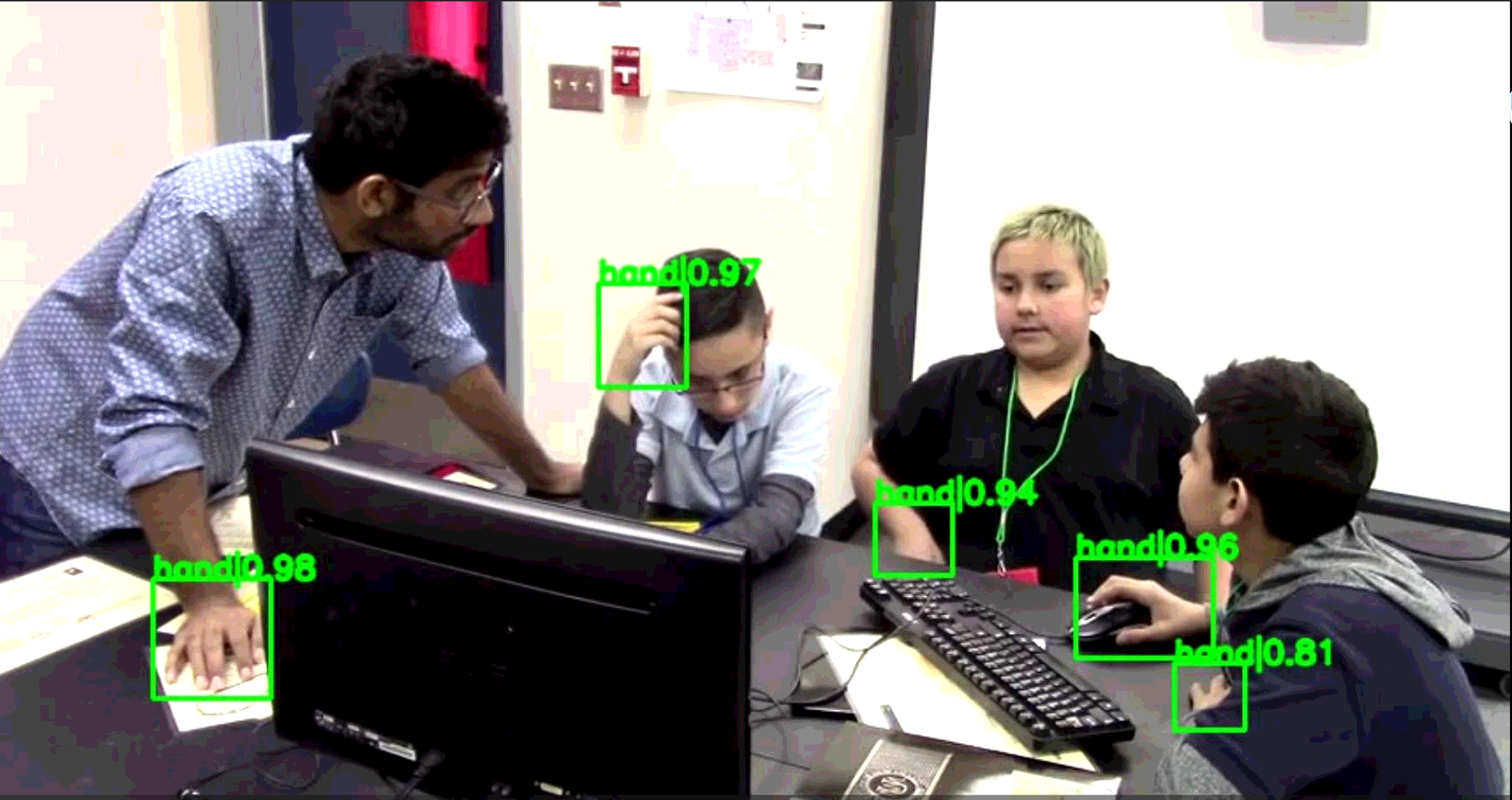}
    \caption{}
    \label{fig:}
  \end{subfigure}
  \begin{subfigure}{0.49\columnwidth}
    \centering
    \includegraphics[width=\textwidth]{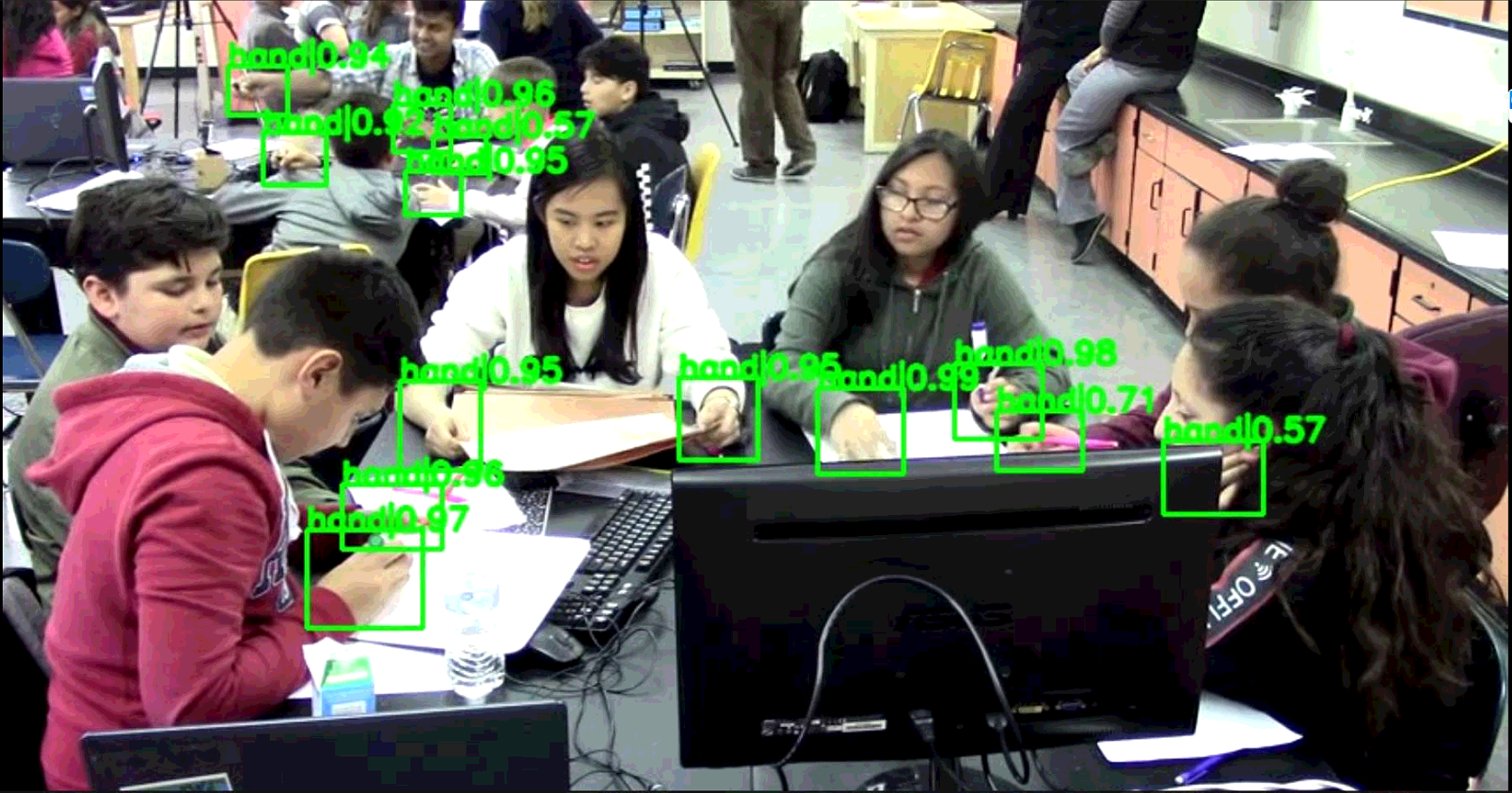}
    \caption{}
    \label{fig:}
  \end{subfigure}
  \begin{subfigure}{0.49\columnwidth}
    \centering
    \includegraphics[width=\textwidth]{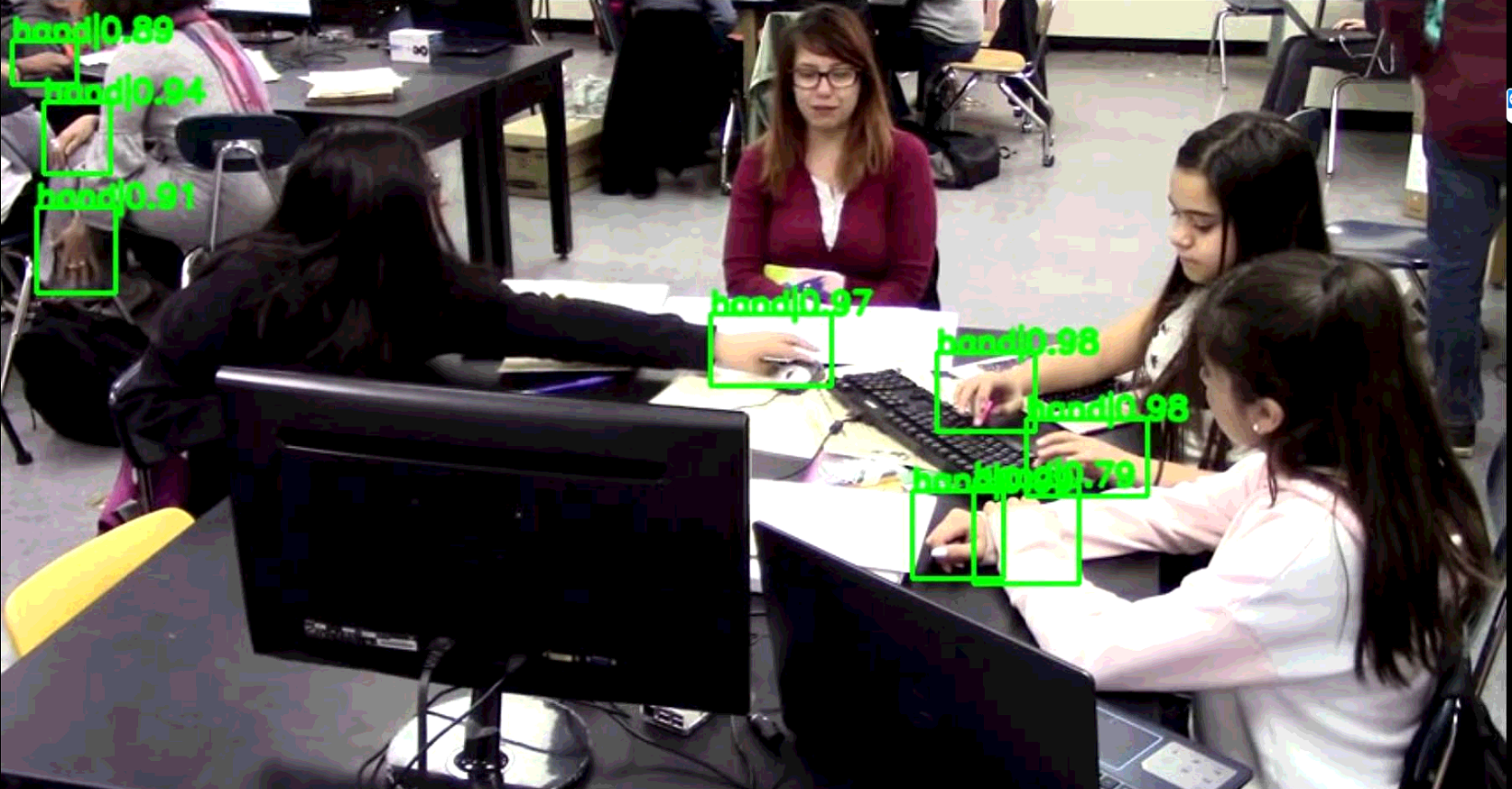}
    \caption{}
    \label{fig:}
  \end{subfigure}
  \caption{Examples of hand detection.}
  \label{fig:hand_det_results}
\end{figure}

Fig. \ref{fig:hand_det_results} shows some of the testing results for hand detection.

As seen in Fig. \ref{fig:hand_det_results}, it is evident that there are many hands detected in the picture. The main goal here is to propose activity regions for writing within the primary table of focus; we need to eliminate the background hands. The other goal is to reduce the number of regions proposed to classify.

\subsection{Performance Evaluation Protocol of Using Projections}

\begin{figure}	[hbt]
  \includegraphics[width=\columnwidth]{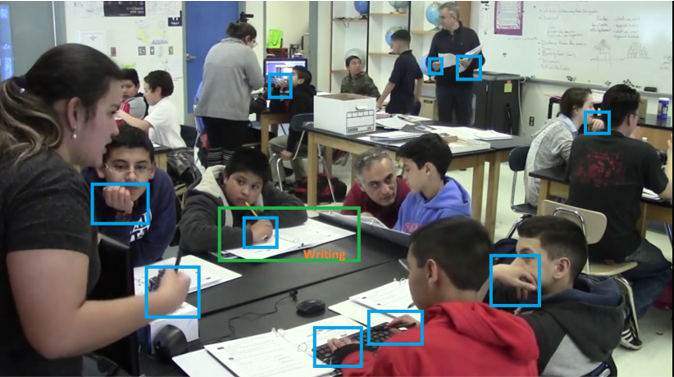}
  \caption{Calculation of performance evaluation.}
  \label{fig:pf}
\end{figure}

Fig. \ref{fig:pf} shows how the performance evaluation is performed.
The boxes marked in blue are results of hand detection and the green box represents ground truth for writing instances.

\begin{figure}[ht!]
  \centering
  \begin{subfigure}{0.49\columnwidth}
    \centering
    \includegraphics[width=\textwidth]{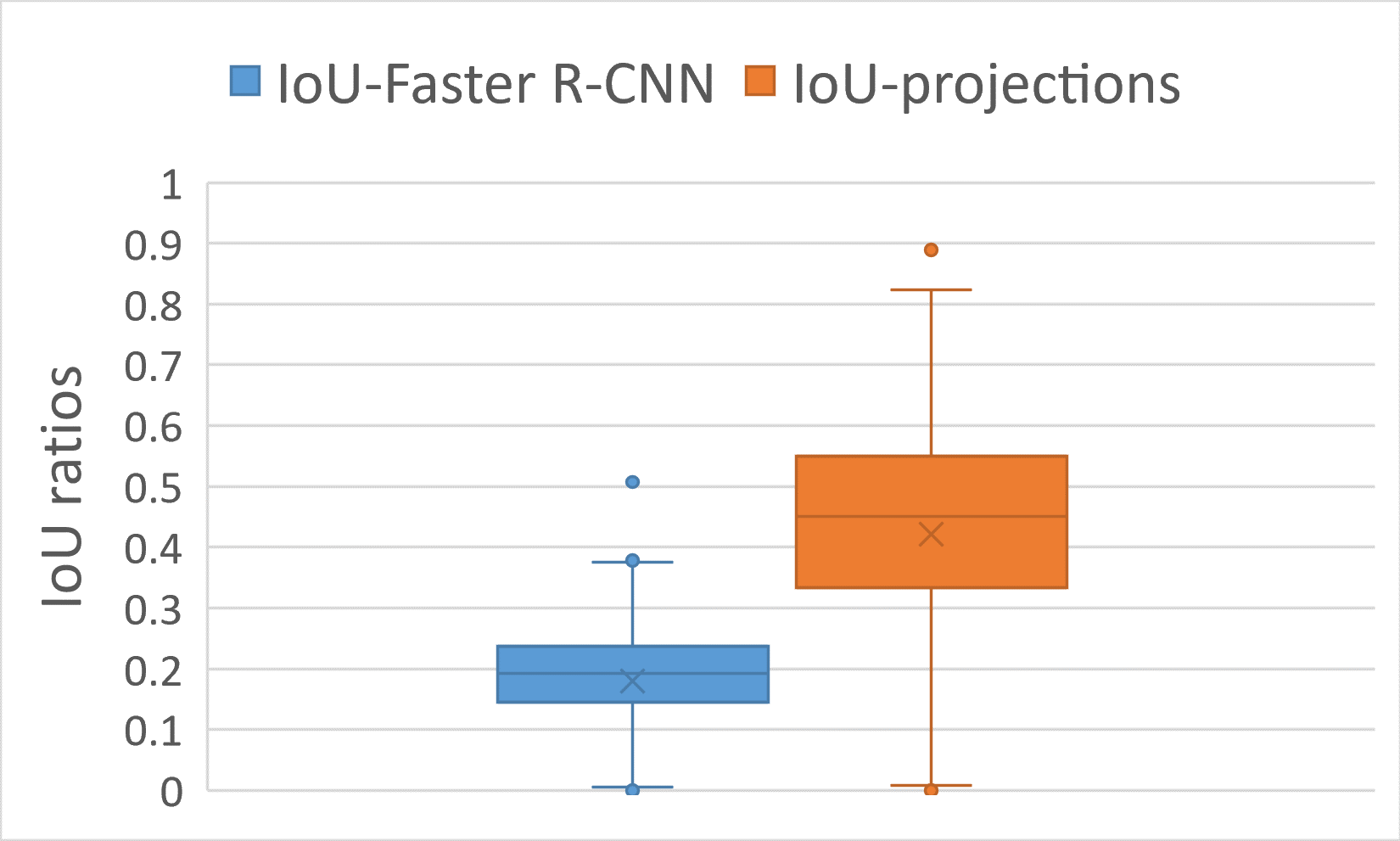}
    \caption{C1L1P-C-Mar30}
  \end{subfigure}\hfill
  \begin{subfigure}{0.49\columnwidth}
    \centering
    \includegraphics[width=\textwidth]{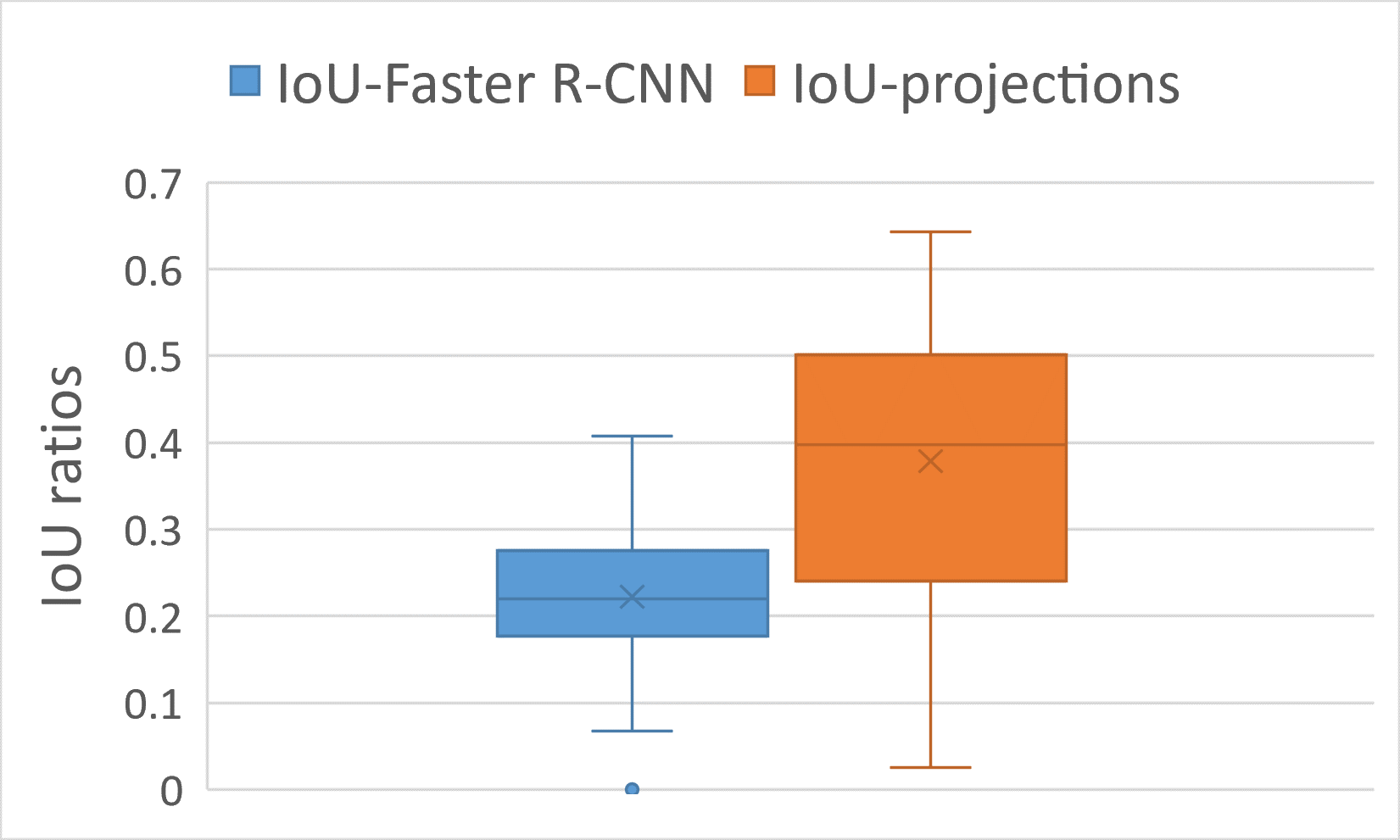}
    \caption{C1L1P-C-Apr13}
  \end{subfigure}
  \begin{subfigure}{0.49\columnwidth}
    \centering
    \includegraphics[width=\textwidth]{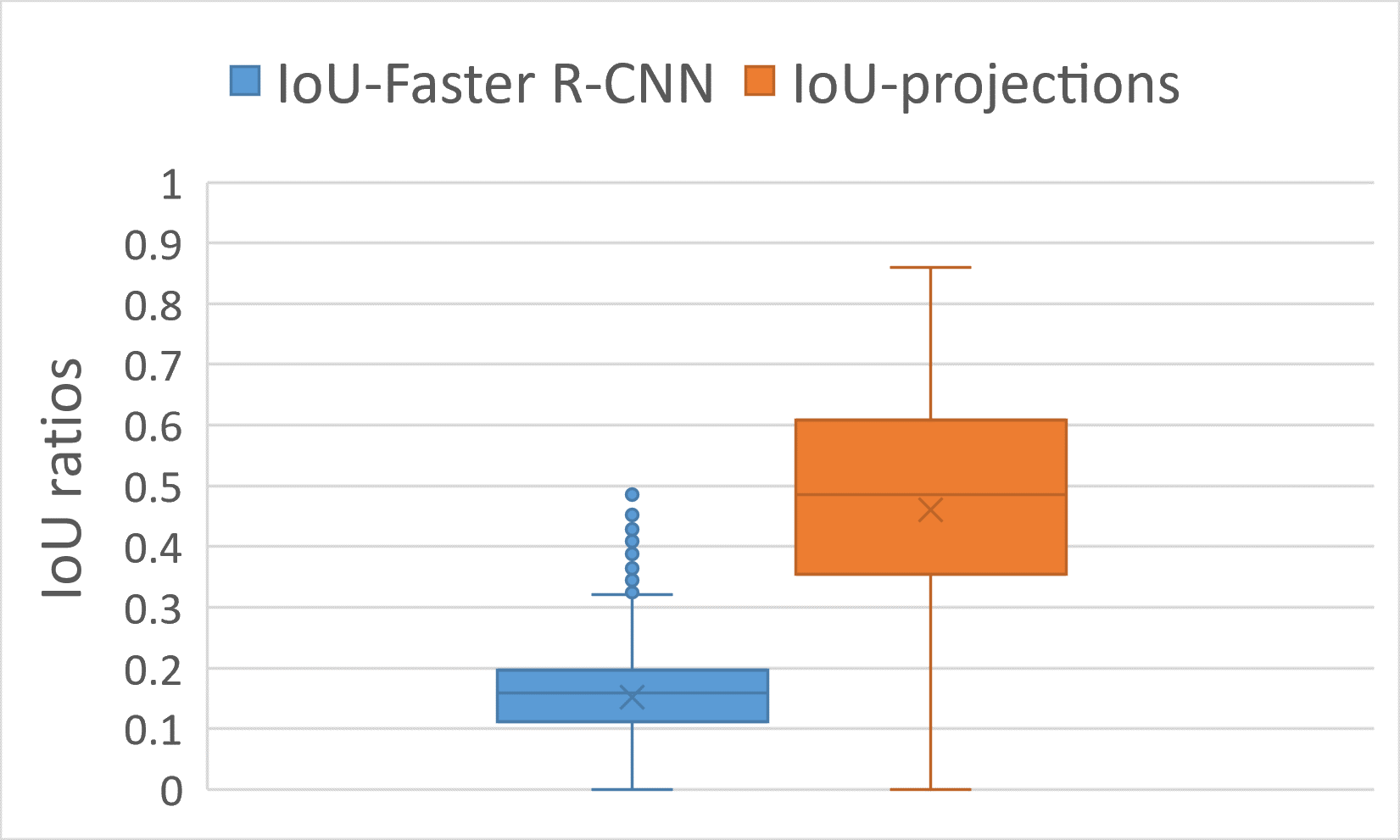}
    \caption{C1L1P-E-Mar02}
  \end{subfigure}\hfill
  \begin{subfigure}{0.49\columnwidth}
    \centering
    \includegraphics[width=\textwidth]{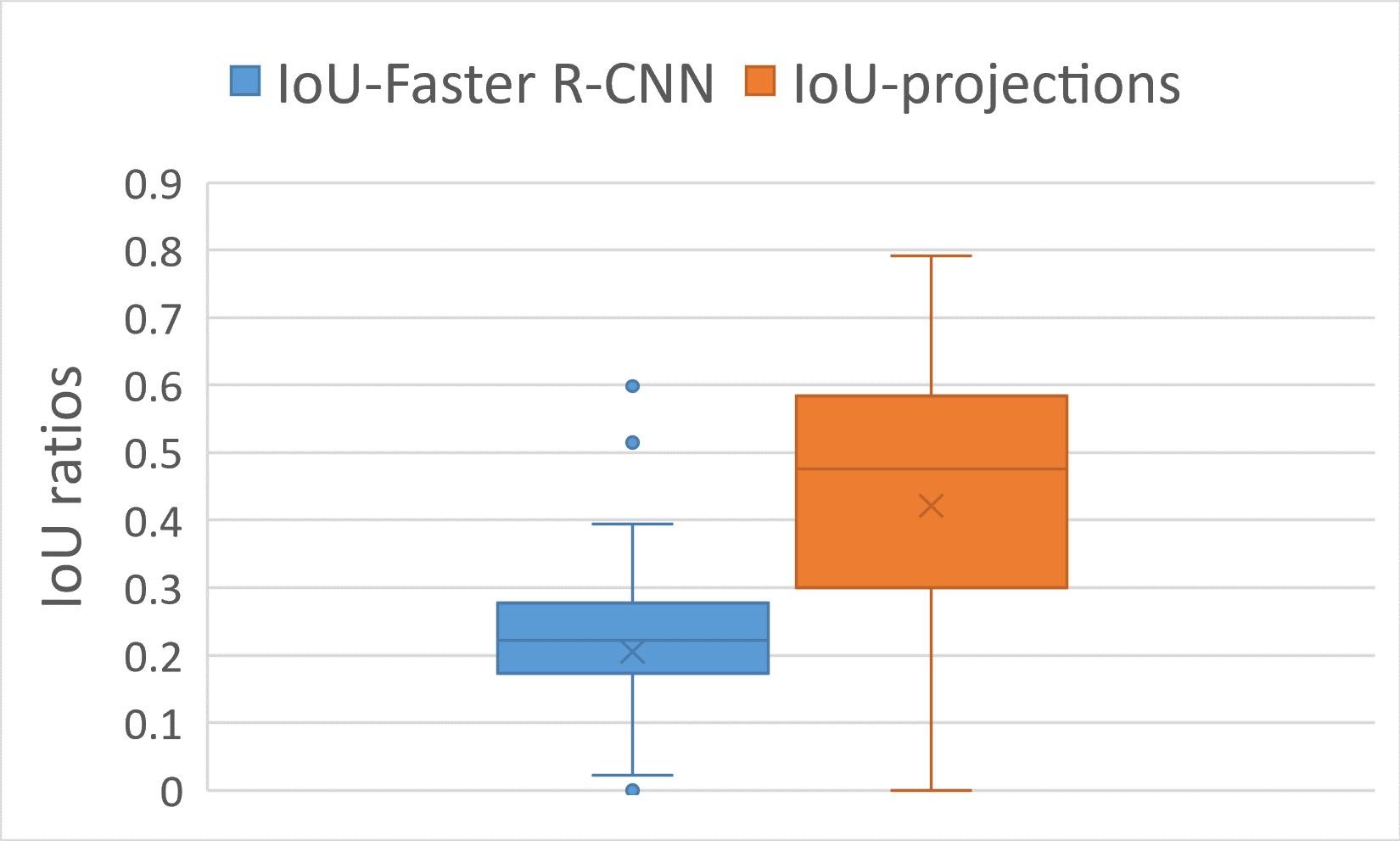}
    \caption{C2L1P-B-Feb23}
    \label{fig:time2}
  \end{subfigure}
  \begin{subfigure}{0.49\columnwidth}
    \centering
    \includegraphics[width=\textwidth]{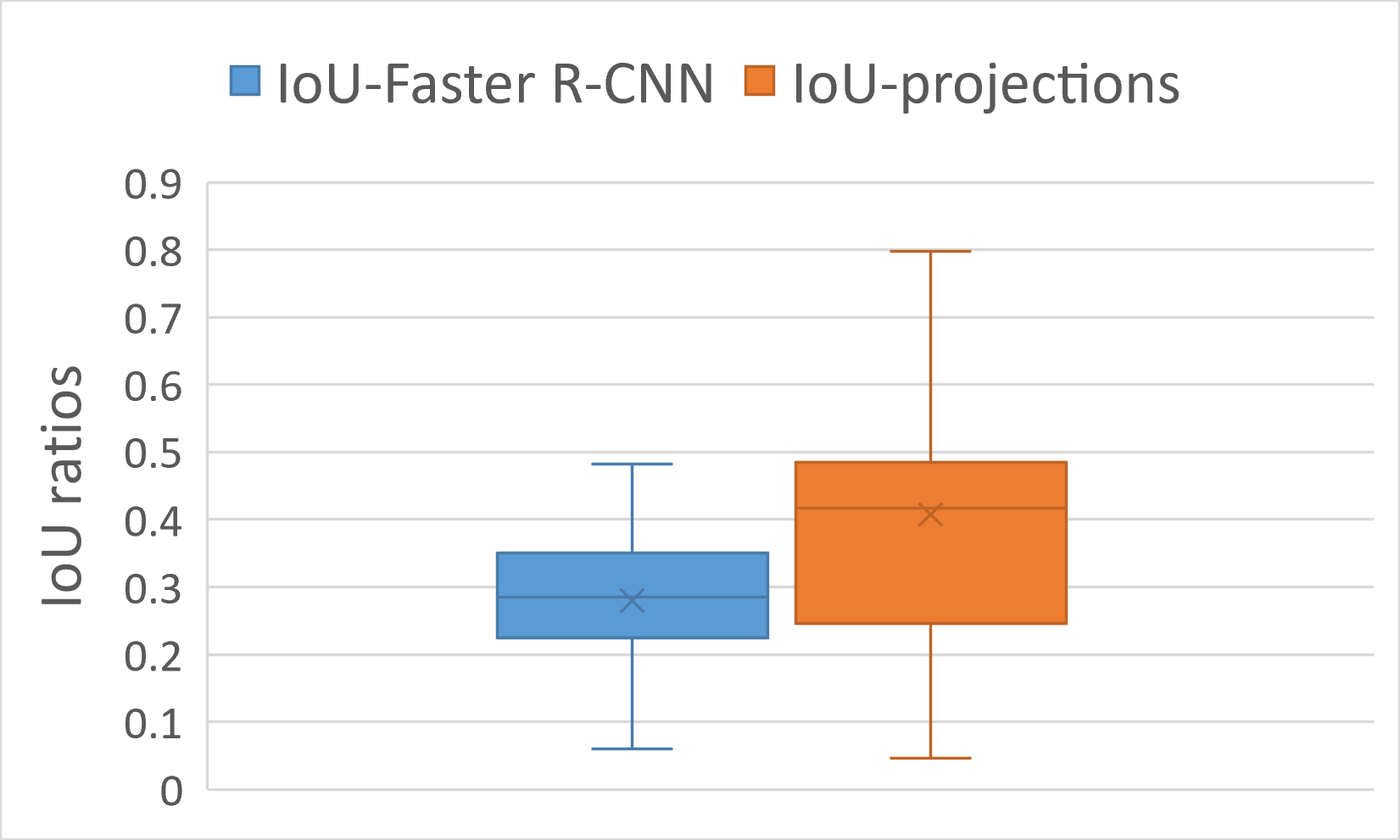}
    \caption{C2L1P-D-Mar08}
  \end{subfigure}\hfill
  \begin{subfigure}{0.49\columnwidth}
    \centering
    \includegraphics[width=\textwidth]{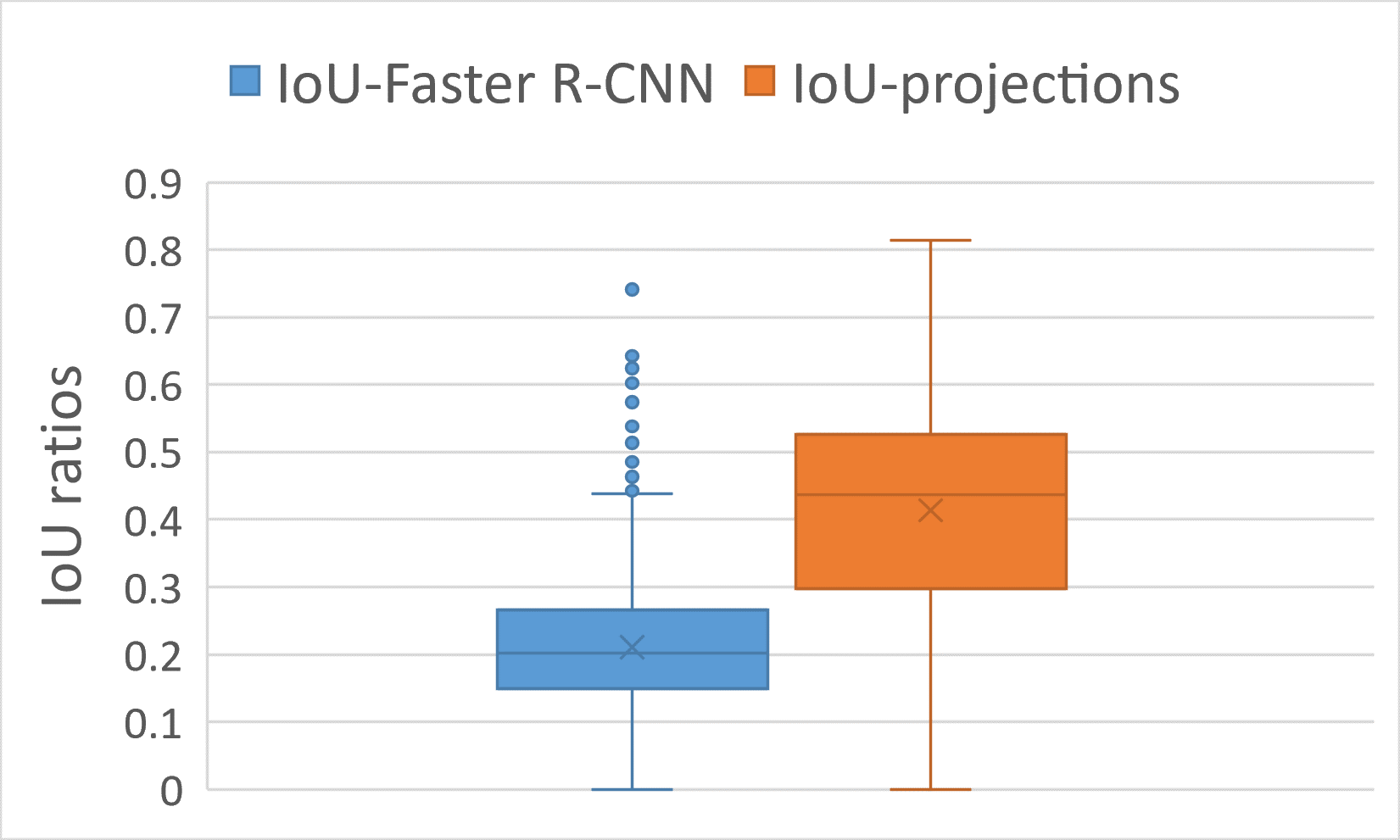}
    \caption{C3L1P-C-Apr11}
  \end{subfigure}	
  \begin{subfigure}{0.49\columnwidth}
    \centering
    \includegraphics[width=\textwidth]{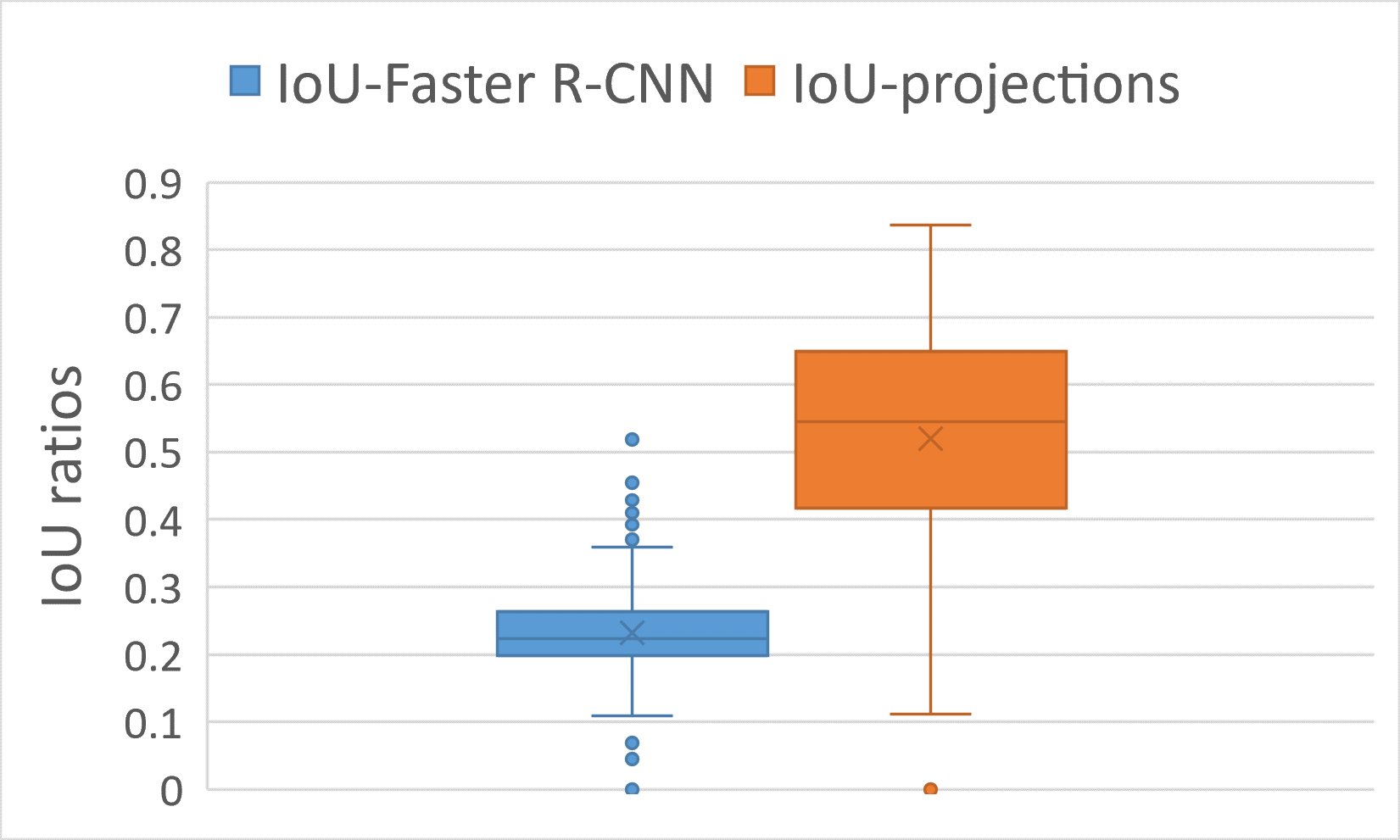}
    \caption{C3L1P-D-Feb21}
  \end{subfigure}\hfill
  \caption{Performance evaluation for all the test sessions.}
  \label{fig:pf_test_sessions}
\end{figure}

Steps involved in performance evaluation are:
\begin{itemize}
\item Classify each second as writing instance if more than half of the frames are labeled as writing in ground Truth.
\item Take all the hand proposals.
\item Calculate IoU ratio for each proposal region w.r.t writing ground truth.
\item Take the value which corresponds to maximum IoU ratio. 
\end{itemize}
Once all the regions and corresponding IoU ratios are obtained, they are plotted using box plot  as shown in figure \ref{fig:pf_test_sessions}.

\noindent\textbf{Naive Region Proposals:} \\
Naive region proposal approach is considered to take all the hand detections into consideration.

\begin{figure}[ht!]	
  \begin{subfigure}{0.49\columnwidth}
    \centering
    \includegraphics[width=\textwidth]{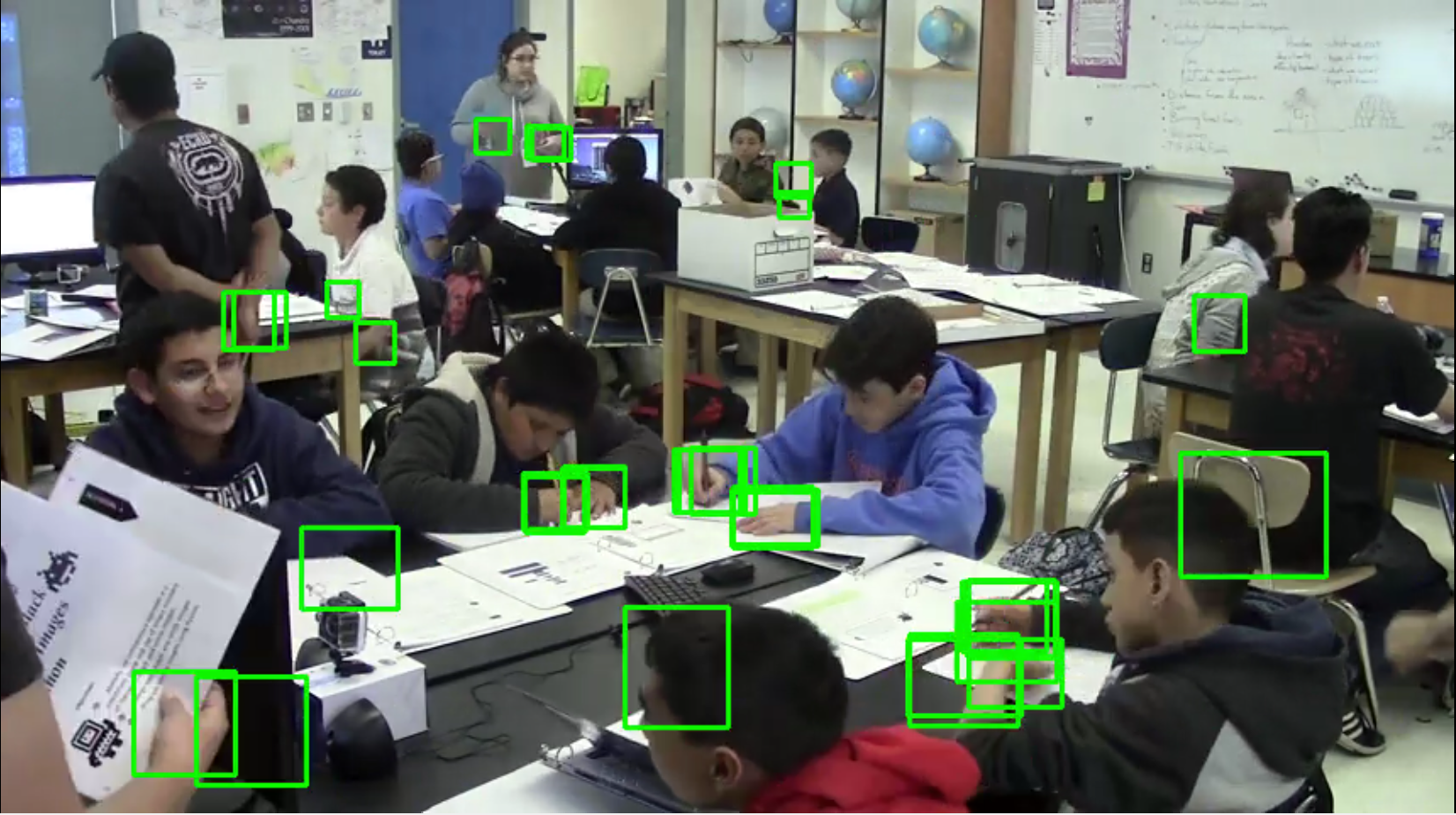}
    \caption{}
  \end{subfigure}\hfill
  \begin{subfigure}{0.49\columnwidth}
    \centering
    \includegraphics[width=\textwidth]{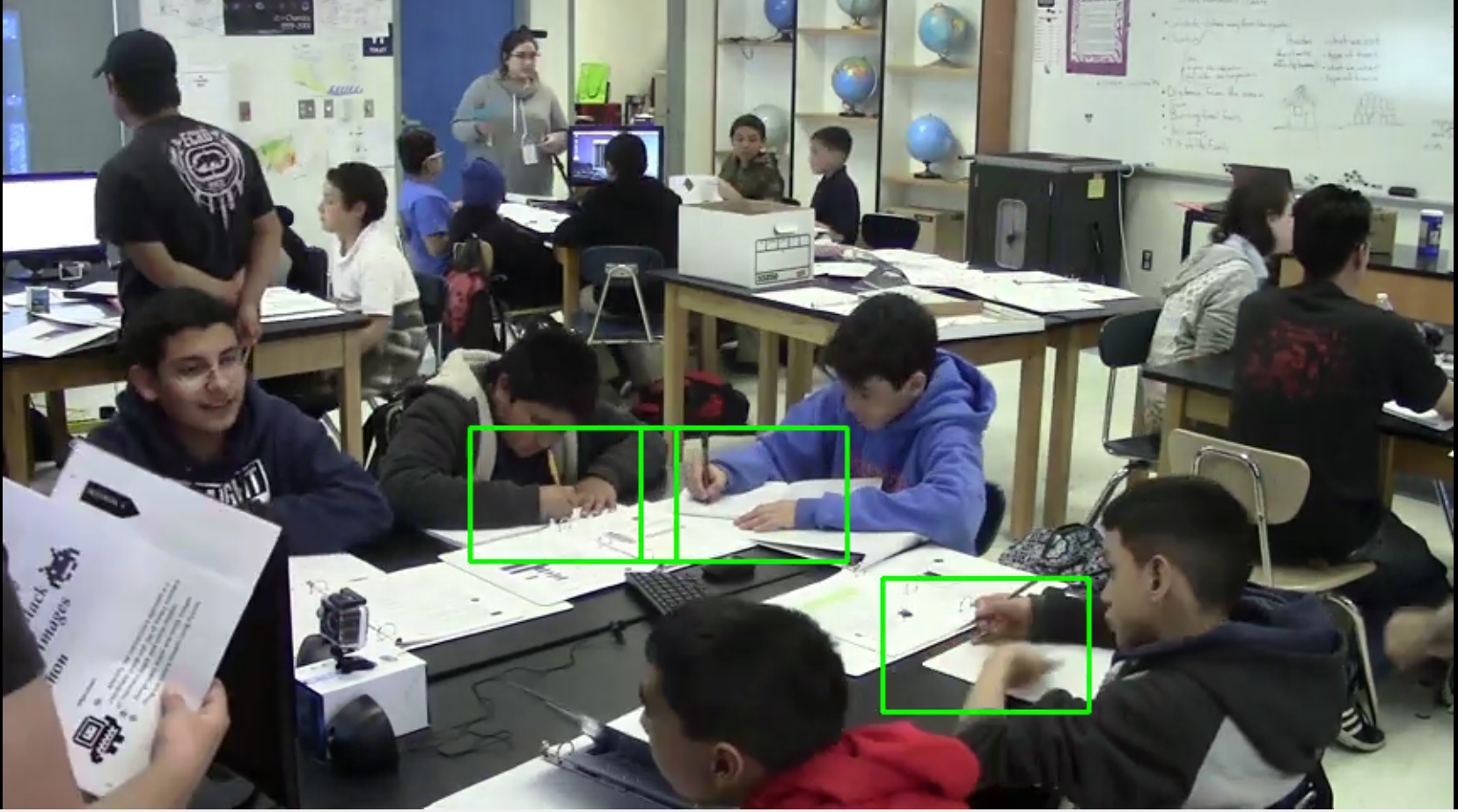}
    \caption{}
  \end{subfigure}
  \begin{subfigure}{0.49\columnwidth}
    \centering
    \includegraphics[width=\textwidth]{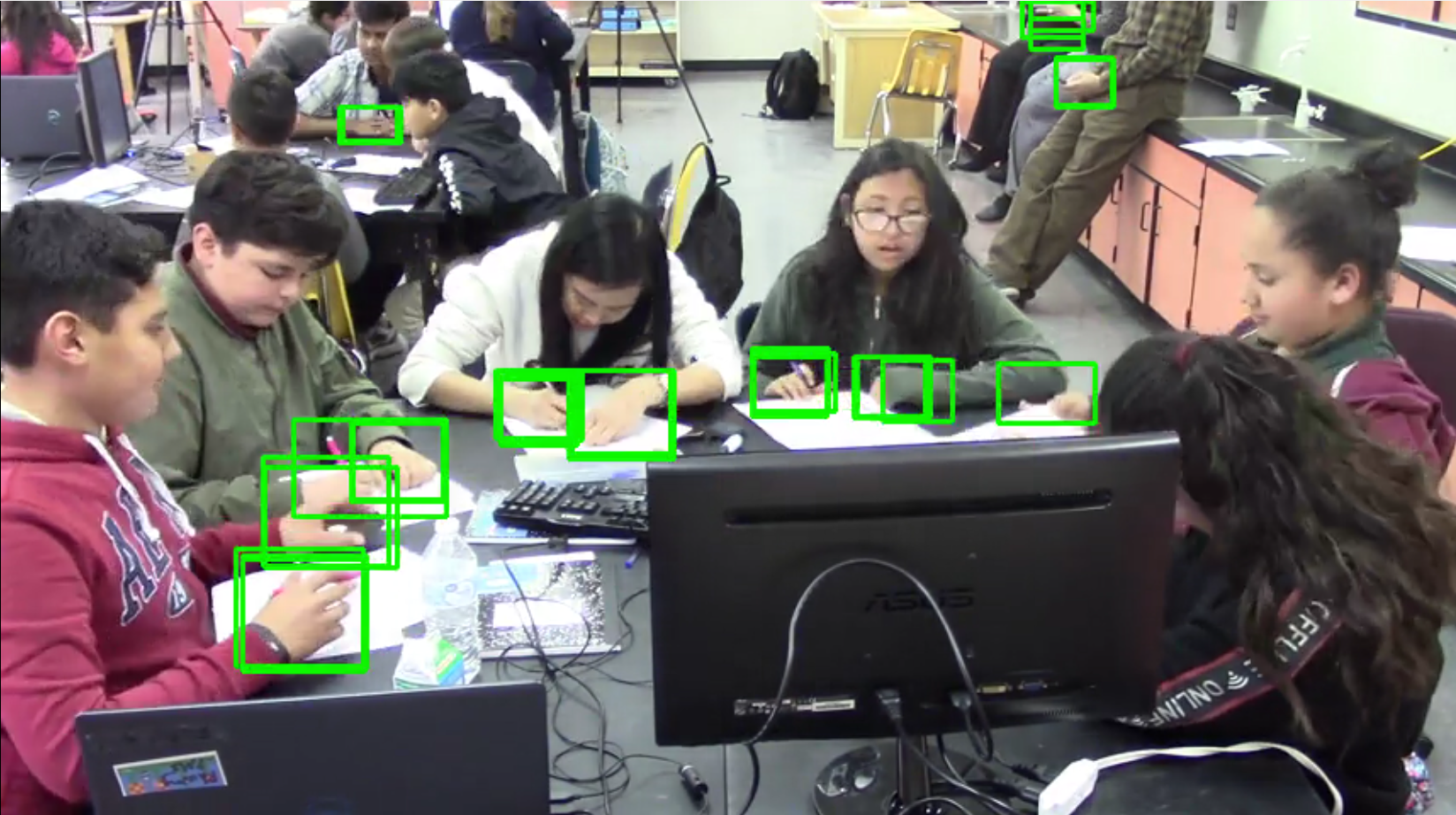}
    \caption{}
    \label{fig:time1}
  \end{subfigure}\hfill
  \begin{subfigure}{0.49\columnwidth}
    \centering
    \includegraphics[width=\textwidth]{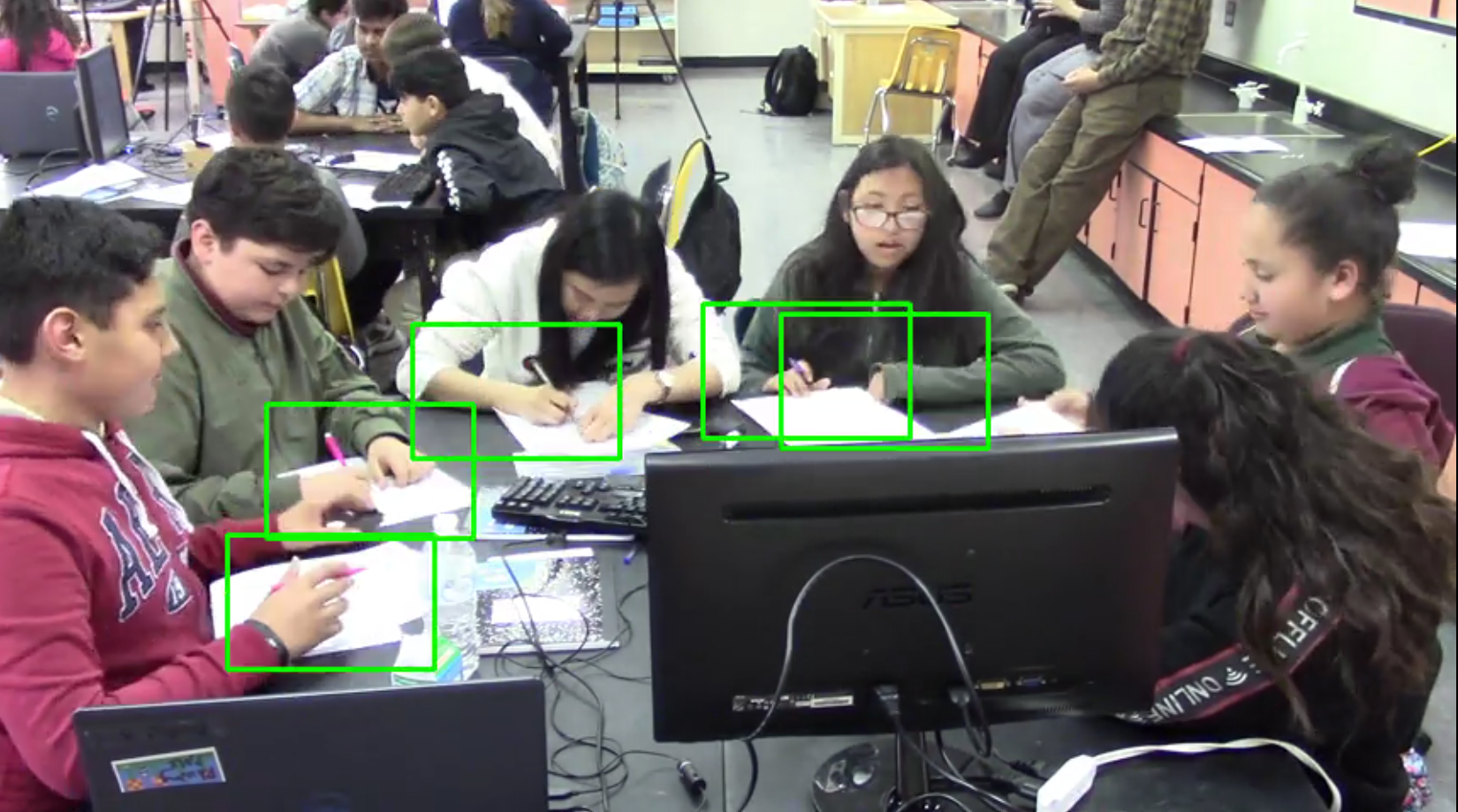}
    \caption{}
  \end{subfigure}
  \begin{subfigure}{0.49\columnwidth}
    \centering
    \includegraphics[width=\textwidth]{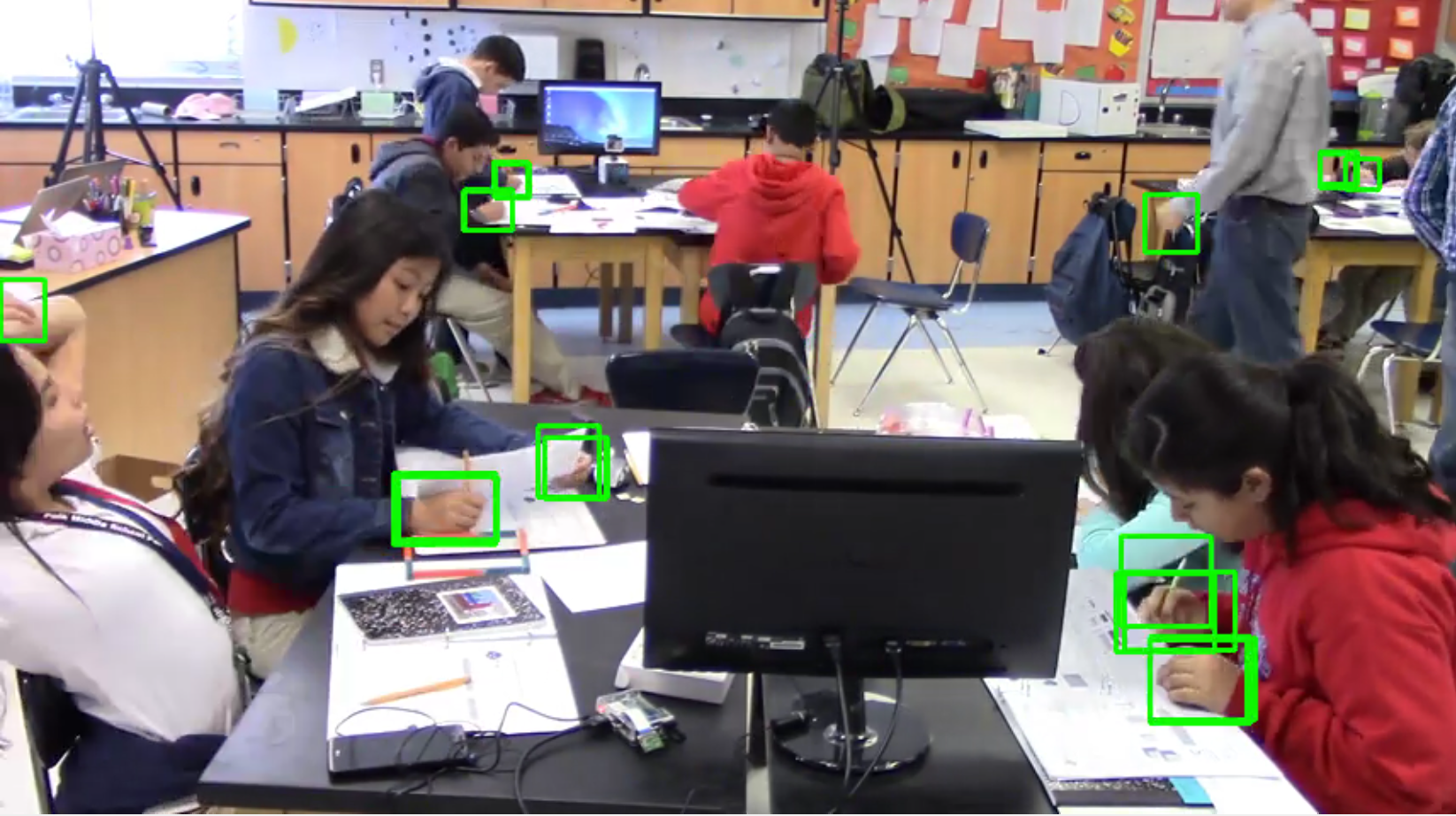}
    \caption{}
  \end{subfigure}\hfill
  \begin{subfigure}{0.49\columnwidth}
    \centering
    \includegraphics[width=\textwidth]{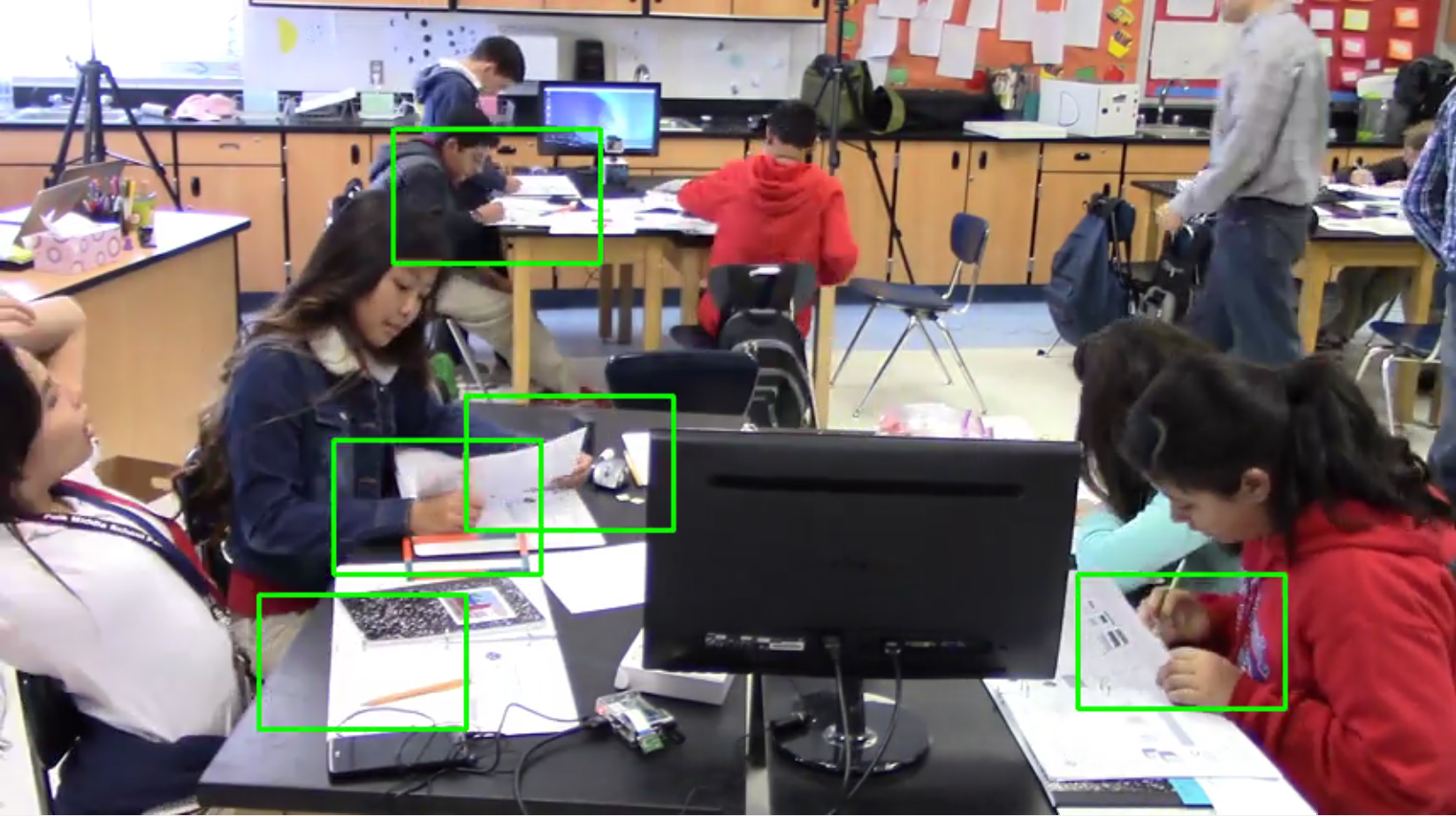}
    \caption{}
  \end{subfigure}	
  \caption{Results showing the reduction of proposal regions and capturing writing instances. The ones on the left are original and the ones on right are after projections and segmentation.}
  \label{fig:vid_res}	
\end{figure}

Figure \ref{fig:vid_res} shows the results on video frames. The figures on the left use the Naive approach, and the ones on the right use Projections. These test frames clearly show the reduction in the number of regions and capturing all writing instances.

Figure \ref{fig:pf_test_sessions} shows the performance of two approaches i) Naive and ii) Projections for all the test sessions. IoU ratios are calculated for every writing instance with the proposals generated by the Naive method and Projections. The approach was very effective in removing false positive detections that correspond to hands from a different group. Yet, the approach was able to detect all the hands from the collaborative group that was closer to the camera (as required). 

Table \ref{tab:pf} shows the percentage of reduction in the number of proposal regions for the two approaches for all the test sessions. The projection approach was able to achieve an average of 80\% reduction in the proposed activity regions and also capture all the instances of collaborative group shown in Fig. \ref{fig:pf_test_sessions}.

\begin{table}[ht!]
  \centering
  \caption{Reduction in number of region proposals for each test session.}
  \label{tab:pf}
  \begin{tabular}{ | p {2cm} | p{2cm} | p{2cm}| p{2cm}| p{2cm} |}
    \hline
    Group & Date & Naive & Using Projections & 	\% Reduction \\
    \hline 
    C1L1P-C & Mar30 & 55914& 9804  & \textbf{82.5}\\ 
    C1L1P-C	& Apr13 & 34665 & 8028 & \textbf{76.8}\\
    C1L1P-E & Mar02 & 50312 & 9968 & \textbf{80.0}\\
    C2L1P-B & Feb23 & 48073 & 9924  & \textbf{79.3}\\
    C2L1P-D & Mar08 & 31875 & 7724 & \textbf{75.7}\\
    C3L1P-C	& Apr11 & 36757 & 9536 &  \textbf{74.0}\\
    C3L1P-D & Mar19 & 57319 & 9536 &  \textbf{83.3}\\
    
    \hline
  \end{tabular}	
\end{table}

\chapter{Conclusion and Future Work}
\section{Conclusion} 
In this thesis, a new method was proposed which uses projections and cluster-based segmentation to identify potential activity regions, which include hand movements. The primary contributions of this thesis include: (i) Robust detection and fast-tracking of keyboard, (ii) a detailed optimal data augmentation parameter study showing an improvement over detection results, and (iii) a novel method that uses projections and cluster-based segmentation to identify hand regions of the current collaborative group.  In each case, the combined algorithm of detection and fast-tracking achieved almost the same accuracy with a lot more speed up of time. The model trained with optimal data augmentation parameters achieved an improvement of 8\% over detection results. Furthermore, the novel method was able to capture all the writing activity regions by reducing the number of region proposals by 80\%. \\

\section{Future Work}
\begin{figure}
  \begin{subfigure}{0.49\columnwidth}
    \centering
    \includegraphics[width=\textwidth]{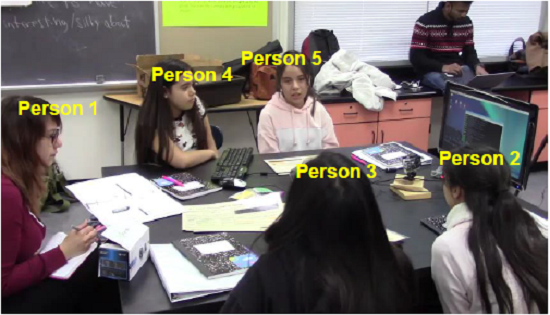}
    \caption{Frame extracted from student interactions.}
  \end{subfigure}	
  \begin{subfigure}{0.49\columnwidth}
    \centering
    \includegraphics[width=\textwidth]{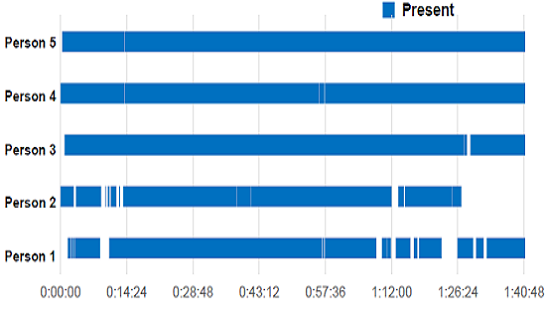}
    \caption{Student participation map.}
  \end{subfigure}
  \caption{Participation map for a session.}
  \label{fig:futurework1}
\end{figure}

\begin{figure}
  \begin{subfigure}{0.49\columnwidth}
    \centering
    \includegraphics[width=\textwidth]{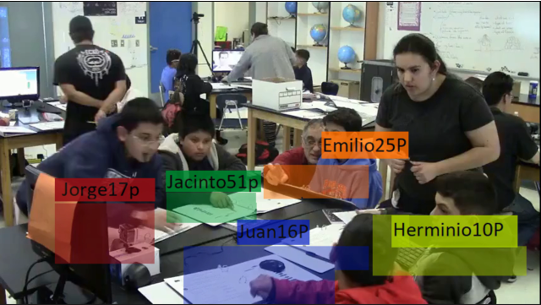}
    \caption{student interaction.}
  \end{subfigure}	
  \begin{subfigure}{0.49\columnwidth}
    \centering
    \includegraphics[width=\textwidth]{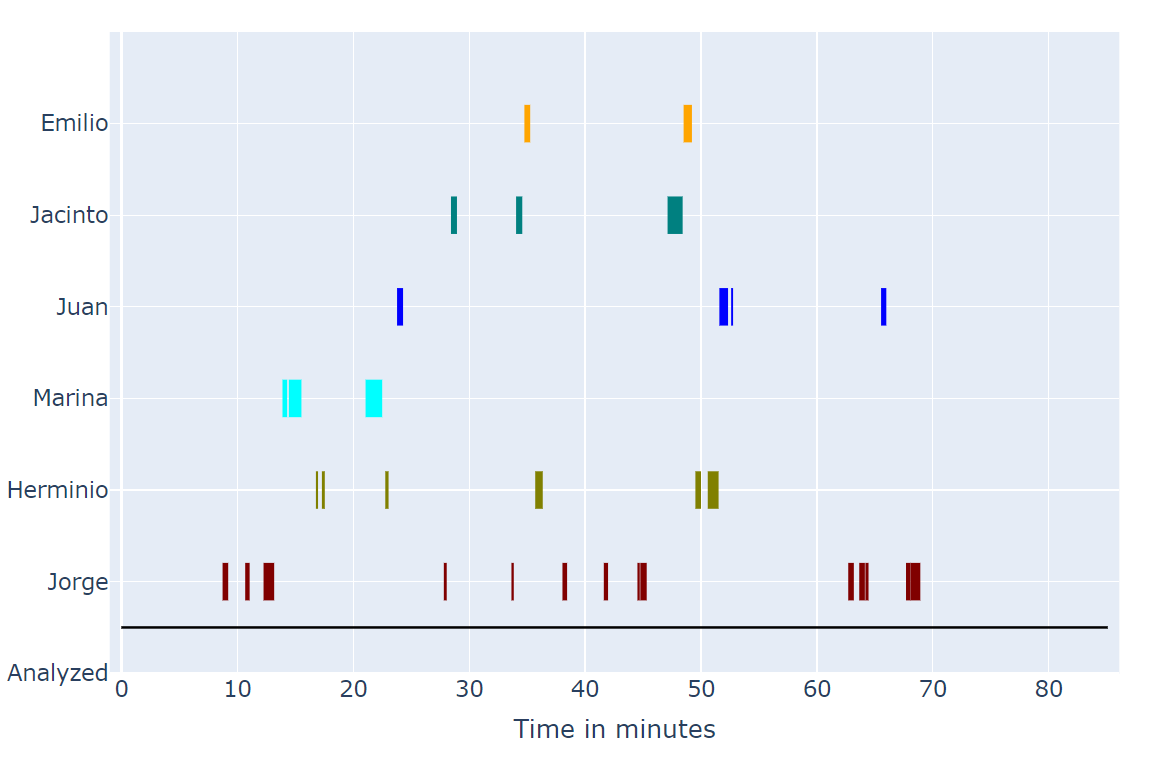}
    \caption{typing activity map for one session.}
  \end{subfigure}
  \caption{Session activity map.}
  \label{fig:futurework2}
\end{figure}

Using the proposed methods, we found improvements of: 
\begin{itemize}
\item \textbf{Matching hand regions to students}. The hand activity clusters can be matched to students in a session to generate student participation maps as shown in \ref{fig:futurework1}
\item \textbf{Region Proposals for Activity Recognition}. This research can be used to generate region proposals for activities such as typing and writing. It can be easily integrated with activity recognition systems to produce student activity maps as shown in Fig. \ref{fig:futurework2}.
\end{itemize}

\addcontentsline{toc}{chapter}{Appendices}


\bibliographystyle{amsplain}
\bibliography{thesis}

\providecommand{\bysame}{\leavevmode\hbox to3em{\hrulefill}\thinspace}
\providecommand{\MR}{\relax\ifhmode\unskip\space\fi MR }
\providecommand{\MRhref}[2]{%
  \href{http://www.ams.org/mathscinet-getitem?mr=#1}{#2}
}
\providecommand{\href}[2]{#2}
\begin{thebibliography}{10}

\bibitem{aziz2020exploring}
Lubna Aziz, Sah bin Haji~Salam, and Sara Ayub, \emph{Exploring deep
  learning-based architecture, strategies, applications and current trends in
  generic object detection: A comprehensive review}, IEEE Access (2020).

\bibitem{babenko2009visual}
Boris Babenko, Ming-Hsuan Yang, and Serge Belongie, \emph{Visual tracking with
  online multiple instance learning}, 2009 IEEE Conference on computer vision
  and Pattern Recognition, IEEE, 2009, pp.~983--990.

\bibitem{ball1965isodata}
Geoffrey~H Ball and David~J Hall, \emph{Isodata, a novel method of data
  analysis and pattern classification}, Tech. report, Stanford research inst
  Menlo Park CA, 1965.

\bibitem{bolme2010visual}
David~S Bolme, J~Ross Beveridge, Bruce~A Draper, and Yui~Man Lui, \emph{Visual
  object tracking using adaptive correlation filters}, 2010 IEEE computer
  society conference on computer vision and pattern recognition, IEEE, 2010,
  pp.~2544--2550.

\bibitem{carranza2020fast}
Cesar Carranza, Daniel Llamocca, and Marios Pattichis, \emph{Fast and scalable
  2d convolutions and cross-correlations for processing image databases and
  videos on cpus}, 2020 IEEE Southwest Symposium on Image Analysis and
  Interpretation (SSIAI), IEEE, 2020, pp.~70--73.

\bibitem{cerna2021deep}
Alvaro E~Ulloa Cerna, Linyuan Jing, Christopher~W Good, Sushravya Raghunath,
  Jonathan~D Suever, Christopher~D Nevius, Gregory~J Wehner, Dustin~N Hartzel,
  Joseph~B Leader, Amro Alsaid, et~al., \emph{Deep-learning-assisted analysis
  of echocardiographic videos improves predictions of all-cause mortality},
  Nature Biomedical Engineering (2021), 1--9.

\bibitem{chen2019mmdetection}
Kai Chen, Jiaqi Wang, Jiangmiao Pang, Yuhang Cao, Yu~Xiong, Xiaoxiao Li,
  Shuyang Sun, Wansen Feng, Ziwei Liu, Jiarui Xu, et~al., \emph{Mmdetection:
  Open mmlab detection toolbox and benchmark}, arXiv preprint arXiv:1906.07155
  (2019).

\bibitem{darsey2018hand}
Callie~J Darsey, \emph{Hand movement detection in collaborative learning
  environment videos},  (2018).

\bibitem{dendorfer2020mot20}
Patrick Dendorfer, Hamid Rezatofighi, Anton Milan, Javen Shi, Daniel Cremers,
  Ian Reid, Stefan Roth, Konrad Schindler, and Laura Leal-Taix{\'e},
  \emph{Mot20: A benchmark for multi object tracking in crowded scenes}, arXiv
  preprint arXiv:2003.09003 (2020).

\bibitem{deng2009imagenet}
Jia Deng, Wei Dong, Richard Socher, Li-Jia Li, Kai Li, and Li~Fei-Fei,
  \emph{Imagenet: A large-scale hierarchical image database}, 2009 IEEE
  conference on computer vision and pattern recognition, Ieee, 2009,
  pp.~248--255.

\bibitem{eilar2016distributed1}
Cody~W Eilar, Venkatesh Jatla, Marios~S Pattichis, Carlos L{\'o}pezLeiva, and
  Sylvia Celed{\'o}n-Pattichis, \emph{Distributed video analysis for the
  advancing out of school learning in mathematics and engineering project},
  2016 50th Asilomar Conference on Signals, Systems and Computers, IEEE, 2016,
  pp.~571--575.

\bibitem{eilar2016distributed}
Cody~Wilson Eilar, \emph{Distributed and scalable video analysis architecture
  for human activity recognition using cloud services},  (2016).

\bibitem{esakki2020adaptive}
Gangadharan Esakki, \emph{Adaptive encoding for constrained video delivery in
  hevc, vp9, av1 and vvc compression standards and adaptation to video
  content}, Ph.D. thesis, The University of New Mexico, 2020.

\bibitem{esakki2016optimal}
Gangadharan Esakki, Venkatesh Jatla, and Marios~S Pattichis, \emph{Optimal hevc
  encoding based on gop configurations}, 2016 IEEE Southwest Symposium on Image
  Analysis and Interpretation (SSIAI), IEEE, 2016, pp.~25--28.

\bibitem{esakki2017adaptive}
\bysame, \emph{Adaptive high efficiency video coding based on camera activity
  classification.}, DCC, 2017, p.~438.

\bibitem{esakki2020comparative}
Gangadharan Esakki, Andreas Panayides, Sravani Teeparthi, and Marios Pattichis,
  \emph{A comparative performance evaluation of vp9, x265, svt-av1, vvc codecs
  leveraging the vmaf perceptual quality metric}, Applications of Digital Image
  Processing XLIII, vol. 11510, International Society for Optics and Photonics,
  2020, p.~1151010.

\bibitem{everingham2007pascal}
Mark Everingham, Luc Van~Gool, Christopher~KI Williams, John Winn, and Andrew
  Zisserman, \emph{The pascal visual object classes challenge 2007 (voc2007)
  results},  (2007).

\bibitem{everingham2011pascal}
Mark Everingham and John Winn, \emph{The pascal visual object classes challenge
  2012 (voc2012) development kit}, Pattern Analysis, Statistical Modelling and
  Computational Learning, Tech. Rep \textbf{8} (2011).

\bibitem{grabner2006real}
Helmut Grabner, Michael Grabner, and Horst Bischof, \emph{Real-time tracking
  via on-line boosting.}, Bmvc, vol.~1, Citeseer, 2006, p.~6.

\bibitem{held2016learning}
David Held, Sebastian Thrun, and Silvio Savarese, \emph{Learning to track at
  100 fps with deep regression networks}, European conference on computer
  vision, Springer, 2016, pp.~749--765.

\bibitem{henriques2014high}
Jo{\~a}o~F Henriques, Rui Caseiro, Pedro Martins, and Jorge Batista,
  \emph{High-speed tracking with kernelized correlation filters}, IEEE
  transactions on pattern analysis and machine intelligence \textbf{37} (2014),
  no.~3, 583--596.

\bibitem{jacoby2017context}
Abigail~R Jacoby, \emph{Context-sensitive human activity classification in
  video utilizing object recognition and motion estimation},  (2017).

\bibitem{jacoby2018context}
Abigail~Ruth Jacoby, Marios~S Pattichis, Sylvia Celed{\'o}n-Pattichis, and
  Carlos L{\'o}pezLeiva, \emph{Context-sensitive human activity classification
  in collaborative learning environments}, 2018 IEEE Southwest Symposium on
  Image Analysis and Interpretation (SSIAI), IEEE, 2018, pp.~1--4.

\bibitem{jatla2016automatic}
Venkatesh Jatla, \emph{Automatic segmentation of coronal holes in solar images
  and solar prediction map classification},  (2016).

\bibitem{jatla2019image}
Venkatesh Jatla, Marios~S Pattichis, and Charles~Nick Arge, \emph{Image
  processing methods for coronal hole segmentation, matching, and map
  classification}, IEEE Transactions on Image Processing \textbf{29} (2019),
  1641--1653.

\bibitem{kalal2010forward}
Zdenek Kalal, Krystian Mikolajczyk, and Jiri Matas, \emph{Forward-backward
  error: Automatic detection of tracking failures}, 2010 20th international
  conference on pattern recognition, IEEE, 2010, pp.~2756--2759.

\bibitem{kalal2011tracking}
\bysame, \emph{Tracking-learning-detection}, IEEE transactions on pattern
  analysis and machine intelligence \textbf{34} (2011), no.~7, 1409--1422.

\bibitem{kent2020design}
Robert~B Kent and Marios~S Pattichis, \emph{Design, implementation, and
  analysis of high-speed single-stage n-sorters and n-filters}, IEEE Access
  (2020).

\bibitem{kristan2018sixth}
Matej Kristan, Ales Leonardis, Jiri Matas, Michael Felsberg, Roman Pflugfelder,
  Luka ˇCehovin~Zajc, Tomas Vojir, Goutam Bhat, Alan Lukezic, Abdelrahman
  Eldesokey, et~al., \emph{The sixth visual object tracking vot2018 challenge
  results}, Proceedings of the European Conference on Computer Vision (ECCV)
  Workshops, 2018, pp.~0--0.

\bibitem{lin2014microsoft}
Tsung-Yi Lin, Michael Maire, Serge Belongie, James Hays, Pietro Perona, Deva
  Ramanan, Piotr Doll{\'a}r, and C~Lawrence Zitnick, \emph{Microsoft coco:
  Common objects in context}, European conference on computer vision, Springer,
  2014, pp.~740--755.

\bibitem{liu2016ssd}
Wei Liu, Dragomir Anguelov, Dumitru Erhan, Christian Szegedy, Scott Reed,
  Cheng-Yang Fu, and Alexander~C Berg, \emph{Ssd: Single shot multibox
  detector}, European conference on computer vision, Springer, 2016,
  pp.~21--37.

\bibitem{lukezic2017discriminative}
Alan Lukezic, Tomas Vojir, Luka ˇCehovin~Zajc, Jiri Matas, and Matej Kristan,
  \emph{Discriminative correlation filter with channel and spatial
  reliability}, Proceedings of the IEEE conference on computer vision and
  pattern recognition, 2017, pp.~6309--6318.

\bibitem{milan2016mot16}
Anton Milan, Laura Leal-Taix{\'e}, Ian Reid, Stefan Roth, and Konrad Schindler,
  \emph{Mot16: A benchmark for multi-object tracking}, arXiv preprint
  arXiv:1603.00831 (2016).

\bibitem{mueller2016benchmark}
Matthias Mueller, Neil Smith, and Bernard Ghanem, \emph{A benchmark and
  simulator for uav tracking}, European conference on computer vision,
  Springer, 2016, pp.~445--461.

\bibitem{muller2018trackingnet}
Matthias Muller, Adel Bibi, Silvio Giancola, Salman Alsubaihi, and Bernard
  Ghanem, \emph{Trackingnet: A large-scale dataset and benchmark for object
  tracking in the wild}, Proceedings of the European Conference on Computer
  Vision (ECCV), 2018, pp.~300--317.

\bibitem{redmon2016you}
Joseph Redmon, Santosh Divvala, Ross Girshick, and Ali Farhadi, \emph{You only
  look once: Unified, real-time object detection}, Proceedings of the IEEE
  conference on computer vision and pattern recognition, 2016, pp.~779--788.

\bibitem{ren2016faster}
Shaoqing Ren, Kaiming He, Ross Girshick, and Jian Sun, \emph{Faster r-cnn:
  towards real-time object detection with region proposal networks}, IEEE
  transactions on pattern analysis and machine intelligence \textbf{39} (2016),
  no.~6, 1137--1149.

\bibitem{shi2016human}
WENJING SHI, \emph{Human attention detection using am-fm representations},
  (2016).

\bibitem{shi2018robust}
Wenjing Shi, Marios~S Pattichis, Sylvia Celed{\'o}n-Pattichis, and Carlos
  L{\'o}pezLeiva, \emph{Robust head detection in collaborative learning
  environments using am-fm representations}, 2018 IEEE Southwest Symposium on
  Image Analysis and Interpretation (SSIAI), IEEE, 2018, pp.~1--4.

\bibitem{make-sense}
Piotr Skalski, \emph{{Make Sense}},
  \url{https://github.com/SkalskiP/make-sense/}, 2019.

\bibitem{tapia2020importance}
Luis~Sanchez Tapia, Marios~S Pattichis, Sylvia Celed{\'o}n-Pattichis, and
  Carlos~L{\'o}pez Leiva, \emph{The importance of the instantaneous phase for
  face detection using simple convolutional neural networks}, 2020 IEEE
  Southwest Symposium on Image Analysis and Interpretation (SSIAI), IEEE, 2020,
  pp.~1--4.

\bibitem{van2014scikit}
Stefan Van~der Walt, Johannes~L Sch{\"o}nberger, Juan Nunez-Iglesias,
  Fran{\c{c}}ois Boulogne, Joshua~D Warner, Neil Yager, Emmanuelle Gouillart,
  and Tony Yu, \emph{scikit-image: image processing in python}, PeerJ
  \textbf{2} (2014), e453.

\bibitem{wang2019fast}
Qiang Wang, Li~Zhang, Luca Bertinetto, Weiming Hu, and Philip~HS Torr,
  \emph{Fast online object tracking and segmentation: A unifying approach},
  Proceedings of the IEEE/CVF Conference on Computer Vision and Pattern
  Recognition, 2019, pp.~1328--1338.

\bibitem{wudetectron2}
Yuxin Wu, Alexander Kirillov, Francisco Massa, Wan-Yen Lo, and Ross Girshick,
  \emph{Detectron2 (2019)}, URL https://github.
  com/facebookresearch/detectron2.

\bibitem{zhang2020fairmot}
Yifu Zhang, Chunyu Wang, Xinggang Wang, Wenjun Zeng, and Wenyu Liu,
  \emph{Fairmot: On the fairness of detection and re-identification in multiple
  object tracking}, arXiv preprint arXiv:2004.01888 (2020).

\end{thebibliography}

\end{document}